\PassOptionsToPackage{table}{xcolor}
\PassOptionsToPackage{numbers,sort}{natbib}
\documentclass[10pt,letterpaper]{fairmeta}
\usepackage[font=small]{caption}
\ifdefined\pdfcompresslevel\pdfcompresslevel=9\fi
\ifdefined\pdfobjcompresslevel\pdfobjcompresslevel=3\fi
\usepackage{amsmath,amssymb,amsfonts,mathtools}
\usepackage{graphicx,booktabs,array,adjustbox}
\usepackage{natbib}
\usepackage{placeins}
\usepackage{xurl}
\usepackage{microtype}
\usepackage{bbding}
\usepackage{hyperref}
\usepackage{event2reward_preprint}

\usepackage{amsmath,amsfonts,bm}

\def\eqref#1{equation~\ref{#1}}

\def\1{\bm{1}}

\DeclareMathAlphabet{\mathsfit}{\encodingdefault}{\sfdefault}{m}{sl}
\SetMathAlphabet{\mathsfit}{bold}{\encodingdefault}{\sfdefault}{bx}{n}

\definecolor{customRed}{RGB}{190,110,113}
\definecolor{customGreen}{RGB}{70,170,80}
\definecolor{customPurple}{RGB}{102,2,60}
\hypersetup{colorlinks=true,linkcolor=metablue,citecolor=metablue,urlcolor=metablue,
 pdftitle={Event2Reward: Learning Robotic Rewards from Progress Relations},
 pdfauthor={Junnan Nie, Zitian Fan, Jiachen Zhang, Junyi Lao, Likui Zhang, Chenghao Liu, Xiaoyang Lu, Yunbo Wang, Yixue Zhang, Linbo Zhai, Jiayi Li, Mengzhe Wang, Xichang Wu, Bin Wang, Chang Liu, Tianle Zhang, Songfang Huang}}
\title{Event2Reward: Learning Robotic Rewards from Progress Relations}
\author[1,*]{Junnan Nie}
\author[1,*]{Zitian Fan}
\author[1]{Jiachen Zhang}
\author[1]{Junyi Lao}
\author[2]{Likui Zhang}
\author[1]{Chenghao Liu}
\author[1]{Xiaoyang Lu}
\author[1]{Yunbo Wang}
\author[1]{Yixue Zhang}
\author[2]{Linbo Zhai}
\author[2]{Jiayi Li}
\author[2]{Mengzhe Wang}
\author[1]{Xichang Wu}
\author[2]{Bin Wang}
\author[1]{Chang Liu}
\author[2]{Tianle Zhang}
\author[1,\text{\normalfont\Envelope}]{Songfang Huang}
\affiliation[1]{Peking University}
\affiliation[2]{JD Explore Academy}
\contribution[*]{Equal contribution}
\contribution[\text{\normalfont\Envelope}]{Corresponding author}
\abstract{
Learning task progress from robot demonstrations provides a practical way to
construct dense shaping rewards for robotic reinforcement learning. A common
approach uses normalized demonstration time as progress labels, implicitly
assuming monotonically increasing task progress. During policy exploration,
however, actual task progress can stall or even regress. Such models may
mistake regressions for progress, assigning positive shaping rewards to
actions that cause progress loss. Real suboptimal and failed executions
offer corrective experience, but assigning absolute progress values to their
states is difficult. Our key observation is that stalls, regressions, and
recoveries establish local relative progress relations even when absolute
progress is unknown. We introduce Event2Reward, which uses normalized
temporal labels to anchor the progress scale and these within-trajectory
relations to constrain local changes, combining absolute and relative
progress supervision. Its progress predictions serve as potentials for
reward shaping. We curate an event-annotated dataset from approximately 100K
raw real-robot execution records, covering successful, suboptimal, and
failed executions, without requiring framewise absolute progress labels.
Under the same training budgets, Event2Reward improves average policy
success by 18.16 percentage points over a baseline using only sparse task
rewards across 40 tasks in four LIBERO suites. On real robots, average
success after 30 minutes of wall-clock training reaches 60.0\% on three
precise insertion tasks and 90.0\% on three object manipulation tasks,
exceeding the sparse-reward baseline under the same budget by 36.7 and 23.3
percentage points, respectively. Ablation studies further show that absolute
progress supervision and relative progress relations are complementary, with
joint supervision yielding better downstream policy learning.
}
\metadata[\resourceicon{browser}Project Page]{\href{https://niejunnan25.github.io/Event2Reward/}{https://niejunnan25.github.io/Event2Reward/}}
\metadata[\resourceicon{github}Code]{\href{https://github.com/niejunnan25/Event2Reward}{https://github.com/niejunnan25/Event2Reward}}

\begin{document}
\maketitle

\begin{figure}[t]
\centering
\includegraphics[width=\linewidth]{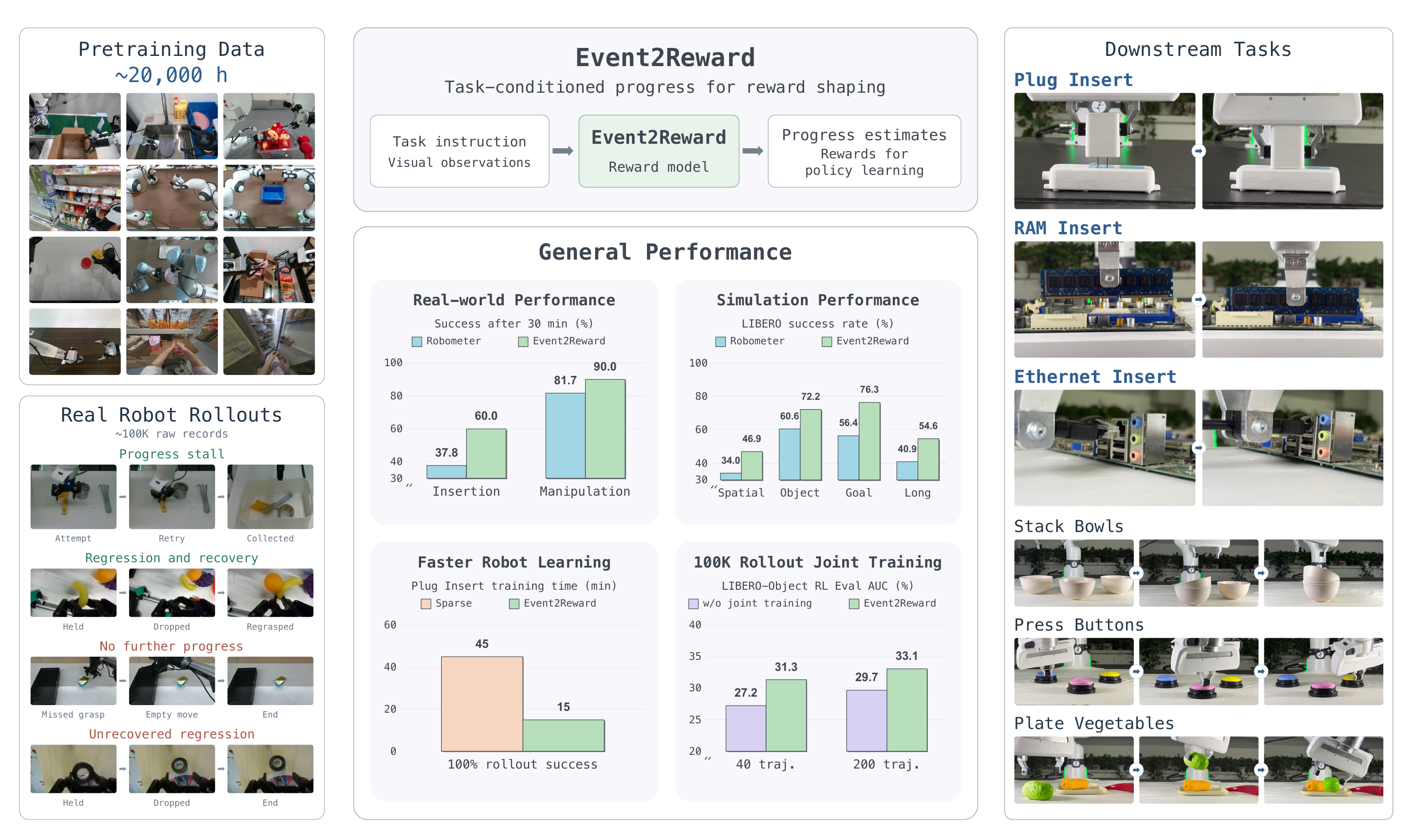}
\caption{\textbf{Overview of Event2Reward.}
    Left: pretraining data and real robot rollouts containing both
    successful and failed executions. Task events provide
    within-trajectory progress relations that complement temporal
    labels from successful demonstrations. Top center: task-conditioned
    progress predictions support reward shaping. Right: six real-world
    downstream tasks. The performance panels summarize real-world and
    LIBERO policy success, plug-insertion learning time, and the effect
    of rollout-based joint training. Plug-insertion time is measured
    until a rolling window of 10 autonomous training episodes reaches
    100\% success.}
\label{fig:roboground-overview}
\end{figure}

\section{Introduction}
\label{sec:introduction}

Reinforcement learning (RL) enables robots to improve manipulation
skills through trial and error~\citep{luo2025hilserl}.
However, learning on real robots
often requires extensive and costly interaction. Reward shaping
offers a way to accelerate policy learning by improving
credit assignment with dense rewards.
A common approach learns task progress from successful demonstrations,
using normalized elapsed time as a
proxy~\citep{liang2026robometer,wang2026wvm}.
This temporal supervision encodes
monotonically increasing task progress. Progress estimates can serve
as potentials for reward shaping, with the shaping term computed from
discounted differences between successive predictions~\citep{ng1999policy}.
During policy exploration,
however, robots routinely produce suboptimal and failed executions,
in which progress can stall or even regress. The reward model may
still predict rising progress during such non-ideal behavior,
assigning positive shaping rewards even when progress
stalls or regresses. Instead of
improving credit assignment, such rewards credit the actions
that caused the loss, hindering rather than accelerating policy
improvement.

A natural remedy is to train the reward model directly on
suboptimal and failed executions. However, normalized time
continues to assign increasing progress values during stalls
and regressions, so it cannot provide valid absolute progress
supervision for these executions. A common workaround repeats
or reverses demonstration frames to synthesize stalls and
regressions with labels known by
construction~\citep{liang2026robometer,wang2026wvm}, but these
edits cannot introduce failure states absent from the original
demonstrations. Another workaround uses outcome labels or
trajectory-level preferences, which do enable reward learning
from real suboptimal and failed
executions~\citep{lee2026roboreward,liang2026robometer}. These
signals describe overall execution quality, whereas progress-based
shaping rewards depend on progress changes between adjacent
observations. Such overall judgments therefore leave the timing
and direction of these changes underdetermined: a failed execution
can retain partial progress, while a successful one can include
regressions and recoveries. How, then, can we obtain local progress
supervision from real executions without assigning an absolute
progress value to every state?

Our key observation is that task events in real executions can
establish local progress relations even when absolute progress
values are unknown. For example, consider a robot moving a cup
from a table onto a shelf. If the cup slips from the gripper and
falls back onto the table, how much progress remains is difficult
to quantify, yet this state clearly holds less task progress than
the state just before the drop. We directly supervise the progress
predictions used for reward shaping with these within-trajectory
progress relations, which encode the direction of change between
temporally anchored states. We combine these relations with
normalized temporal labels from successful demonstrations to
train our reward model, Event2Reward (Figure~\ref{fig:roboground-overview}).
The temporal labels anchor the numerical scale of predicted
progress, while the relations constrain the direction of local
progress changes in real suboptimal and failed executions.

We also contribute an event-annotated dataset curated from
approximately 100K raw real-robot execution records across multiple
sources, spanning successful, suboptimal, and failed executions.
The annotations identify where task progress stalls, regresses,
or recovers, and link these events to temporal regions and
reference states. They turn observed execution events into local
progress comparisons beyond terminal outcomes, without requiring
framewise completion percentages. We combine two-pass VLM annotation
with independent human annotation and review
(Appendix~\ref{app:annotation-procedure}).
We will release the structured annotations, data splits, and tools
for constructing progress comparisons, providing a reusable resource
for reward learning and execution analysis.
We train Event2Reward through pretraining and rollout joint training,
then adapt it using a small set of target-task executions.

We evaluate whether Event2Reward improves policy learning on 40
LIBERO~\citep{liu2023libero} tasks and six real-world tasks on Franka
robots. The real-world experiments cover three precise insertion tasks
trained with SERL~\citep{luo2025serl,luo2025hilserl} and three manipulation
tasks trained with DSRL~\citep{wagenmaker2025dsrl} using a
$\pi_0$~\citep{black2024pi0} base policy. Across the four LIBERO suites,
Event2Reward improves average policy success by 9.96--13.76 percentage
points over the strongest baseline in each suite under matched
interaction budgets. In real-world experiments with a matched 30-minute
training budget, Event2Reward increases average success from 37.8\% to
60.0\% on precise insertion tasks and from 81.7\% to 90.0\% on manipulation
tasks, compared with Robometer, the strongest learned reward baseline
in each setting. Our ablations show that combining
temporal labels with within-trajectory progress relations yields
larger gains in downstream policy learning than combining them
with trajectory-level preferences. Combining temporal and relational
supervision outperforms either alone in downstream policy learning.

\section{Related Work}
\label{sec:related_work}

\paragraph{Visual reward and value models.}
Learning rewards from visual observations reduces the need for task-specific reward engineering in robot learning.
VIP~\citep{ma2023vip} learns goal-directed visual representations through value-based pretraining, and LIV~\citep{ma2023liv} extends this approach to language and image goals.
For vision-language reasoning, GVL~\citep{ma2024gvl} estimates progress in context without parameter updates, while ProgressLM~\citep{zhang2026progresslm} learns to assess progress relative to visual or textual demonstrations.
WVM~\citep{wang2026wvm} instead uses video world-model representations to predict distributions over progress sequences.
Our work focuses on the supervision of progress estimates: how real suboptimal and failed executions can provide evidence about local progress changes.

\paragraph{Progress supervision from demonstrations.}
Demonstrations provide temporal order and semantic task structure for progress supervision.
Temporal order supports ranking-based objectives: Rank2Reward~\citep{yang2024rank2reward} learns rankings on expert videos and combines them with an expert--policy discriminator, while Video-Language Critic~\citep{alakuijala2025vlc} combines sequential ranking with video--text matching and uses failed executions as contrastive negatives.
VLAC~\citep{zhai2025vlac} derives signed progress differences from temporal offsets.
RynnValue~\citep{huang2026rynnvalue} learns absolute temporal distance to task completion and relative temporal displacement from timestamps, using temporal-order shuffling and value-isolation attention to reduce shortcut learning.
Task-stage annotations provide semantic structure: SARM~\citep{chen2025sarm} aligns numerical progress targets with annotated stages, and Robo-Dopamine~\citep{tan2025robodopamine} derives signed progress differences from stage-based progress coordinates.
Augmentation extends this supervision to nonmonotonic sequences.
ReWiND~\citep{zhang2025rewind} reverses demonstration segments to synthesize regressions, while WVM~\citep{wang2026wvm} constructs increasing, stable, and decreasing profiles through frame preservation, repetition, and reversal.
Our approach complements these signals with progress relations grounded in stalls, regressions, and recoveries in real suboptimal and failed executions.

\paragraph{Reward learning from nonideal experience.}
Preference-based reward learning uses relative judgments without requiring absolute numerical ratings~\citep{christiano2017preferences}.
At the observation level, RL-VLM-F~\citep{wang2024rlvlmf} asks a VLM to compare images, and \citet{venkataraman2025offlinevlm} apply this approach to reward labeling for real-world offline RL.
At the trajectory level, RoboReward~\citep{lee2026roboreward} combines instruction relabeling and temporal clipping of successful demonstrations with naturally successful and failed rollouts carrying episode-level completion scores.
Robometer~\citep{liang2026robometer} combines normalized temporal progress labels from expert demonstrations with success prediction and trajectory preferences, allowing real failures to contribute supervision while also producing per-frame progress estimates.
Beyond these judgments, execution events provide structure for datasets and evaluation.
WVM's Suboptimal-Value-Bench~\citep{wang2026wvm} converts annotated hesitation and retry intervals into reference progress curves, while REBOOT~\citep{oforiampofo2026reboot} pairs precision-assembly demonstrations with phase boundaries, failure and recovery timestamps, and a phase-level evaluation protocol.
We use event-grounded within-trajectory relations to supervise the progress estimates used for reward shaping.

\section{Method}
\label{sec:method}

\begin{figure}[t]
    \centering
    \includegraphics[width=\linewidth]{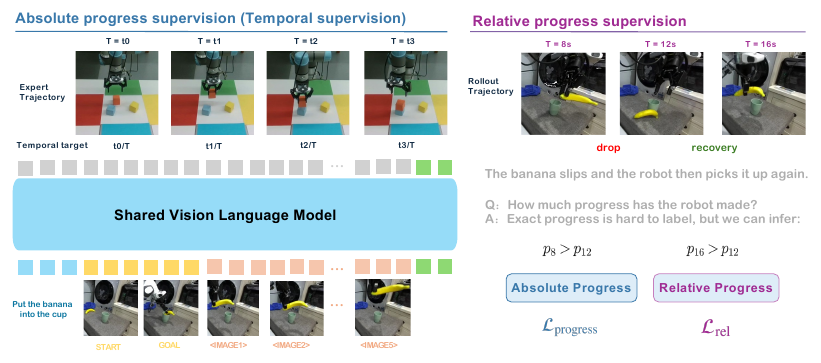}
    \caption{\textbf{Overview of progress supervision in Event2Reward.}
    Left: normalized temporal labels from successful expert trajectories
    anchor the progress scale of a shared vision-language model.
    Right: execution events in robot rollouts support relative constraints,
    illustrated by a drop and subsequent recovery
    ($p_8>p_{12}$ and $p_{16}>p_{12}$).
    These complementary signals supervise the same progress model;
    relation supervision uses both successful and failed executions.}
    \label{fig:progress-supervision-overview}
\end{figure}

Event2Reward learns task progress from temporal labels and within-trajectory
progress relations. After adaptation to downstream tasks, the model's
scalar progress predictions serve as the potential for reward shaping.
Figure~\ref{fig:progress-supervision-overview} illustrates these
complementary supervision signals.

\subsection{Task Progress Model}
\label{sec:progress-model}

Event2Reward is a task-conditioned model that estimates task progress
from visual observations. At time $t$, its input is
$x_t=(\ell,o_0,H_t,g)$, where $\ell$ is the task instruction,
$o_0$ is a reference start observation, $H_t$ is a short visual
history ending at the current observation, and $g$ is an optional
goal image depicting task completion. It outputs a scalar progress
estimate $p_\theta(x_t)\in[0,1]$, with higher values indicating
greater progress toward completing the instructed task. The visual
history provides context for interpreting local changes: progress
may remain stable during stalled execution or decrease when
previously achieved task states are lost.

We implement Event2Reward with a Qwen3.5-4B vision-language backbone,
a progress head, and an auxiliary trend head. The progress head
predicts a distribution over discretized progress values in
$[0,1]$, whose expectation gives $p_\theta(x_t)$. As detailed in
Sections~\ref{sec:event-dataset} and~\ref{sec:joint-learning},
temporal labels supervise this distribution, while within-trajectory
progress relations constrain its scalar expectation. The same
scalar serves as the potential for downstream reward shaping.
The trend head predicts whether progress increases, remains stable,
or decreases within the visual history as an auxiliary training
task. Input construction and prediction-head details are given in
Appendix~\ref{app:model-inputs}.

\subsection{Progress Supervision from Robot Executions}
\label{sec:event-dataset}

We curate an event-annotated dataset from approximately 100K
raw real-robot execution records.
The annotations connect observable execution events to
temporal anchors and reference states, providing local
progress comparisons beyond terminal success or failure.
Normal successful trajectories supply normalized-time
labels as a progress proxy over their task-relevant
execution intervals.
Trajectories containing events supply within-trajectory
relations that capture stalled execution, loss of achieved
task states, and recovery.

For each execution event, we record its category, a reference state,
and the relevant temporal interval; regression-and-recovery events
additionally identify a recovery state. Together with the supporting
visual observations, these temporal anchors identify the within-trajectory
progress comparisons used for supervision. Annotation therefore consists
of identifying task events and their temporal anchors, without assigning
framewise completion percentages.

Our annotation protocol organizes events by execution outcome,
while their progress relations are grounded in observable
task-state changes. For successful executions, we annotate
\emph{progress stall}, an interval with approximately unchanged
task progress, and \emph{regression and recovery}, a decrease
in task progress followed by recovery to the pre-regression
reference level. For failed executions, we distinguish
\emph{no further progress}, where no advancement beyond a
reference state is observed before recording ends, and
\emph{unrecovered regression}, where task progress falls below
its pre-regression reference level and does not return to
that level before the recording ends.
This organization does not imply that stalls or local recoveries
occur only in successful executions. Event judgments depend on
the task instruction and observable changes relative to the
reference states. An unsuccessful grasp attempt is not regression
unless a grasp had already been established, and renewed motion is not
recovery unless the relevant task state is restored.
A failed execution may retain partial progress, so failure alone
does not establish unrecovered regression.

We construct event annotations through two-pass annotation with
GPT-5.6 Sol and independent human annotation. GPT-5.6 Sol first
identifies candidate events and temporal anchors across each execution,
then refines them using denser local observations and later context
when needed. Human inspection covers the raw real-robot collection,
and each retained trajectory is independently annotated without access
to the machine labels. Disagreements and unresolved cases undergo
third-party review, while stratified samples of agreement cases are
independently audited. Only visually supported progress comparisons
are used for supervision. On the 8,146 retained RoboArena trajectories,
machine and independent human annotations agree on the primary category
for 87.56\% of trajectories. Full annotation procedures and audit
statistics are provided in Appendix~\ref{app:annotation-procedure}.

We convert the reviewed event annotations into within-trajectory progress
comparisons. For each event $e$, its temporal interval locates candidate
observations, and only those supported by the annotated task-state
relation enter the comparison set $\mathcal M_e$. For example, when a
dropped cup is regrasped and lifted, observations showing
the loss of the established grasp should have lower progress than both
the pre-drop reference and the recovered state. The annotations specify
the observations to compare and the direction of their progress relation;
the model learns the numerical scores through the relation loss in
Section~\ref{sec:joint-learning}. Additional definitions and examples
are provided in Appendix~\ref{app:event-definitions}.

\subsection{Learning Objectives}
\label{sec:joint-learning}

\paragraph{Temporal supervision.}
For successful trajectories without annotated events, temporal labels
are $y_t=(t-t_0)/(t_S-t_0)$ over the task-relevant interval from $t_0$
to completion $t_S$. We train the progress distribution with
cross-entropy against a soft encoding with mean $y_t$, yielding
$\mathcal L_{\mathrm{progress}}$. During pretraining and replay,
successful histories and their temporal transformations provide labels
for the auxiliary trend cross-entropy loss $\mathcal L_{\mathrm{trend}}$
(Appendix~\ref{app:progress-trend}). Reference and comparison
observations used for relation supervision receive neither temporal nor
trend labels.

\paragraph{Progress relations.}
Event categories select constraints on the scalar progress prediction
$p_t=p_\theta(x_t)$, rather than serving as classification targets.
Reference and comparison scores use the same model and task conditions;
reference scores are recomputed as the model changes, with gradients
stopped through them in the relation loss. For a regression-and-recovery
event $e$, let $\bar p_a$ and $\bar p_r$ be the reference and recovery
scores. For each $t\in\mathcal M_e$, we use
\begin{equation}
\ell_e(t)=\bigl[p_t-\min(\bar p_a,\bar p_r)+m_e\bigr]_+^2,
    \label{eq:event-relations}
\end{equation}
where $[z]_+=\max(z,0)$ and $m_e$ is a margin determined by the training
rule, not an annotated amount of progress loss. This penalizes
insufficient separation below either reference. Progress stall permits a
tolerance around the reference; no further progress penalizes scores
above it; unrecovered regression requires a decrease margin without a
recovery reference. We average the corresponding penalties within each
event and then across events to obtain $\mathcal L_{\mathrm{rel}}$. Full
loss and margin definitions are given in
Appendix~\ref{app:event-losses}.

\paragraph{Joint objective.}
We jointly optimize $\mathcal L=\lambda_p\mathcal
L_{\mathrm{progress}}+\lambda_{\mathrm{rel}}\mathcal
L_{\mathrm{rel}}+\lambda_t\mathcal L_{\mathrm{trend}}$, with weights
controlling the three contributions. Temporal labels anchor the
numerical progress scale, while progress relations constrain local
changes in real executions without assigning absolute targets to event
observations. This distinction matters for reward shaping, whose
magnitude depends on the prediction scale at a fixed shaping
coefficient. Section~\ref{sec:ablation} evaluates the complementary
contributions of these two supervision signals.

\paragraph{Training and adaptation.}
Training proceeds through pretraining, rollout joint training, and
target-task adaptation. We pretrain Event2Reward on approximately
20,000 hours of robot expert
demonstrations and human manipulation videos using temporal labels and
auxiliary trend supervision. Rollout joint training then combines temporal
supervision from normal successful rollouts, relation supervision from
event-bearing rollouts, and replay of pretraining data. Target-task
adaptation uses demonstrations and event-bearing executions under the
same joint objective. Training settings are provided in
Appendix~\ref{app:training-configuration}, and adaptation
protocols are given in Section~\ref{sec:experiments}.

\subsection{Progress-Based Reward Shaping}
\label{sec:reward-shaping}

Event2Reward converts learned task progress into dense reward signals
for downstream policy learning. After target-task adaptation, we freeze
the reward model and predict task progress $p_t$ during execution.
Following potential-based reward shaping~\citep{ng1999policy}, we add
the discounted difference between successive progress predictions to
the environment reward,
\begin{equation}
r_t^{\mathrm{shaped}}=r_t^{\mathrm{env}}+\beta\left[\gamma p_{t+1}-p_t\right],
    \label{eq:progress-reward}
\end{equation}
where $r_t^{\mathrm{env}}$ is the environment reward and $\gamma$ is
the RL discount factor. We set $\beta=1$ by default and assign zero
potential to absorbing terminal states.

The environment reward provides feedback on task completion, while
the shaping term supplies local progress feedback during execution,
providing a learning signal for policy updates before completion.
Progress predictions should therefore reflect stalls, regressions,
and recoveries. In particular, they should decrease when previously
achieved task states are lost, to avoid assigning misleading positive
shaping rewards to actions that cause progress loss.

Event annotations are used to train the progress model. During policy
learning, rewards are computed directly from its predictions, without
requiring online event categories or recovery annotations. We use this
reward construction in both simulation and real-robot experiments.
Action-chunk reward computation and episode-boundary handling are
detailed in Appendices~\ref{app:ablation}
and~\ref{app:real-world-protocols}.

\section{Experiments}
\label{sec:experiments}

We first use controlled ablations to compare supervision from
within-trajectory progress relations and trajectory-level preferences,
and to examine the complementary roles of progress relations and temporal
labels. We then compare Event2Reward with baseline methods for policy
learning in simulation. To assess the contribution of rollout-based joint
training, we compare models under the same target-task adaptation budget.
Finally, we evaluate whether Event2Reward improves policy success and
learning efficiency on real robots.

\subsection{Supervision Ablations}
\label{sec:ablation}

\begin{figure}[t]
    \centering
    \includegraphics[width=\linewidth]{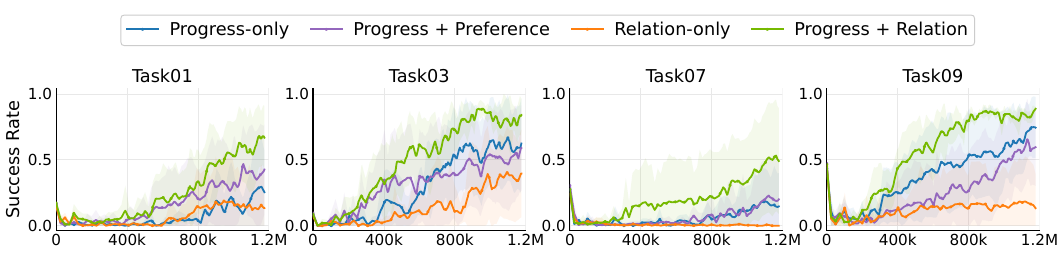}
    \caption{\textbf{Policy learning on four LIBERO-10 tasks with different reward supervision.}
    Lines show mean independent-evaluation success rates across five
    environment seeds; shaded bands indicate $\pm1.96\,\mathrm{SEM}$.
    Horizontal axes show environment steps up to the shared 1.2M budget.}
    \label{fig:ablation-supervision}
\end{figure}

The controlled supervision ablations use Qwen3-VL-4B-Instruct, matching
Robometer's backbone. Event2Reward in
Sections~\ref{sec:simulation}--\ref{sec:real-world} uses Qwen3.5-4B.
All controlled variants start from the same checkpoint trained with
temporal labels on normal successful LIBERO-10 rollouts.
We vary the supervision during continued training
and evaluate the resulting reward models through downstream policy
learning on four LIBERO-10 tasks. Policy learning uses a residual SAC
framework adapted from PLD~\citep{xiao2025pld}, with both the base policy
and reward model frozen; predicted progress provides shaping rewards
for residual policy learning.
Within each task, all variants share the same downstream policy
initialization and a budget of 1.2 million environment steps.
Full protocols, statistical details, and additional
comparisons are provided in Appendix~\ref{app:ablation}.

\paragraph{Supervision type.}
We compare temporal labels alone (\emph{Progress-only}) with
temporal labels combined with trajectory-level preferences
(\emph{Progress + Preference}) or within-trajectory progress
relations (\emph{Progress + Relation}).
The latter two use the same additional trajectory pool and
identical temporal supervision.
We measure downstream policy learning using normalized evaluation
AUC, which summarizes average evaluation success over a common
interaction interval.
Progress + Relation achieves the highest mean AUC on all four tasks,
with a four-task average of 36.7\%, compared with 19.6\% for
Progress-only and 19.4\% for Progress + Preference
(Figure~\ref{fig:ablation-supervision}).
To distinguish the effect of relation supervision from that of mixed
sampling, we also evaluate \emph{Relation-loss-off}, which retains the
mixed-sampling scheme of Progress + Relation but disables the relation
loss. Its four-task mean AUC is 19.1\%, close to Progress-only and below
Progress + Relation (Table~\ref{tab:ablation-full-rl-results}
in Appendix~\ref{app:ablation}).
These results support a contribution from relation supervision beyond
mixed sampling and show that adding progress relations to temporal
labels yields larger downstream gains than adding trajectory-level
preferences in this setting.

\paragraph{Joint supervision.}
To examine the complementary roles of temporal labels and progress
relations, we compare Progress + Relation with Progress-only and
\emph{Relation-only}. The latter uses only progress relations during
continued training. Joint supervision achieves higher mean evaluation
AUC than either alternative on all four tasks
(Figure~\ref{fig:ablation-supervision}).

We diagnose this gap through relation satisfaction and within-trajectory
prediction range, without rescaling model outputs
(Table~\ref{tab:main-progress-diagnostics}; protocols in
Appendix~\ref{app:ablation-diagnostics}).
Relation-only has the highest relation satisfaction (58.86\%) but
near-zero outputs with a mean range of only 0.0021, consistent with weak
shaping at a fixed coefficient.
Joint supervision improves relation satisfaction over Progress-only
while retaining prediction variation. Prediction range alone does not
establish accuracy. Together with the RL results, these findings support
temporal labels for scale anchoring and relations for local constraints
in this setting.

\begin{table}[!htbp]
    \centering
    \caption{\textbf{Progress prediction diagnostics.}
    Relation satisfaction is evaluated on 46 unrecovered regression
    trajectories and averaged equally across their nine tasks.
    Prediction range is averaged equally across all 103 held-out
    trajectories covering ten LIBERO-10 tasks.}
    \label{tab:main-progress-diagnostics}
    \begin{tabular}{@{}lcc@{}}
        \toprule
        Supervision
        & Relation satisfaction rate (\%)
        & Mean prediction range \\
        \midrule
        Progress-only      & 41.62 & 0.5928 \\
        Relation-only      & 58.86 & 0.0021 \\
        Progress + Relation & 54.47 & 0.5097 \\
        \bottomrule
    \end{tabular}
\end{table}

\subsection{Policy Learning in Simulation}
\label{sec:simulation}

We compare Event2Reward with Sparse (environment rewards only),
Robometer~\citep{liang2026robometer}, and
\emph{Robo-Dopamine}~\citep{tan2025robodopamine} on four
LIBERO~\citep{liu2023libero} suites---Spatial, Object, Goal,
and LIBERO-10---covering 40 tasks.
Event2Reward's pretraining and rollout joint training exclude LIBERO data.
For each suite, Event2Reward and Robometer are adapted using the
same 200 target-suite trajectories, while Robo-Dopamine follows
its one-shot adaptation protocol with one expert
demonstration per task.
All reward models remain frozen during policy learning.
Within each task, all conditions share the same frozen
$\pi_{0.5}$~\citep{physicalintelligence2025pi05} base policy and use
the residual SAC framework from Section~\ref{sec:ablation} under matched
interaction budgets. Adaptation, policy-training, and evaluation details
are provided in Appendix~\ref{app:policy-learning-details}.

\begin{table}[tbp]
    \centering
    \begingroup
\definecolor{E2RShade}{HTML}{EEF0FF}
\definecolor{E2RRed}{HTML}{D51D27}
    \small
    \setlength{\tabcolsep}{5pt}
    \renewcommand{\arraystretch}{1.12}
    \caption{\textbf{RL performance on LIBERO.}
Success rates (\%) are mean $\pm$ sample standard deviation across five
 environment seeds. Each run contributes its last periodic independent
 evaluation within 200k environment steps for Spatial, Object, and Goal,
 and 1.2M for LIBERO-10.}
    \label{tab:libero-policy-performance}
    \begin{adjustbox}{max width=\linewidth}
\begin{tabular}{lcccc}
    \toprule
    Method & LIBERO-Spatial & LIBERO-Object & LIBERO-Goal & LIBERO-10 \\
    \midrule
Sparse & $36.80 \pm 6.80$ & $47.40 \pm 3.13$ & $66.36 \pm 3.39$ & $26.84 \pm 8.80$ \\
Robo-Dopamine & $36.72 \pm 4.77$ & $57.56 \pm 6.09$ & $64.00 \pm 3.88$ & $31.72 \pm 4.92$ \\
Robometer & $34.04 \pm 5.03$ & $60.56 \pm 6.53$ & $56.44 \pm 5.67$ & $40.88 \pm 19.00$ \\
\textbf{Event2Reward} & $\mathbf{46.92} \pm 6.94$ & $\mathbf{72.16} \pm 2.97$ & $\mathbf{76.32} \pm 3.57$ & $\mathbf{54.64} \pm 8.70$ \\
\rowcolor{E2RShade}
$\Delta$ & \textcolor{E2RRed}{$\mathbf{+10.12}$} & \textcolor{E2RRed}{$\mathbf{+11.60}$} & \textcolor{E2RRed}{$\mathbf{+9.96}$} & \textcolor{E2RRed}{$\mathbf{+13.76}$} \\
    \bottomrule
    \end{tabular}
    \end{adjustbox}
    \endgroup
\end{table}

\begin{figure}[tbp]
    \centering
    \includegraphics[width=\linewidth]{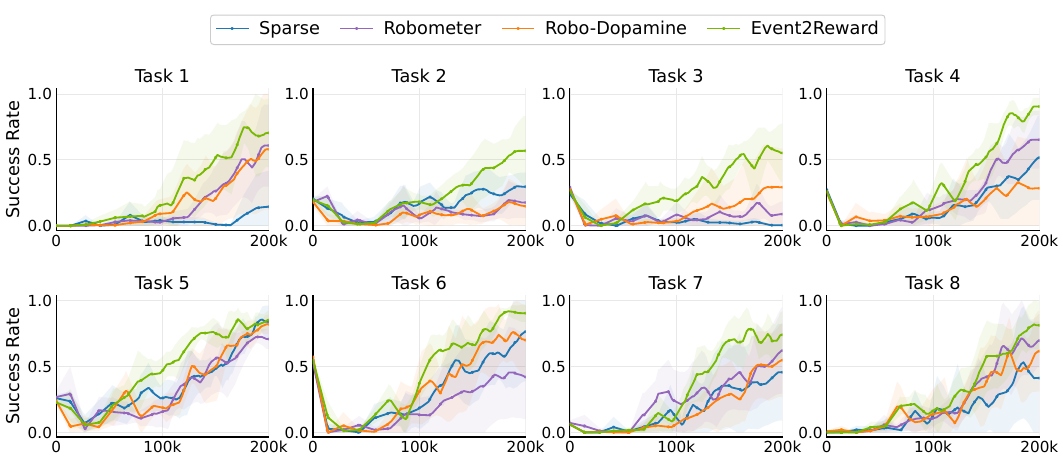}
    \caption{\textbf{Policy learning on eight LIBERO tasks with different reward models.}
    Lines show mean independent-evaluation success rates across five
    environment seeds; shaded bands indicate $\pm1.96\,\mathrm{SEM}$.
    Horizontal axes show environment steps up to the shared 200k budget.}
    \label{fig:simulation-selected-tasks}
\end{figure}

Event2Reward achieves the highest mean policy success on all four
LIBERO suites (Table~\ref{tab:libero-policy-performance}).
Compared with the strongest baseline in each suite, it improves success
by 10.12, 11.60, 9.96, and 13.76 percentage points on Spatial, Object,
Goal, and LIBERO-10, respectively.
Figure~\ref{fig:simulation-selected-tasks} shows learning curves for eight
of these tasks, which are listed in Table~\ref{tab:figure4-tasks};
Appendix~\ref{app:simulation-task-results} reports curves for all 40 tasks.

\FloatBarrier

\subsection{Contribution of Rollout-Based Joint Training}
\label{sec:rollout-joint-ablation}

We compare Event2Reward with a variant that skips rollout-based joint
training (\emph{w/o rollout joint training}). Both start from the same pretrained
checkpoint and use identical target-adaptation data, objectives, and
update budgets. We evaluate them on LIBERO-Object's ten tasks using an
average of 20 or 4 adaptation trajectories per task (the 200-trajectory
set from Section~\ref{sec:simulation} and its fixed 40-trajectory subset),
with the same base policy, RL settings, and 200k-step interaction budget.

\begingroup
\setlength{\intextsep}{6pt plus 2pt minus 2pt}
\begin{table}[!htbp]
    \centering
    \begingroup
    \definecolor{E2RShade}{HTML}{EEF0FF}
    \definecolor{E2RRed}{HTML}{D51D27}
    \small
    \setlength{\tabcolsep}{5pt}
    \renewcommand{\arraystretch}{1.12}
    \caption{\textbf{Effect of rollout joint training on LIBERO-Object.}
    Evaluation AUC and last periodic evaluation success (\%) within 200k
    environment steps are mean $\pm$ sample standard deviation across
    five environment seeds, averaged over ten tasks.}
    \label{tab:rollout-joint-ablation}
    \begin{adjustbox}{max width=\linewidth}
\begin{tabular}{lcccc}
    \toprule
    & \multicolumn{2}{c}{20 trajectories/task (avg.)}
    & \multicolumn{2}{c}{4 trajectories/task (avg.)} \\
    \cmidrule(lr){2-3}\cmidrule(lr){4-5}
    Method & Eval AUC & Success & Eval AUC & Success \\
    \midrule
    w/o rollout joint training & $29.67 \pm 2.66$ & $70.12 \pm 9.51$ & $27.22 \pm 2.24$ & $64.48 \pm 7.16$ \\
    \textbf{Event2Reward} & $\mathbf{33.06} \pm 2.28$ & $\mathbf{72.16} \pm 2.97$ & $\mathbf{31.34} \pm 2.58$ & $\mathbf{67.04} \pm 5.47$ \\
    \rowcolor{E2RShade}
    $\Delta$ & \textcolor{E2RRed}{$\mathbf{+3.39}$} & \textcolor{E2RRed}{$\mathbf{+2.04}$} & \textcolor{E2RRed}{$\mathbf{+4.13}$} & \textcolor{E2RRed}{$\mathbf{+2.56}$} \\
    \bottomrule
    \end{tabular}
    \end{adjustbox}
    \endgroup
\end{table}
\endgroup

Rollout-based joint training improves mean evaluation AUC by
3.39 and 4.13 percentage points with an average of 20 and 4 adaptation
trajectories per task, respectively, and success by 2.04 and 2.56 points
(Table~\ref{tab:rollout-joint-ablation}). Mean task-level AUC improves on
nine of ten tasks at each data size, supporting benefits even with fewer
adaptation trajectories. This comparison evaluates the training stage as
a whole, including its data mixture and additional optimization.
Appendix~\ref{app:rollout-joint-ablation} provides detailed protocols and
additional results.

\FloatBarrier

\subsection{Policy Learning on Real Robots}
\label{sec:real-world}

\begin{figure}[!htbp]
    \centering
    \begin{minipage}[t]{0.32\linewidth}
        \raggedright\small (a) Plug Insert\par\smallskip
        \includegraphics[width=\linewidth]{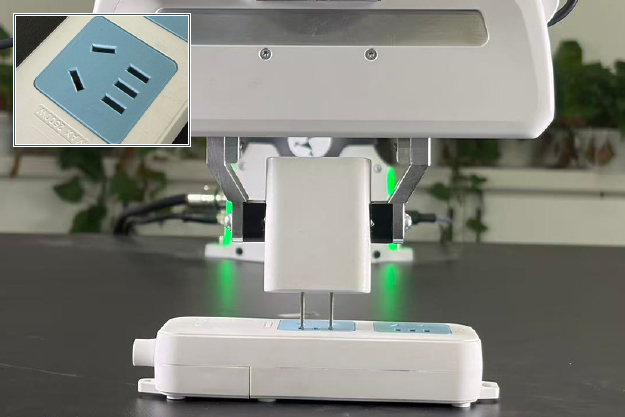}
    \end{minipage}\hfill
    \begin{minipage}[t]{0.32\linewidth}
        \raggedright\small (b) RAM Insert\par\smallskip
        \includegraphics[width=\linewidth]{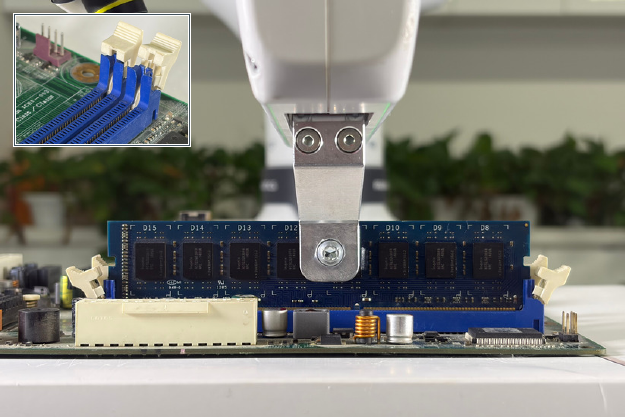}
    \end{minipage}\hfill
    \begin{minipage}[t]{0.32\linewidth}
        \raggedright\small (c) Ethernet Insert\par\smallskip
        \includegraphics[width=\linewidth]{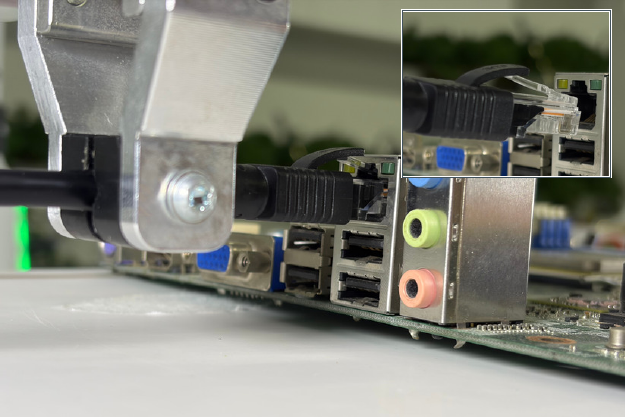}
    \end{minipage}
    \par\medskip
    \begin{minipage}[t]{0.32\linewidth}
        \raggedright\small (d) Stack Bowls\par\smallskip
        \includegraphics[width=\linewidth]{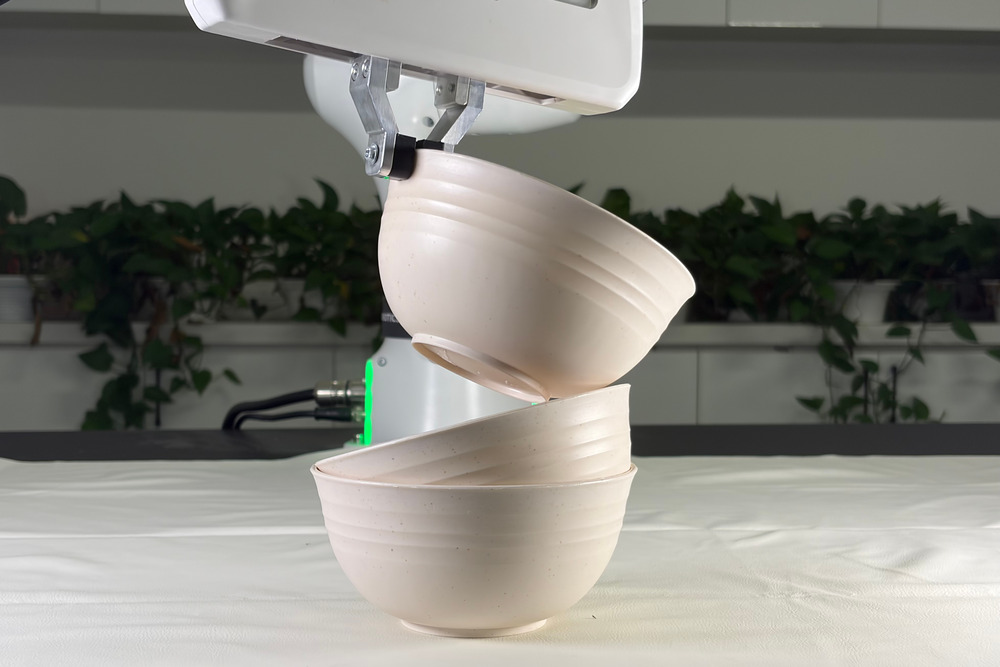}
    \end{minipage}\hfill
    \begin{minipage}[t]{0.32\linewidth}
        \raggedright\small (e) Press Buttons\par\smallskip
        \includegraphics[width=\linewidth]{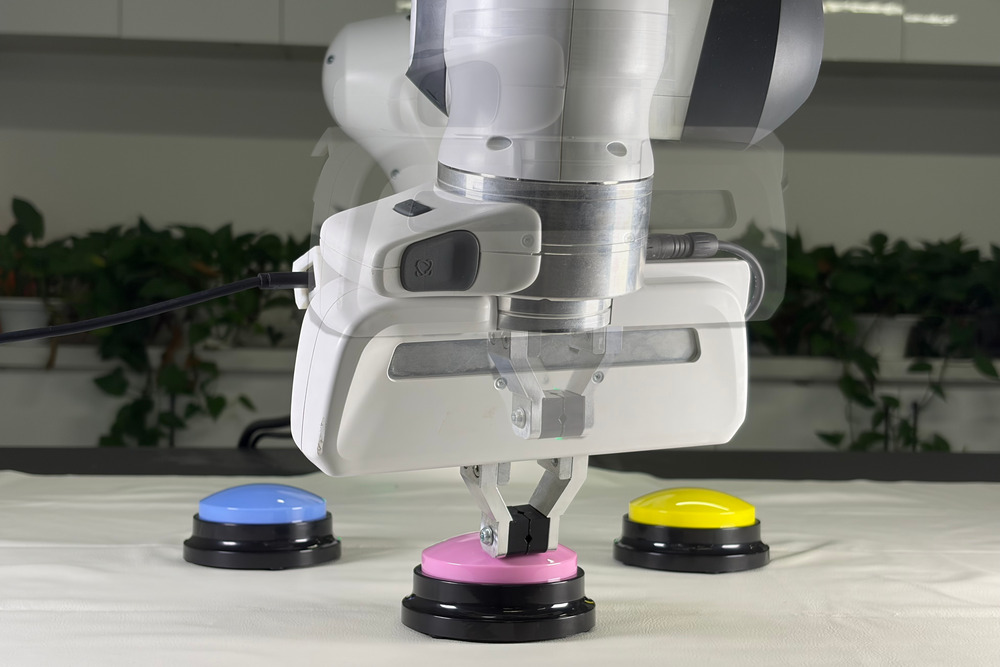}
    \end{minipage}\hfill
    \begin{minipage}[t]{0.32\linewidth}
        \raggedright\small (f) Plate Vegetables\par\smallskip
        \includegraphics[width=\linewidth]{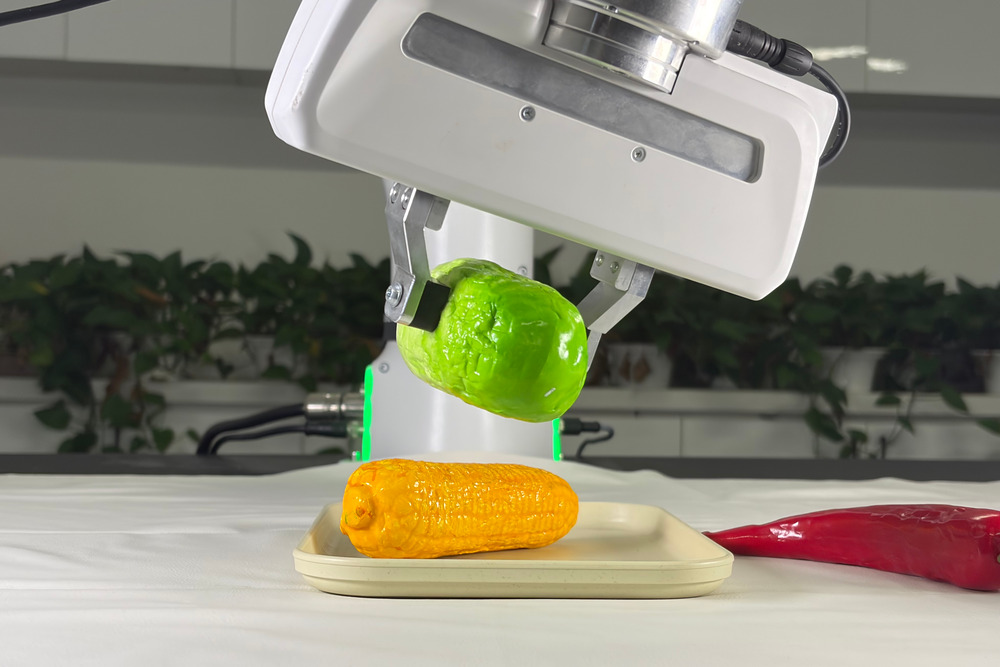}
    \end{minipage}
    \caption{\textbf{Real-world tasks for policy learning.}
        Top: insertion tasks trained with SERL. Bottom: manipulation tasks
        trained with DSRL using a $\pi_0$ base policy. Insets in (a) and (b)
        show oblique views of the socket and RAM slots.
        The inset in (c) shows the Ethernet plug and port
        in an earlier frame before insertion. In Press Buttons, the translucent
        overlay depicts an earlier end-effector pose during the second press,
        while the opaque pose shows contact with the pink button.}
    \label{fig:real-world-tasks}
\end{figure}

We evaluate Event2Reward on six real-world tasks using Franka robots:
three precise insertion tasks trained with SERL and three manipulation
tasks trained with DSRL using a $\pi_0$ base policy
(Figure~\ref{fig:real-world-tasks}).

\begin{figure}[!htbp]
    \centering
    \includegraphics[width=\linewidth]{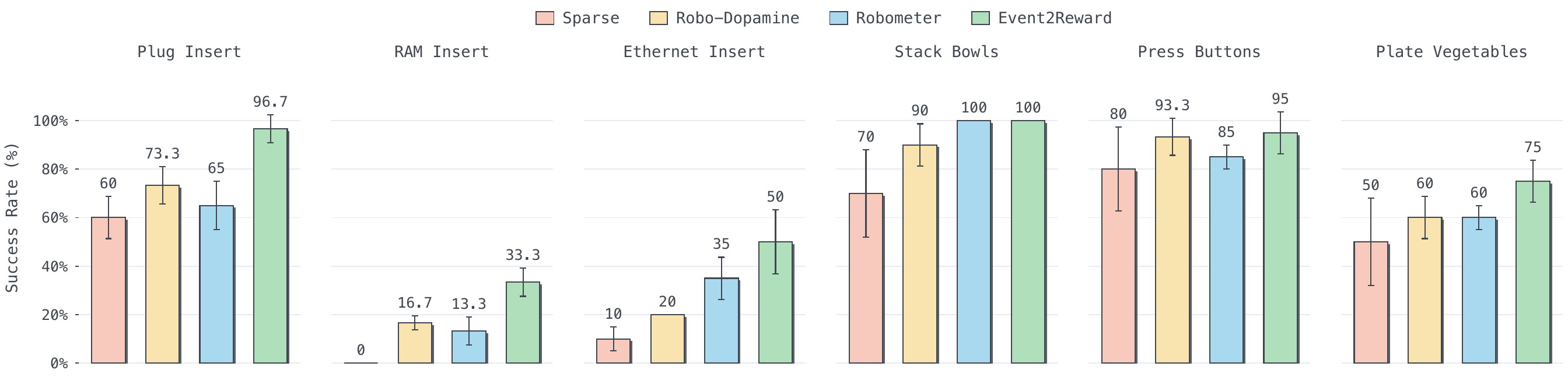}
    \caption{\textbf{Success rate (\%) after 30 minutes of wall-clock training,
        including initial buffer warm-up where applicable, resets, and reward
        computation.} Bars show means and error bars show sample standard deviations across
        three independent training runs per task and reward condition.
        Each run's checkpoint at 30 minutes is evaluated over
        20 autonomous episodes.}
    \label{fig:real-world-policy-performance}
\end{figure}

Within each framework, we compare Event2Reward with Sparse, Robometer,
and Robo-Dopamine. Event2Reward and Robometer use the same five successful
and five suboptimal adaptation trajectories per task, while Robo-Dopamine
follows its official one-shot adaptation protocol. Annotating events on
the five suboptimal trajectories takes about one minute of human effort
per task. Within each task,
reward models remain frozen during policy learning, and initialization
protocols and learning settings are matched. We independently evaluate
policies after 30 minutes of wall-clock training, including initial buffer
warm-up where applicable, resets, and reward computation. Full protocols
are provided in Appendix~\ref{app:real-world-protocols}.

Within this budget, Event2Reward increases average success
from 37.8\% to 60.0\% on precise insertion tasks and from 81.7\% to 90.0\%
on manipulation tasks, compared with Robometer, the strongest learned
reward baseline on average in each setting, gaining 22.2 and 8.3
percentage points, respectively. Event2Reward achieves the highest
mean success rate on five of the six tasks and ties with Robometer on
Stack Bowls (Figure~\ref{fig:real-world-policy-performance}).

We conduct an additional real-robot training experiment to assess the
end-to-end learning efficiency of the system. On plug insertion, we
compare Event2Reward shaping with sparse task rewards within the same
SERL framework, measuring wall-clock time to the first ten consecutive
successful autonomous training episodes, including warm-up data
collection and resets. In one additional training run per method,
Event2Reward reaches this criterion in 15 minutes, whereas using only
sparse rewards takes approximately 45 minutes
(Figure~\ref{fig:roboground-overview}; protocol in
Appendix~\ref{app:real-world-protocols}).

\FloatBarrier

\section{Conclusion}
\label{sec:conclusion}

We introduce Event2Reward, which combines temporal labels from
demonstrations with progress relations grounded in real execution events
to learn a task progress model for reward shaping in robotic
reinforcement learning. We curate an event-annotated dataset covering
successful, suboptimal, and failed executions to supervise local progress
changes. Experiments on 40 LIBERO tasks and six real-robot tasks show
improved policy success under the same training budgets. Controlled
ablations further support the complementary roles of temporal labels in
anchoring the numerical scale and relative progress relations in
constraining local changes. These results show that event-grounded
relations can provide effective supervision for robotic reward learning
even when absolute progress values are difficult to assign to states in
real suboptimal and failed executions.

\newpage
\bibliographystyle{plainnat}
\bibliography{refs}

\clearpage
\appendix

\section{Event-Annotated Robot Rollouts}
\label{app:event-dataset}

\subsection{Data Sources, Composition, and Splits}
\label{app:dataset-composition}

\paragraph{Data sources and scope.}
We curate approximately 100K raw real-robot execution records from multiple
sources. Public sources include RoboChallenge~\citep{yakefu2025robochallenge},
RW-RL~\citep{boden2026rwrl}, RoboArena~\citep{atreya2025roboarena},
ViFailback~\citep{zeng2026diagnose}, MolmoAct2~\citep{fang2026molmoact2},
ArmNetBench~\citep{selvaraj2026armnetbench}, the real-world subset of
RoboFAC~\citep{ye2026robofac}, and AutoEval~\citep{zhou2025autoeval}.
We organize visual observations, task instructions, and outcomes in a common
format for within-trajectory event annotation.

\paragraph{Execution quality and dataset composition.}
Normal successes complete the task without events identified by our protocol;
suboptimal successes complete it with events, and failures remain incomplete
when recording ends. Retention requires a task instruction, usable visual
observations, and either a valid normal-success completion endpoint or at
least one supported progress relation. Some retained trajectories have
uncertain outcomes but supported local relations.
Table~\ref{tab:dataset-source-composition} gives retained counts by source,
execution quality, and split.

\paragraph{Data splits.}
Each rollout and its derived clips and event annotations remain in one split;
source records identified as the same execution are grouped first.
Table~\ref{tab:dataset-source-composition} reports split sizes.
Validation and test trajectories are excluded from training, and training
goal images come only from the training split.

\begin{table}[t]
    \centering
    \caption{
        \textbf{Source composition of the real-robot data.}
        Raw counts enumerate source records; Retained counts
        enumerate unique trajectories included in the training,
        validation, or test partitions.
        Both the outcome columns and the split columns
        sum to Retained.
        Uncertain denotes trajectories with usable local
        progress relations but unresolved terminal outcomes.
    }
    \label{tab:dataset-source-composition}
    \setlength{\tabcolsep}{2.9pt}
    \renewcommand{\arraystretch}{1.15}
    \begin{adjustbox}{max width=\linewidth}
\begin{tabular}{@{}lrrrrrrrrr@{}}
        \toprule
        Source & Raw & Retained
        & \shortstack{Normal\\success}
        & \shortstack{Suboptimal\\success}
        & Failure & Uncertain
        & Train & Val. & Test \\
        \midrule
        \href{https://huggingface.co/RoboChallenge}
        {RoboChallenge}
        & 40,857 & 29,817 & 1,821 & 1,271 & 26,725 & 0
        & 23,761 & 3,044 & 3,012 \\

        \href{https://huggingface.co/datasets/MINT-SJTU/RW-RL-Dataset}
        {RW-RL}
        & 20,264 & 13,376 & 8,921 & 4,427 & 28 & 0
        & 12,073 & 651 & 652 \\

        \href{https://huggingface.co/datasets/RoboArena/DataDump_07-17-2026}
        {RoboArena}
        & 10,783 & 8,146 & 370 & 428 & 7,348 & 0
        & 6,974 & 565 & 607 \\

        \href{https://huggingface.co/datasets/sii-rhos-ai/ViFailback-Dataset}
        {ViFailback}
        & 5,202 & 3,689 & 480 & 53 & 3,156 & 0
        & 3,338 & 182 & 169 \\

        \href{https://huggingface.co/allenai}
        {MolmoAct2}
        & 2,412 & 1,632 & 106 & 152 & 1,374 & 0
        & 1,234 & 68 & 330 \\

        \href{https://huggingface.co/villekuosmanen}
        {ArmNetBench}
        & 3,718 & 2,325 & 528 & 178 & 1,619 & 0
        & 2,002 & 199 & 124 \\

        \href{https://huggingface.co/datasets/MINT-SJTU/RoboFAC-dataset}
        {RoboFAC (real)}
        & 602 & 596 & 1 & 152 & 443 & 0
        & 535 & 36 & 25 \\

        \href{https://huggingface.co/datasets/zhouzypaul/auto_eval}
        {AutoEval}
        & 12,054 & 10,503 & 4,902 & 776 & 4,825 & 0
        & 9,537 & 415 & 551 \\

        Other
        & 3,306 & 2,508 & 800 & 226 & 1,440 & 42
        & 2,237 & 136 & 135 \\
        \midrule
        Total
        & 99,198 & 72,592 & 17,929 & 7,663 & 46,958 & 42
        & 61,691 & 5,296 & 5,605 \\
        \bottomrule
    \end{tabular}
    \end{adjustbox}

    \par\smallskip
    \begin{minipage}{\linewidth}
    RoboFAC includes only its real-world subset.
    The raw ArmNetBench count includes 600 reference
    demonstrations, which are excluded from Retained.
    Of 73,084 trajectories with usable annotations,
    492 are outside these partitions: 401 ViFailback
    benchmark trajectories and 91 trajectories from Other not
    included under the selected supervision and camera
    settings.
    Outcome categories follow the selected annotations;
    Failure includes executions that remain incomplete
    at the end of recording.
    \end{minipage}
\end{table}

\subsection{Event Annotations and Examples}
\label{app:event-definitions}

\paragraph{Annotation contents.}
For each event, records contain its category, reference state, and temporal
region; regression-and-recovery also supplies a recovery state. These anchors
and supporting observations define justified within-trajectory comparisons
without framewise completion percentages; available descriptions explain
attained states and their changes. Table~\ref{tab:event-definitions} summarizes the
four categories, interpreted against the complete task instruction and
observable states; a failed outcome alone does not imply lost progress.

\begin{table}[t]
    \centering
    \caption{\textbf{Execution events and the progress relations they support.}
        Comparisons apply to observations that support the annotated relation.}
    \label{tab:event-definitions}
    \setlength{\tabcolsep}{4pt}
    \renewcommand{\arraystretch}{1.15}
    \begin{adjustbox}{max width=\linewidth}
\begin{tabular}{p{0.23\linewidth}p{0.71\linewidth}}
        \toprule
        Execution event & Definition and supported progress relation \\
        \midrule
        Progress stall
        & A local interval without observable task advancement; progress remains approximately unchanged relative to the reference state. \\
        Regression and recovery
        & An achieved task state is lost and subsequently restored. Supported states during the loss have lower progress than both the reference and recovery states. \\
        No further progress
        & The remaining execution provides no evidence of advancement beyond the reference state before recording ends. Supported states have progress no greater than the reference. \\
        Unrecovered regression
        & An achieved task state is lost and remains unrecovered when recording ends. Supported post-loss states have lower progress than the reference. \\
        \bottomrule
    \end{tabular}
    \end{adjustbox}
\end{table}

These distinctions depend on task-state evidence rather than motion alone.
Task-required waiting is not automatically a progress stall; an unsuccessful grasp
attempt is not regression unless a grasp had been established; and renewed
motion is not recovery unless the relevant task state is restored. No further
progress does not require constant progress throughout the remaining
execution, and unrecovered regression does not require a frame-by-frame
monotonic decline.

An annotated interval locates an event; only observations supporting the
stated relation are used for comparison. Later advancement or recovery is
considered when interpreting the event, and comparisons with insufficient
evidence are excluded. A locally supported relation can be retained even
when the terminal outcome is uncertain. These relations supervise the
progress model as described in Section~\ref{sec:joint-learning} and
Appendix~\ref{app:event-losses}.

\paragraph{Illustrative annotations in simulation and on real robots.}
The following pages illustrate the four event categories in
LIBERO~\citep{liu2023libero}
(pp.~\pageref{app:libero-event-examples}--\pageref{app:libero-event-examples-end})
and real-robot executions
(pp.~\pageref{app:real-event-examples}--\pageref{app:real-event-examples-end}).
Each domain contains 16 distinct trajectories: four per event category,
with seven frames per trajectory spanning the recording start and endpoint.
In this display, stall and recovery examples end in success, while
no-further-progress and unrecovered-regression examples end in failure.
These outcomes provide context; the supported progress relations depend
on observable changes to task-relevant states rather than terminal
outcome alone, and do not assign completion percentages.

\paragraph{Example sources and annotation provenance.}
The real-robot examples include seven RoboChallenge~\citep{yakefu2025robochallenge}
successes and nine RoboArena~\citep{atreya2025roboarena} executions
(one success and eight failures).
All displayed real-robot examples use the final annotation records produced
by the annotation and review procedure in
Appendix~\ref{app:annotation-procedure}.
The displayed LIBERO success events retain their existing human annotations.
The LIBERO failure events retain machine annotations and are not presented
as human-verified.
These examples illustrate event semantics and do not constitute an
additional annotation-quality evaluation.

\subsection{Annotation Procedure and Quality Control}
\label{app:annotation-procedure}

\paragraph{Data scope and temporal alignment.}
We manually inspected the full collection of approximately 100K raw
real-robot execution records. Records were organized by task instruction,
source outcome, and observation timeline, with available camera views
retained. Source success and failure outcomes were preserved. For
successful rollouts used for progress supervision, we identified the
task-relevant completion endpoint, retaining release and stabilization
needed for completion while excluding unrelated post-task motion or
waiting. RoboChallenge success videos additionally used a
robot-state-derived execution timeline that compresses long static
intervals while retaining context around motion boundaries. Annotations
were aligned with the corresponding observation timeline, with mappings
back to the source recordings.

\paragraph{Machine annotation.}
We used GPT-5.6 Sol with medium reasoning effort to generate initial
annotations (V1) for all real-robot data sources from task instructions and
indexed visual observations in two passes. A coarse pass identified
candidate events, reference states, and temporal regions across the
execution. A refinement pass used denser observations around these regions
and later context when needed to refine event interpretation and boundary
localization, including whether a lost task state was subsequently
recovered. The annotations recorded event types and instances, temporal
anchors, and brief descriptions of the supporting task-state changes.

\paragraph{Independent human annotation and review.}
Each retained trajectory had an independent human annotation (V2), produced
without access to V1. We compared the two versions for event type, temporal
order, instance count, and boundary localization, and checked the
completeness of applicable information. Records with event or boundary
conflicts, or unresolved required information, were referred for
third-party review (V3). Reviewers first annotated independently without
seeing either version, then compared their annotations with the frozen V1
and V2 records. Trajectories selected by stratified sampling from the set
with no detected conflicts and no unresolved required information underwent
independent auditing. Auditors were blinded to the original annotations and
prior adjudication results.

\paragraph{Consolidation and final records.}
Final annotations incorporated confirmations and revisions from review and
independent auditing, while the original machine and human records and
review provenance were retained. Completion of review did not imply that
every required field could be determined: unresolved information remained
explicitly marked, and comparisons lacking sufficient support were excluded
from supervision. Retained trajectories satisfied the eligibility criteria
in Appendix~\ref{app:dataset-composition}. The dataset composition
statistics and real-robot examples in this paper use the final annotation
records.

\paragraph{RoboArena audit example.}
Table~\ref{tab:roboarena-annotation-audit} illustrates the shared procedure
for RoboArena's 8,146 retained trajectories, each independently annotated
by one of ten human annotators. Corresponding V1/V2 event boundaries were
mapped to a common processed observation timeline and normalized within
the effective trajectory interval. Any boundary difference of at least
0.10 triggered third-party review, allowing comparison across frame rates.
Independent audits of the concordant set were stratified by primary
category: all regression-and-recovery and multiple-event success
trajectories were checked, with random samples from the remaining strata.
An audited trajectory required revision if either frozen annotation
needed correction and was counted only once.

\begin{table}[t]
    \centering
    \caption{\textbf{RoboArena annotation audit.}
        Statistics concern the 8,146 retained trajectories.}
    \label{tab:roboarena-annotation-audit}
    \setlength{\tabcolsep}{4pt}
    \renewcommand{\arraystretch}{1.15}
    \begin{adjustbox}{max width=\linewidth}
\begin{tabular}{@{}p{0.32\linewidth}p{0.64\linewidth}@{}}
        \toprule
        Audit item & Result \\
        \midrule
        Primary-category agreement
        & 7,133 / 8,146 (87.56\%) \\
        Third-party review
        & All 1,370 referred trajectories reviewed \\
        Third-party conclusions
        & 1,079 supported V2; 230 supported V1;
          47 required revisions to both; 14 remained unresolved \\
        Independent audit of the concordant set
        & 800 / 6,776 trajectories inspected;
          26 required revision (3.25\% of the sample) \\
        Stratified weighted revision rate
        & 3.24\% \\
        Final annotation completeness
        & 8,132 with complete applicable information;
          14 with unresolved information \\
        \bottomrule
    \end{tabular}
    \end{adjustbox}

    \par\smallskip
    \begin{minipage}{\linewidth}
    Primary-category agreement does not establish agreement on all event
    boundaries or required information. Revision rates concern the frozen
    annotations in the initial concordant set, rather than the final
    dataset's error rate. Unresolved records remain in the respective
    review and completeness denominators.
    \end{minipage}
\end{table}

\subsection{Release Contents and Usage}
\label{app:dataset-release}

We will release event annotations, data splits, source-media references,
and documented tools for loading records and constructing supervision.
Records link task instructions, execution outcomes, event regions, and
reference and recovery states, where applicable, to their source
trajectories. They retain annotation provenance and human-review status.
Accompanying semantic descriptions support interpretation and retrieval;
they are not an additional training objective. Event definitions and
training objectives are specified in Appendices~\ref{app:event-definitions}
and~\ref{app:method-training}, respectively.

\clearpage
\begingroup
\raggedbottom
\setlength{\parindent}{0pt}
\normalsize
\edef\eventExampleParskip{\the\parskip}
\clearpage
\begingroup
\definecolor{LiberoSuccess}{HTML}{15803D}\definecolor{LiberoFailure}{HTML}{B91C1C}
\begingroup
\setlength{\parskip}{\eventExampleParskip}
\section*{LIBERO: Progress stall}
\label{app:libero-event-examples}
{\normalsize Progress stays approximately unchanged over $[a,r)$. Black borders mark the stall onset $a$ and resumption $r$. All four videos finish successfully; each row shows F0, the stall, later progress, and the recording endpoint $T$.\par}
\begin{minipage}{\linewidth}
\setlength{\parskip}{\eventExampleParskip}
{\normalsize \textbf{(a)} Put the yellow-and-white mug in the microwave and close it. The mug waits on the table, then is put inside and the door closes. \mbox{\textbf{Stall: 21.6\,s}}\par}
\noindent \begin{minipage}[t]{0.137\linewidth}\centering
\setlength{\fboxsep}{0pt}\setlength{\fboxrule}{0.8pt}
\fcolorbox{white}{white}{\includegraphics[width=\dimexpr\linewidth-1.6pt\relax]{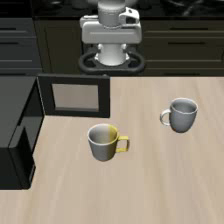}}\\[-2pt]
{\scriptsize F0 / 0.0s\strut}\\[-1pt]
{\scriptsize Context\strut}\end{minipage}\hfill%
\begin{minipage}[t]{0.137\linewidth}\centering
\setlength{\fboxsep}{0pt}\setlength{\fboxrule}{0.8pt}
\fcolorbox{black}{white}{\includegraphics[width=\dimexpr\linewidth-1.6pt\relax]{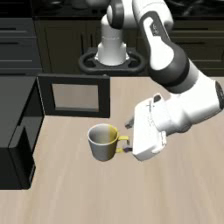}}\\[-2pt]
{\scriptsize \textbf{F96 / 9.6s}\strut}\\[-1pt]
{\scriptsize \textbf{Start $a$}\strut}\end{minipage}\hfill%
\begin{minipage}[t]{0.137\linewidth}\centering
\setlength{\fboxsep}{0pt}\setlength{\fboxrule}{0.8pt}
\fcolorbox{white}{white}{\includegraphics[width=\dimexpr\linewidth-1.6pt\relax]{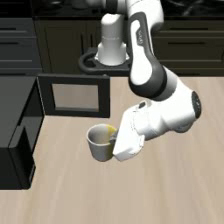}}\\[-2pt]
{\scriptsize F168 / 16.8s\strut}\\[-1pt]
{\scriptsize Stall\strut}\end{minipage}\hfill%
\begin{minipage}[t]{0.137\linewidth}\centering
\setlength{\fboxsep}{0pt}\setlength{\fboxrule}{0.8pt}
\fcolorbox{white}{white}{\includegraphics[width=\dimexpr\linewidth-1.6pt\relax]{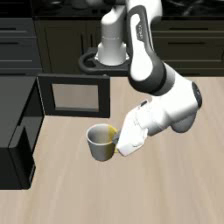}}\\[-2pt]
{\scriptsize F270 / 27.0s\strut}\\[-1pt]
{\scriptsize Stall\strut}\end{minipage}\hfill%
\begin{minipage}[t]{0.137\linewidth}\centering
\setlength{\fboxsep}{0pt}\setlength{\fboxrule}{0.8pt}
\fcolorbox{black}{white}{\includegraphics[width=\dimexpr\linewidth-1.6pt\relax]{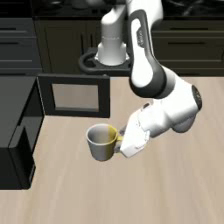}}\\[-2pt]
{\scriptsize \textbf{F312 / 31.2s}\strut}\\[-1pt]
{\scriptsize \textbf{Resume $r$}\strut}\end{minipage}\hfill%
\begin{minipage}[t]{0.137\linewidth}\centering
\setlength{\fboxsep}{0pt}\setlength{\fboxrule}{0.8pt}
\fcolorbox{white}{white}{\includegraphics[width=\dimexpr\linewidth-1.6pt\relax]{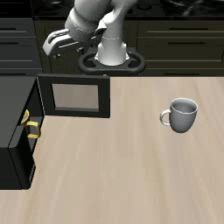}}\\[-2pt]
{\scriptsize F437 / 43.7s\strut}\\[-1pt]
{\scriptsize Later progress\strut}\end{minipage}\hfill%
\begin{minipage}[t]{0.137\linewidth}\centering
\setlength{\fboxsep}{0pt}\setlength{\fboxrule}{0.8pt}
\fcolorbox{LiberoSuccess}{white}{\includegraphics[width=\dimexpr\linewidth-1.6pt\relax]{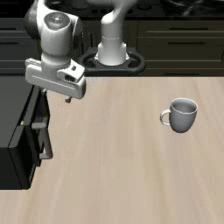}}\\[-2pt]
{\scriptsize F477 / 47.7s\strut}\\[-1pt]
{\scriptsize End $T$\strut}\\[-1pt]{\scriptsize\color{LiberoSuccess}\textbf{Success}\strut}\end{minipage}
\end{minipage}\par\vspace{10pt}
\begin{minipage}{\linewidth}
\setlength{\parskip}{\eventExampleParskip}
{\normalsize \textbf{(b)} Put the black bowl in the bottom drawer and close it. After the pause, the bowl is placed inside and the drawer closes. \mbox{\textbf{Stall: 8.8\,s}}\par}
\noindent \begin{minipage}[t]{0.137\linewidth}\centering
\setlength{\fboxsep}{0pt}\setlength{\fboxrule}{0.8pt}
\fcolorbox{white}{white}{\includegraphics[width=\dimexpr\linewidth-1.6pt\relax]{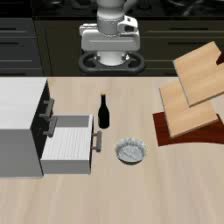}}\\[-2pt]
{\scriptsize F0 / 0.0s\strut}\\[-1pt]
{\scriptsize Context\strut}\end{minipage}\hfill%
\begin{minipage}[t]{0.137\linewidth}\centering
\setlength{\fboxsep}{0pt}\setlength{\fboxrule}{0.8pt}
\fcolorbox{black}{white}{\includegraphics[width=\dimexpr\linewidth-1.6pt\relax]{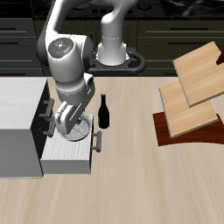}}\\[-2pt]
{\scriptsize \textbf{F120 / 12.0s}\strut}\\[-1pt]
{\scriptsize \textbf{Start $a$}\strut}\end{minipage}\hfill%
\begin{minipage}[t]{0.137\linewidth}\centering
\setlength{\fboxsep}{0pt}\setlength{\fboxrule}{0.8pt}
\fcolorbox{white}{white}{\includegraphics[width=\dimexpr\linewidth-1.6pt\relax]{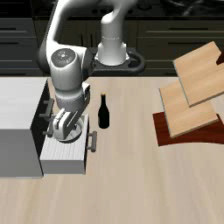}}\\[-2pt]
{\scriptsize F149 / 14.9s\strut}\\[-1pt]
{\scriptsize Stall\strut}\end{minipage}\hfill%
\begin{minipage}[t]{0.137\linewidth}\centering
\setlength{\fboxsep}{0pt}\setlength{\fboxrule}{0.8pt}
\fcolorbox{white}{white}{\includegraphics[width=\dimexpr\linewidth-1.6pt\relax]{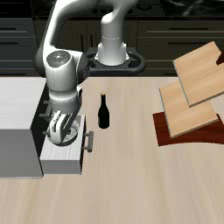}}\\[-2pt]
{\scriptsize F190 / 19.0s\strut}\\[-1pt]
{\scriptsize Stall\strut}\end{minipage}\hfill%
\begin{minipage}[t]{0.137\linewidth}\centering
\setlength{\fboxsep}{0pt}\setlength{\fboxrule}{0.8pt}
\fcolorbox{black}{white}{\includegraphics[width=\dimexpr\linewidth-1.6pt\relax]{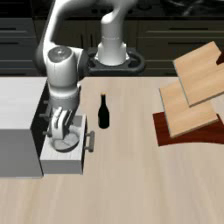}}\\[-2pt]
{\scriptsize \textbf{F208 / 20.8s}\strut}\\[-1pt]
{\scriptsize \textbf{Resume $r$}\strut}\end{minipage}\hfill%
\begin{minipage}[t]{0.137\linewidth}\centering
\setlength{\fboxsep}{0pt}\setlength{\fboxrule}{0.8pt}
\fcolorbox{white}{white}{\includegraphics[width=\dimexpr\linewidth-1.6pt\relax]{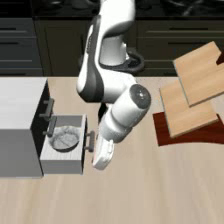}}\\[-2pt]
{\scriptsize F249 / 24.9s\strut}\\[-1pt]
{\scriptsize Later progress\strut}\end{minipage}\hfill%
\begin{minipage}[t]{0.137\linewidth}\centering
\setlength{\fboxsep}{0pt}\setlength{\fboxrule}{0.8pt}
\fcolorbox{LiberoSuccess}{white}{\includegraphics[width=\dimexpr\linewidth-1.6pt\relax]{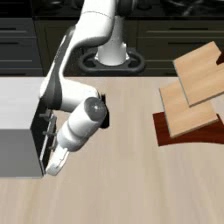}}\\[-2pt]
{\scriptsize F269 / 26.9s\strut}\\[-1pt]
{\scriptsize End $T$\strut}\\[-1pt]{\scriptsize\color{LiberoSuccess}\textbf{Success}\strut}\end{minipage}
\end{minipage}\par\vspace{10pt}
\begin{minipage}{\linewidth}
\setlength{\parskip}{\eventExampleParskip}
{\normalsize \textbf{(c)} Turn on the stove and put the moka pot on it. The pot stays on the table during the pause, then reaches the lit stove. \mbox{\textbf{Stall: 9.3\,s}}\par}
\noindent \begin{minipage}[t]{0.137\linewidth}\centering
\setlength{\fboxsep}{0pt}\setlength{\fboxrule}{0.8pt}
\fcolorbox{white}{white}{\includegraphics[width=\dimexpr\linewidth-1.6pt\relax]{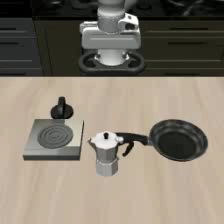}}\\[-2pt]
{\scriptsize F0 / 0.0s\strut}\\[-1pt]
{\scriptsize Context\strut}\end{minipage}\hfill%
\begin{minipage}[t]{0.137\linewidth}\centering
\setlength{\fboxsep}{0pt}\setlength{\fboxrule}{0.8pt}
\fcolorbox{black}{white}{\includegraphics[width=\dimexpr\linewidth-1.6pt\relax]{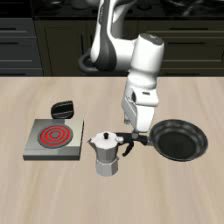}}\\[-2pt]
{\scriptsize \textbf{F170 / 17.0s}\strut}\\[-1pt]
{\scriptsize \textbf{Start $a$}\strut}\end{minipage}\hfill%
\begin{minipage}[t]{0.137\linewidth}\centering
\setlength{\fboxsep}{0pt}\setlength{\fboxrule}{0.8pt}
\fcolorbox{white}{white}{\includegraphics[width=\dimexpr\linewidth-1.6pt\relax]{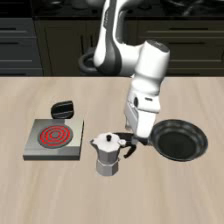}}\\[-2pt]
{\scriptsize F201 / 20.1s\strut}\\[-1pt]
{\scriptsize Stall\strut}\end{minipage}\hfill%
\begin{minipage}[t]{0.137\linewidth}\centering
\setlength{\fboxsep}{0pt}\setlength{\fboxrule}{0.8pt}
\fcolorbox{white}{white}{\includegraphics[width=\dimexpr\linewidth-1.6pt\relax]{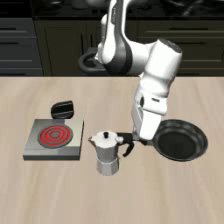}}\\[-2pt]
{\scriptsize F244 / 24.4s\strut}\\[-1pt]
{\scriptsize Stall\strut}\end{minipage}\hfill%
\begin{minipage}[t]{0.137\linewidth}\centering
\setlength{\fboxsep}{0pt}\setlength{\fboxrule}{0.8pt}
\fcolorbox{black}{white}{\includegraphics[width=\dimexpr\linewidth-1.6pt\relax]{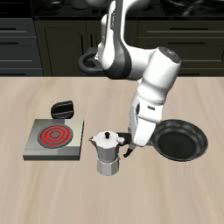}}\\[-2pt]
{\scriptsize \textbf{F263 / 26.3s}\strut}\\[-1pt]
{\scriptsize \textbf{Resume $r$}\strut}\end{minipage}\hfill%
\begin{minipage}[t]{0.137\linewidth}\centering
\setlength{\fboxsep}{0pt}\setlength{\fboxrule}{0.8pt}
\fcolorbox{white}{white}{\includegraphics[width=\dimexpr\linewidth-1.6pt\relax]{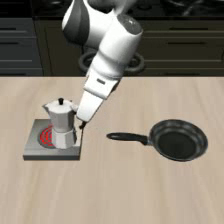}}\\[-2pt]
{\scriptsize F306 / 30.6s\strut}\\[-1pt]
{\scriptsize Later progress\strut}\end{minipage}\hfill%
\begin{minipage}[t]{0.137\linewidth}\centering
\setlength{\fboxsep}{0pt}\setlength{\fboxrule}{0.8pt}
\fcolorbox{LiberoSuccess}{white}{\includegraphics[width=\dimexpr\linewidth-1.6pt\relax]{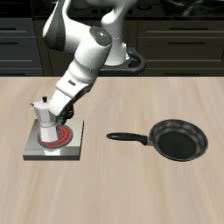}}\\[-2pt]
{\scriptsize F326 / 32.6s\strut}\\[-1pt]
{\scriptsize End $T$\strut}\\[-1pt]{\scriptsize\color{LiberoSuccess}\textbf{Success}\strut}\end{minipage}
\end{minipage}\par\vspace{10pt}
\begin{minipage}{\linewidth}
\setlength{\parskip}{\eventExampleParskip}
{\normalsize \textbf{(d)} Put both moka pots on the stove. The second pot stays on the table during the stall; both pots reach the stove afterward. \mbox{\textbf{Stall: 15.0\,s}}\par}
\noindent \begin{minipage}[t]{0.137\linewidth}\centering
\setlength{\fboxsep}{0pt}\setlength{\fboxrule}{0.8pt}
\fcolorbox{white}{white}{\includegraphics[width=\dimexpr\linewidth-1.6pt\relax]{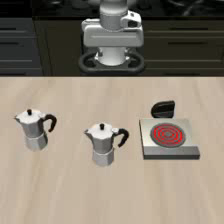}}\\[-2pt]
{\scriptsize F0 / 0.0s\strut}\\[-1pt]
{\scriptsize Context\strut}\end{minipage}\hfill%
\begin{minipage}[t]{0.137\linewidth}\centering
\setlength{\fboxsep}{0pt}\setlength{\fboxrule}{0.8pt}
\fcolorbox{black}{white}{\includegraphics[width=\dimexpr\linewidth-1.6pt\relax]{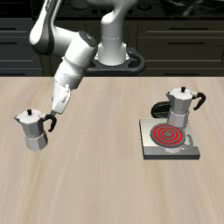}}\\[-2pt]
{\scriptsize \textbf{F270 / 27.0s}\strut}\\[-1pt]
{\scriptsize \textbf{Start $a$}\strut}\end{minipage}\hfill%
\begin{minipage}[t]{0.137\linewidth}\centering
\setlength{\fboxsep}{0pt}\setlength{\fboxrule}{0.8pt}
\fcolorbox{white}{white}{\includegraphics[width=\dimexpr\linewidth-1.6pt\relax]{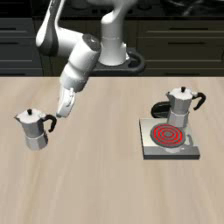}}\\[-2pt]
{\scriptsize F320 / 32.0s\strut}\\[-1pt]
{\scriptsize Stall\strut}\end{minipage}\hfill%
\begin{minipage}[t]{0.137\linewidth}\centering
\setlength{\fboxsep}{0pt}\setlength{\fboxrule}{0.8pt}
\fcolorbox{white}{white}{\includegraphics[width=\dimexpr\linewidth-1.6pt\relax]{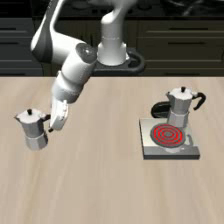}}\\[-2pt]
{\scriptsize F390 / 39.0s\strut}\\[-1pt]
{\scriptsize Stall\strut}\end{minipage}\hfill%
\begin{minipage}[t]{0.137\linewidth}\centering
\setlength{\fboxsep}{0pt}\setlength{\fboxrule}{0.8pt}
\fcolorbox{black}{white}{\includegraphics[width=\dimexpr\linewidth-1.6pt\relax]{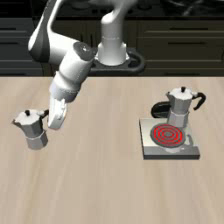}}\\[-2pt]
{\scriptsize \textbf{F420 / 42.0s}\strut}\\[-1pt]
{\scriptsize \textbf{Resume $r$}\strut}\end{minipage}\hfill%
\begin{minipage}[t]{0.137\linewidth}\centering
\setlength{\fboxsep}{0pt}\setlength{\fboxrule}{0.8pt}
\fcolorbox{white}{white}{\includegraphics[width=\dimexpr\linewidth-1.6pt\relax]{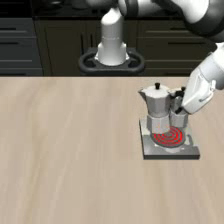}}\\[-2pt]
{\scriptsize F485 / 48.5s\strut}\\[-1pt]
{\scriptsize Later progress\strut}\end{minipage}\hfill%
\begin{minipage}[t]{0.137\linewidth}\centering
\setlength{\fboxsep}{0pt}\setlength{\fboxrule}{0.8pt}
\fcolorbox{LiberoSuccess}{white}{\includegraphics[width=\dimexpr\linewidth-1.6pt\relax]{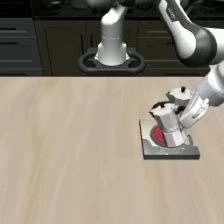}}\\[-2pt]
{\scriptsize F505 / 50.5s\strut}\\[-1pt]
{\scriptsize End $T$\strut}\\[-1pt]{\scriptsize\color{LiberoSuccess}\textbf{Success}\strut}\end{minipage}
\end{minipage}\par\vspace{10pt}
{\scriptsize Source: LIBERO; existing human annotations. Frames span F0 to the recording endpoint $T$ (10 fps). W: wrist-camera view. Black borders mark annotated event anchors; green/red indicates success/failure.\par}
\endgroup
\clearpage

\begingroup
\setlength{\parskip}{\eventExampleParskip}
\section*{LIBERO: Regression and recovery}
{\normalsize A previously attained task state is lost and restored. Black borders mark the reference $a$ and recovery $r$; intermediate frames show the loss and regrasping. Each row begins at F0 and ends at the successful recording endpoint $T$.\par}
\begin{minipage}{\linewidth}
\setlength{\parskip}{\eventExampleParskip}
{\normalsize \textbf{(a)} Put the alphabet-soup can and cream-cheese box in the basket. The can drops and is regrasped; both targets end in the basket.\par}
\noindent \begin{minipage}[t]{0.137\linewidth}\centering
\setlength{\fboxsep}{0pt}\setlength{\fboxrule}{0.8pt}
\fcolorbox{white}{white}{\includegraphics[width=\dimexpr\linewidth-1.6pt\relax]{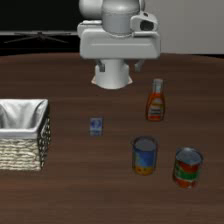}}\\[-2pt]
{\scriptsize F0 / 0.0s\strut}\\[-1pt]
{\scriptsize Context\strut}\end{minipage}\hfill%
\begin{minipage}[t]{0.137\linewidth}\centering
\setlength{\fboxsep}{0pt}\setlength{\fboxrule}{0.8pt}
\fcolorbox{black}{white}{\includegraphics[width=\dimexpr\linewidth-1.6pt\relax]{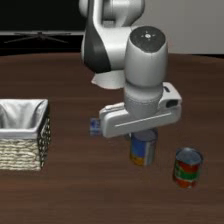}}\\[-2pt]
{\scriptsize \textbf{F71 / 7.1s}\strut}\\[-1pt]
{\scriptsize \textbf{Ref. $a$}\strut}\end{minipage}\hfill%
\begin{minipage}[t]{0.137\linewidth}\centering
\setlength{\fboxsep}{0pt}\setlength{\fboxrule}{0.8pt}
\fcolorbox{white}{white}{\includegraphics[width=\dimexpr\linewidth-1.6pt\relax]{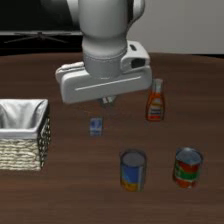}}\\[-2pt]
{\scriptsize F81 / 8.1s\strut}\\[-1pt]
{\scriptsize Loss evidence\strut}\end{minipage}\hfill%
\begin{minipage}[t]{0.137\linewidth}\centering
\setlength{\fboxsep}{0pt}\setlength{\fboxrule}{0.8pt}
\fcolorbox{white}{white}{\includegraphics[width=\dimexpr\linewidth-1.6pt\relax]{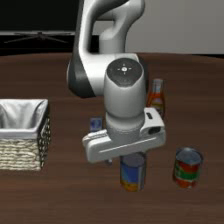}}\\[-2pt]
{\scriptsize F176 / 17.6s\strut}\\[-1pt]
{\scriptsize Regrasp attempt\strut}\end{minipage}\hfill%
\begin{minipage}[t]{0.137\linewidth}\centering
\setlength{\fboxsep}{0pt}\setlength{\fboxrule}{0.8pt}
\fcolorbox{black}{white}{\includegraphics[width=\dimexpr\linewidth-1.6pt\relax]{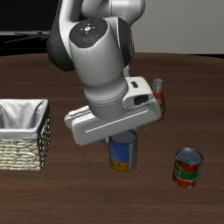}}\\[-2pt]
{\scriptsize \textbf{F282 / 28.2s}\strut}\\[-1pt]
{\scriptsize \textbf{Recovery $r$}\strut}\end{minipage}\hfill%
\begin{minipage}[t]{0.137\linewidth}\centering
\setlength{\fboxsep}{0pt}\setlength{\fboxrule}{0.8pt}
\fcolorbox{white}{white}{\includegraphics[width=\dimexpr\linewidth-1.6pt\relax]{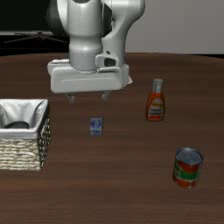}}\\[-2pt]
{\scriptsize F362 / 36.2s\strut}\\[-1pt]
{\scriptsize Later progress\strut}\end{minipage}\hfill%
\begin{minipage}[t]{0.137\linewidth}\centering
\setlength{\fboxsep}{0pt}\setlength{\fboxrule}{0.8pt}
\fcolorbox{LiberoSuccess}{white}{\includegraphics[width=\dimexpr\linewidth-1.6pt\relax]{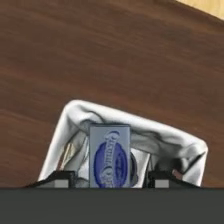}}\\[-2pt]
{\scriptsize F442 / 44.2s W\strut}\\[-1pt]
{\scriptsize End $T$\strut}\\[-1pt]{\scriptsize\color{LiberoSuccess}\textbf{Success}\strut}\end{minipage}
\end{minipage}\par\vspace{10pt}
\begin{minipage}{\linewidth}
\setlength{\parskip}{\eventExampleParskip}
{\normalsize \textbf{(b)} Put the soup and tomato-sauce cans in the basket. The soup can is lost and regrasped; both cans reach the basket.\par}
\noindent \begin{minipage}[t]{0.137\linewidth}\centering
\setlength{\fboxsep}{0pt}\setlength{\fboxrule}{0.8pt}
\fcolorbox{white}{white}{\includegraphics[width=\dimexpr\linewidth-1.6pt\relax]{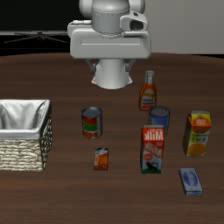}}\\[-2pt]
{\scriptsize F0 / 0.0s\strut}\\[-1pt]
{\scriptsize Context\strut}\end{minipage}\hfill%
\begin{minipage}[t]{0.137\linewidth}\centering
\setlength{\fboxsep}{0pt}\setlength{\fboxrule}{0.8pt}
\fcolorbox{black}{white}{\includegraphics[width=\dimexpr\linewidth-1.6pt\relax]{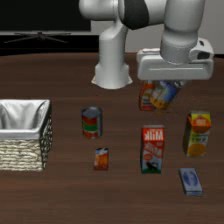}}\\[-2pt]
{\scriptsize \textbf{F106 / 10.6s}\strut}\\[-1pt]
{\scriptsize \textbf{Ref. $a$}\strut}\end{minipage}\hfill%
\begin{minipage}[t]{0.137\linewidth}\centering
\setlength{\fboxsep}{0pt}\setlength{\fboxrule}{0.8pt}
\fcolorbox{white}{white}{\includegraphics[width=\dimexpr\linewidth-1.6pt\relax]{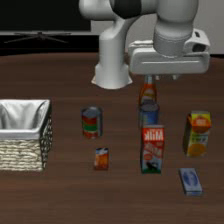}}\\[-2pt]
{\scriptsize F116 / 11.6s\strut}\\[-1pt]
{\scriptsize Loss evidence\strut}\end{minipage}\hfill%
\begin{minipage}[t]{0.137\linewidth}\centering
\setlength{\fboxsep}{0pt}\setlength{\fboxrule}{0.8pt}
\fcolorbox{white}{white}{\includegraphics[width=\dimexpr\linewidth-1.6pt\relax]{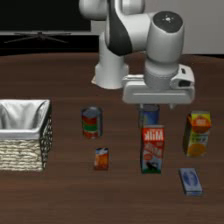}}\\[-2pt]
{\scriptsize F135 / 13.5s\strut}\\[-1pt]
{\scriptsize Regrasp attempt\strut}\end{minipage}\hfill%
\begin{minipage}[t]{0.137\linewidth}\centering
\setlength{\fboxsep}{0pt}\setlength{\fboxrule}{0.8pt}
\fcolorbox{black}{white}{\includegraphics[width=\dimexpr\linewidth-1.6pt\relax]{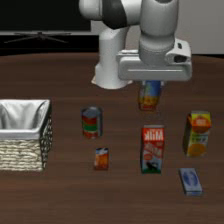}}\\[-2pt]
{\scriptsize \textbf{F164 / 16.4s}\strut}\\[-1pt]
{\scriptsize \textbf{Recovery $r$}\strut}\end{minipage}\hfill%
\begin{minipage}[t]{0.137\linewidth}\centering
\setlength{\fboxsep}{0pt}\setlength{\fboxrule}{0.8pt}
\fcolorbox{white}{white}{\includegraphics[width=\dimexpr\linewidth-1.6pt\relax]{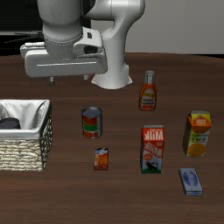}}\\[-2pt]
{\scriptsize F249 / 24.9s\strut}\\[-1pt]
{\scriptsize Later progress\strut}\end{minipage}\hfill%
\begin{minipage}[t]{0.137\linewidth}\centering
\setlength{\fboxsep}{0pt}\setlength{\fboxrule}{0.8pt}
\fcolorbox{LiberoSuccess}{white}{\includegraphics[width=\dimexpr\linewidth-1.6pt\relax]{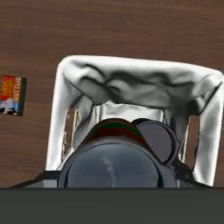}}\\[-2pt]
{\scriptsize F334 / 33.4s W\strut}\\[-1pt]
{\scriptsize End $T$\strut}\\[-1pt]{\scriptsize\color{LiberoSuccess}\textbf{Success}\strut}\end{minipage}
\end{minipage}\par\vspace{10pt}
\begin{minipage}{\linewidth}
\setlength{\parskip}{\eventExampleParskip}
{\normalsize \textbf{(c)} Put the cream-cheese box and butter in the basket. The butter slips during lifting, is grasped again, and joins the cheese box in the basket.\par}
\noindent \begin{minipage}[t]{0.137\linewidth}\centering
\setlength{\fboxsep}{0pt}\setlength{\fboxrule}{0.8pt}
\fcolorbox{white}{white}{\includegraphics[width=\dimexpr\linewidth-1.6pt\relax]{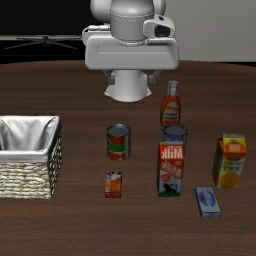}}\\[-2pt]
{\scriptsize F0 / 0.0s\strut}\\[-1pt]
{\scriptsize Context\strut}\end{minipage}\hfill%
\begin{minipage}[t]{0.137\linewidth}\centering
\setlength{\fboxsep}{0pt}\setlength{\fboxrule}{0.8pt}
\fcolorbox{black}{white}{\includegraphics[width=\dimexpr\linewidth-1.6pt\relax]{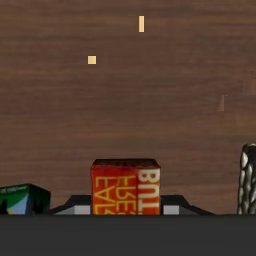}}\\[-2pt]
{\scriptsize \textbf{F223 / 22.3s W}\strut}\\[-1pt]
{\scriptsize \textbf{Ref. $a$}\strut}\end{minipage}\hfill%
\begin{minipage}[t]{0.137\linewidth}\centering
\setlength{\fboxsep}{0pt}\setlength{\fboxrule}{0.8pt}
\fcolorbox{white}{white}{\includegraphics[width=\dimexpr\linewidth-1.6pt\relax]{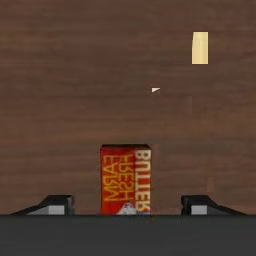}}\\[-2pt]
{\scriptsize F233 / 23.3s W\strut}\\[-1pt]
{\scriptsize Loss evidence\strut}\end{minipage}\hfill%
\begin{minipage}[t]{0.137\linewidth}\centering
\setlength{\fboxsep}{0pt}\setlength{\fboxrule}{0.8pt}
\fcolorbox{white}{white}{\includegraphics[width=\dimexpr\linewidth-1.6pt\relax]{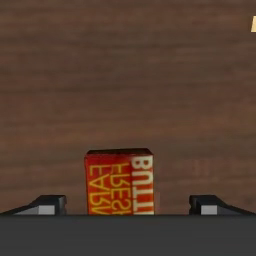}}\\[-2pt]
{\scriptsize F243 / 24.3s W\strut}\\[-1pt]
{\scriptsize Regrasp attempt\strut}\end{minipage}\hfill%
\begin{minipage}[t]{0.137\linewidth}\centering
\setlength{\fboxsep}{0pt}\setlength{\fboxrule}{0.8pt}
\fcolorbox{black}{white}{\includegraphics[width=\dimexpr\linewidth-1.6pt\relax]{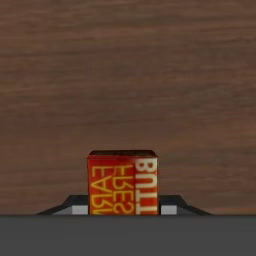}}\\[-2pt]
{\scriptsize \textbf{F264 / 26.4s W}\strut}\\[-1pt]
{\scriptsize \textbf{Recovery $r$}\strut}\end{minipage}\hfill%
\begin{minipage}[t]{0.137\linewidth}\centering
\setlength{\fboxsep}{0pt}\setlength{\fboxrule}{0.8pt}
\fcolorbox{white}{white}{\includegraphics[width=\dimexpr\linewidth-1.6pt\relax]{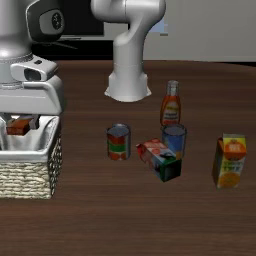}}\\[-2pt]
{\scriptsize F300 / 30.0s\strut}\\[-1pt]
{\scriptsize Later progress\strut}\end{minipage}\hfill%
\begin{minipage}[t]{0.137\linewidth}\centering
\setlength{\fboxsep}{0pt}\setlength{\fboxrule}{0.8pt}
\fcolorbox{LiberoSuccess}{white}{\includegraphics[width=\dimexpr\linewidth-1.6pt\relax]{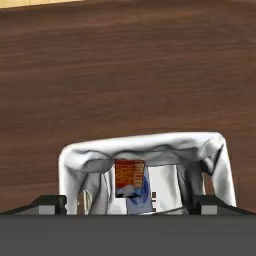}}\\[-2pt]
{\scriptsize F320 / 32.0s W\strut}\\[-1pt]
{\scriptsize End $T$\strut}\\[-1pt]{\scriptsize\color{LiberoSuccess}\textbf{Success}\strut}\end{minipage}
\end{minipage}\par\vspace{10pt}
\begin{minipage}{\linewidth}
\setlength{\parskip}{\eventExampleParskip}
{\normalsize \textbf{(d)} Put the soup and tomato-sauce cans in the basket. The tomato-sauce can drops, is lifted again, and moves into the basket.\par}
\noindent \begin{minipage}[t]{0.137\linewidth}\centering
\setlength{\fboxsep}{0pt}\setlength{\fboxrule}{0.8pt}
\fcolorbox{white}{white}{\includegraphics[width=\dimexpr\linewidth-1.6pt\relax]{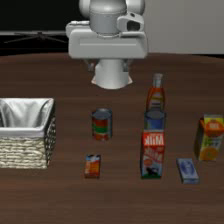}}\\[-2pt]
{\scriptsize F0 / 0.0s\strut}\\[-1pt]
{\scriptsize Context\strut}\end{minipage}\hfill%
\begin{minipage}[t]{0.137\linewidth}\centering
\setlength{\fboxsep}{0pt}\setlength{\fboxrule}{0.8pt}
\fcolorbox{black}{white}{\includegraphics[width=\dimexpr\linewidth-1.6pt\relax]{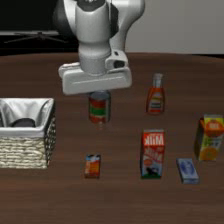}}\\[-2pt]
{\scriptsize \textbf{F213 / 21.3s}\strut}\\[-1pt]
{\scriptsize \textbf{Ref. $a$}\strut}\end{minipage}\hfill%
\begin{minipage}[t]{0.137\linewidth}\centering
\setlength{\fboxsep}{0pt}\setlength{\fboxrule}{0.8pt}
\fcolorbox{white}{white}{\includegraphics[width=\dimexpr\linewidth-1.6pt\relax]{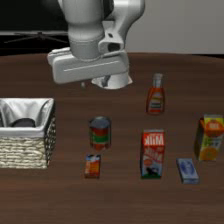}}\\[-2pt]
{\scriptsize F223 / 22.3s\strut}\\[-1pt]
{\scriptsize Loss evidence\strut}\end{minipage}\hfill%
\begin{minipage}[t]{0.137\linewidth}\centering
\setlength{\fboxsep}{0pt}\setlength{\fboxrule}{0.8pt}
\fcolorbox{white}{white}{\includegraphics[width=\dimexpr\linewidth-1.6pt\relax]{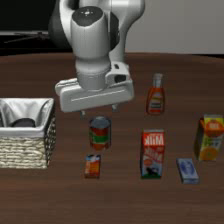}}\\[-2pt]
{\scriptsize F240 / 24.0s\strut}\\[-1pt]
{\scriptsize Regrasp attempt\strut}\end{minipage}\hfill%
\begin{minipage}[t]{0.137\linewidth}\centering
\setlength{\fboxsep}{0pt}\setlength{\fboxrule}{0.8pt}
\fcolorbox{black}{white}{\includegraphics[width=\dimexpr\linewidth-1.6pt\relax]{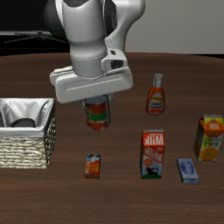}}\\[-2pt]
{\scriptsize \textbf{F266 / 26.6s}\strut}\\[-1pt]
{\scriptsize \textbf{Recovery $r$}\strut}\end{minipage}\hfill%
\begin{minipage}[t]{0.137\linewidth}\centering
\setlength{\fboxsep}{0pt}\setlength{\fboxrule}{0.8pt}
\fcolorbox{white}{white}{\includegraphics[width=\dimexpr\linewidth-1.6pt\relax]{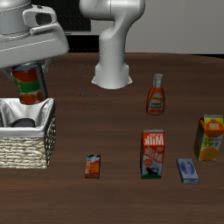}}\\[-2pt]
{\scriptsize F286 / 28.6s\strut}\\[-1pt]
{\scriptsize Later progress\strut}\end{minipage}\hfill%
\begin{minipage}[t]{0.137\linewidth}\centering
\setlength{\fboxsep}{0pt}\setlength{\fboxrule}{0.8pt}
\fcolorbox{LiberoSuccess}{white}{\includegraphics[width=\dimexpr\linewidth-1.6pt\relax]{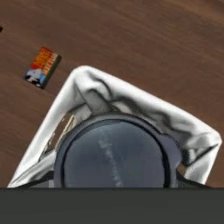}}\\[-2pt]
{\scriptsize F307 / 30.7s W\strut}\\[-1pt]
{\scriptsize End $T$\strut}\\[-1pt]{\scriptsize\color{LiberoSuccess}\textbf{Success}\strut}\end{minipage}
\end{minipage}\par\vspace{10pt}
{\scriptsize Source: LIBERO; existing human annotations. Frames span F0 to the recording endpoint $T$ (10 fps). W: wrist-camera view. Black borders mark annotated event anchors; green/red indicates success/failure.\par}
\endgroup
\clearpage

\begingroup
\setlength{\parskip}{\eventExampleParskip}
\section*{LIBERO: No further progress}
{\normalsize Black borders mark the annotated reference $a$ after the last observed task advancement. Earlier frames provide context; later states do not advance beyond $p_a$. Each row begins at F0 and ends at the failed recording endpoint $T$.\par}
\begin{minipage}{\linewidth}
\setlength{\parskip}{\eventExampleParskip}
{\normalsize \textbf{(a)} Put the bowl in the bottom drawer and close it. The bowl enters the drawer, but closing stops partway.\par}
\noindent \begin{minipage}[t]{0.137\linewidth}\centering
\setlength{\fboxsep}{0pt}\setlength{\fboxrule}{0.8pt}
\fcolorbox{white}{white}{\includegraphics[width=\dimexpr\linewidth-1.6pt\relax]{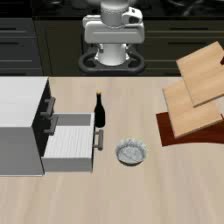}}\\[-2pt]
{\scriptsize F0 / 0.0s\strut}\\[-1pt]
{\scriptsize Context\strut}\end{minipage}\hfill%
\begin{minipage}[t]{0.137\linewidth}\centering
\setlength{\fboxsep}{0pt}\setlength{\fboxrule}{0.8pt}
\fcolorbox{white}{white}{\includegraphics[width=\dimexpr\linewidth-1.6pt\relax]{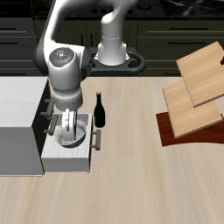}}\\[-2pt]
{\scriptsize F207 / 20.7s\strut}\\[-1pt]
{\scriptsize Earlier progress\strut}\end{minipage}\hfill%
\begin{minipage}[t]{0.137\linewidth}\centering
\setlength{\fboxsep}{0pt}\setlength{\fboxrule}{0.8pt}
\fcolorbox{black}{white}{\includegraphics[width=\dimexpr\linewidth-1.6pt\relax]{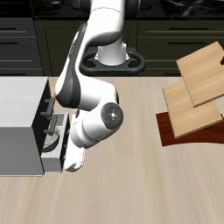}}\\[-2pt]
{\scriptsize \textbf{F267 / 26.7s}\strut}\\[-1pt]
{\scriptsize \textbf{Ref. $a$}\strut}\end{minipage}\hfill%
\begin{minipage}[t]{0.137\linewidth}\centering
\setlength{\fboxsep}{0pt}\setlength{\fboxrule}{0.8pt}
\fcolorbox{white}{white}{\includegraphics[width=\dimexpr\linewidth-1.6pt\relax]{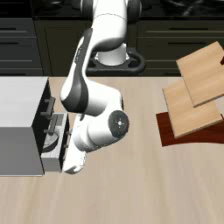}}\\[-2pt]
{\scriptsize F330 / 33.0s\strut}\\[-1pt]
{\scriptsize No advancement\strut}\end{minipage}\hfill%
\begin{minipage}[t]{0.137\linewidth}\centering
\setlength{\fboxsep}{0pt}\setlength{\fboxrule}{0.8pt}
\fcolorbox{white}{white}{\includegraphics[width=\dimexpr\linewidth-1.6pt\relax]{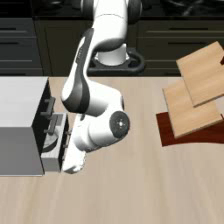}}\\[-2pt]
{\scriptsize F393 / 39.3s\strut}\\[-1pt]
{\scriptsize No advancement\strut}\end{minipage}\hfill%
\begin{minipage}[t]{0.137\linewidth}\centering
\setlength{\fboxsep}{0pt}\setlength{\fboxrule}{0.8pt}
\fcolorbox{white}{white}{\includegraphics[width=\dimexpr\linewidth-1.6pt\relax]{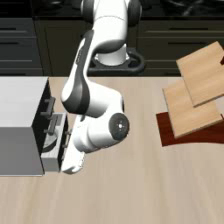}}\\[-2pt]
{\scriptsize F456 / 45.6s\strut}\\[-1pt]
{\scriptsize No advancement\strut}\end{minipage}\hfill%
\begin{minipage}[t]{0.137\linewidth}\centering
\setlength{\fboxsep}{0pt}\setlength{\fboxrule}{0.8pt}
\fcolorbox{LiberoFailure}{white}{\includegraphics[width=\dimexpr\linewidth-1.6pt\relax]{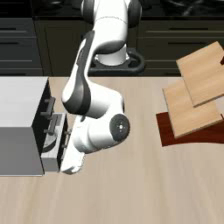}}\\[-2pt]
{\scriptsize F519 / 51.9s\strut}\\[-1pt]
{\scriptsize End $T$\strut}\\[-1pt]{\scriptsize\color{LiberoFailure}\textbf{Failure}\strut}\end{minipage}
\end{minipage}\par\vspace{10pt}
\begin{minipage}{\linewidth}
\setlength{\parskip}{\eventExampleParskip}
{\normalsize \textbf{(b)} Put the bowl in the bottom drawer and close it. A short closing motion ends at the boundary; the drawer stays ajar.\par}
\noindent \begin{minipage}[t]{0.137\linewidth}\centering
\setlength{\fboxsep}{0pt}\setlength{\fboxrule}{0.8pt}
\fcolorbox{white}{white}{\includegraphics[width=\dimexpr\linewidth-1.6pt\relax]{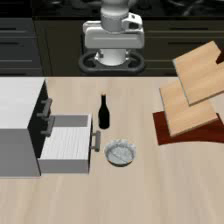}}\\[-2pt]
{\scriptsize F0 / 0.0s\strut}\\[-1pt]
{\scriptsize Context\strut}\end{minipage}\hfill%
\begin{minipage}[t]{0.137\linewidth}\centering
\setlength{\fboxsep}{0pt}\setlength{\fboxrule}{0.8pt}
\fcolorbox{white}{white}{\includegraphics[width=\dimexpr\linewidth-1.6pt\relax]{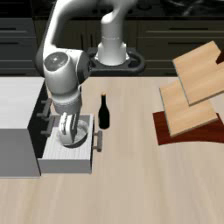}}\\[-2pt]
{\scriptsize F280 / 28.0s\strut}\\[-1pt]
{\scriptsize Earlier progress\strut}\end{minipage}\hfill%
\begin{minipage}[t]{0.137\linewidth}\centering
\setlength{\fboxsep}{0pt}\setlength{\fboxrule}{0.8pt}
\fcolorbox{black}{white}{\includegraphics[width=\dimexpr\linewidth-1.6pt\relax]{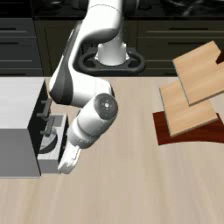}}\\[-2pt]
{\scriptsize \textbf{F340 / 34.0s}\strut}\\[-1pt]
{\scriptsize \textbf{Ref. $a$}\strut}\end{minipage}\hfill%
\begin{minipage}[t]{0.137\linewidth}\centering
\setlength{\fboxsep}{0pt}\setlength{\fboxrule}{0.8pt}
\fcolorbox{white}{white}{\includegraphics[width=\dimexpr\linewidth-1.6pt\relax]{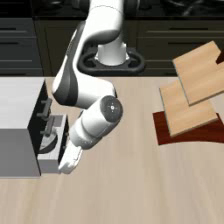}}\\[-2pt]
{\scriptsize F384 / 38.4s\strut}\\[-1pt]
{\scriptsize No advancement\strut}\end{minipage}\hfill%
\begin{minipage}[t]{0.137\linewidth}\centering
\setlength{\fboxsep}{0pt}\setlength{\fboxrule}{0.8pt}
\fcolorbox{white}{white}{\includegraphics[width=\dimexpr\linewidth-1.6pt\relax]{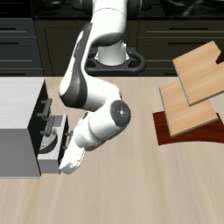}}\\[-2pt]
{\scriptsize F429 / 42.9s\strut}\\[-1pt]
{\scriptsize No advancement\strut}\end{minipage}\hfill%
\begin{minipage}[t]{0.137\linewidth}\centering
\setlength{\fboxsep}{0pt}\setlength{\fboxrule}{0.8pt}
\fcolorbox{white}{white}{\includegraphics[width=\dimexpr\linewidth-1.6pt\relax]{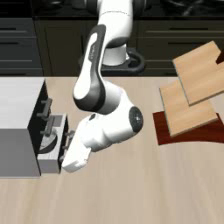}}\\[-2pt]
{\scriptsize F474 / 47.4s\strut}\\[-1pt]
{\scriptsize No advancement\strut}\end{minipage}\hfill%
\begin{minipage}[t]{0.137\linewidth}\centering
\setlength{\fboxsep}{0pt}\setlength{\fboxrule}{0.8pt}
\fcolorbox{LiberoFailure}{white}{\includegraphics[width=\dimexpr\linewidth-1.6pt\relax]{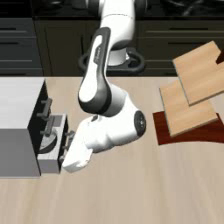}}\\[-2pt]
{\scriptsize F519 / 51.9s\strut}\\[-1pt]
{\scriptsize End $T$\strut}\\[-1pt]{\scriptsize\color{LiberoFailure}\textbf{Failure}\strut}\end{minipage}
\end{minipage}\par\vspace{10pt}
\begin{minipage}{\linewidth}
\setlength{\parskip}{\eventExampleParskip}
{\normalsize \textbf{(c)} Put the yellow-and-white mug in the microwave and close it. The mug is inside, but the door stops closing and remains ajar.\par}
\noindent \begin{minipage}[t]{0.137\linewidth}\centering
\setlength{\fboxsep}{0pt}\setlength{\fboxrule}{0.8pt}
\fcolorbox{white}{white}{\includegraphics[width=\dimexpr\linewidth-1.6pt\relax]{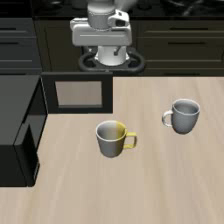}}\\[-2pt]
{\scriptsize F0 / 0.0s\strut}\\[-1pt]
{\scriptsize Context\strut}\end{minipage}\hfill%
\begin{minipage}[t]{0.137\linewidth}\centering
\setlength{\fboxsep}{0pt}\setlength{\fboxrule}{0.8pt}
\fcolorbox{white}{white}{\includegraphics[width=\dimexpr\linewidth-1.6pt\relax]{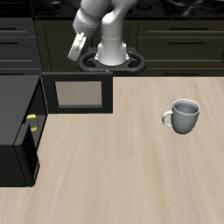}}\\[-2pt]
{\scriptsize F218 / 21.8s\strut}\\[-1pt]
{\scriptsize Earlier progress\strut}\end{minipage}\hfill%
\begin{minipage}[t]{0.137\linewidth}\centering
\setlength{\fboxsep}{0pt}\setlength{\fboxrule}{0.8pt}
\fcolorbox{black}{white}{\includegraphics[width=\dimexpr\linewidth-1.6pt\relax]{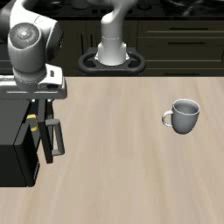}}\\[-2pt]
{\scriptsize \textbf{F278 / 27.8s}\strut}\\[-1pt]
{\scriptsize \textbf{Ref. $a$}\strut}\end{minipage}\hfill%
\begin{minipage}[t]{0.137\linewidth}\centering
\setlength{\fboxsep}{0pt}\setlength{\fboxrule}{0.8pt}
\fcolorbox{white}{white}{\includegraphics[width=\dimexpr\linewidth-1.6pt\relax]{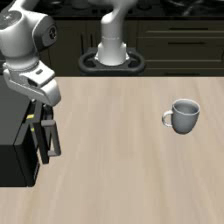}}\\[-2pt]
{\scriptsize F338 / 33.8s\strut}\\[-1pt]
{\scriptsize No advancement\strut}\end{minipage}\hfill%
\begin{minipage}[t]{0.137\linewidth}\centering
\setlength{\fboxsep}{0pt}\setlength{\fboxrule}{0.8pt}
\fcolorbox{white}{white}{\includegraphics[width=\dimexpr\linewidth-1.6pt\relax]{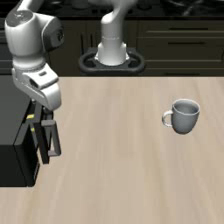}}\\[-2pt]
{\scriptsize F398 / 39.8s\strut}\\[-1pt]
{\scriptsize No advancement\strut}\end{minipage}\hfill%
\begin{minipage}[t]{0.137\linewidth}\centering
\setlength{\fboxsep}{0pt}\setlength{\fboxrule}{0.8pt}
\fcolorbox{white}{white}{\includegraphics[width=\dimexpr\linewidth-1.6pt\relax]{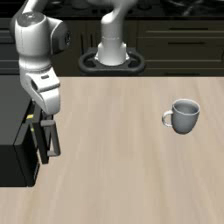}}\\[-2pt]
{\scriptsize F458 / 45.8s\strut}\\[-1pt]
{\scriptsize No advancement\strut}\end{minipage}\hfill%
\begin{minipage}[t]{0.137\linewidth}\centering
\setlength{\fboxsep}{0pt}\setlength{\fboxrule}{0.8pt}
\fcolorbox{LiberoFailure}{white}{\includegraphics[width=\dimexpr\linewidth-1.6pt\relax]{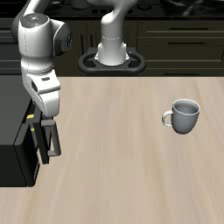}}\\[-2pt]
{\scriptsize F519 / 51.9s\strut}\\[-1pt]
{\scriptsize End $T$\strut}\\[-1pt]{\scriptsize\color{LiberoFailure}\textbf{Failure}\strut}\end{minipage}
\end{minipage}\par\vspace{10pt}
\begin{minipage}{\linewidth}
\setlength{\parskip}{\eventExampleParskip}
{\normalsize \textbf{(d)} Put the mug on the plate and the pudding to its right. The pudding reaches the plate side but stays short of the required right-side placement.\par}
\noindent \begin{minipage}[t]{0.137\linewidth}\centering
\setlength{\fboxsep}{0pt}\setlength{\fboxrule}{0.8pt}
\fcolorbox{white}{white}{\includegraphics[width=\dimexpr\linewidth-1.6pt\relax]{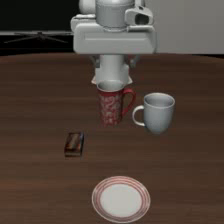}}\\[-2pt]
{\scriptsize F0 / 0.0s\strut}\\[-1pt]
{\scriptsize Context\strut}\end{minipage}\hfill%
\begin{minipage}[t]{0.137\linewidth}\centering
\setlength{\fboxsep}{0pt}\setlength{\fboxrule}{0.8pt}
\fcolorbox{white}{white}{\includegraphics[width=\dimexpr\linewidth-1.6pt\relax]{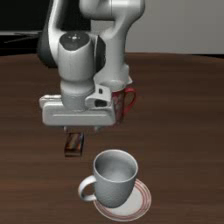}}\\[-2pt]
{\scriptsize F238 / 23.8s\strut}\\[-1pt]
{\scriptsize Earlier progress\strut}\end{minipage}\hfill%
\begin{minipage}[t]{0.137\linewidth}\centering
\setlength{\fboxsep}{0pt}\setlength{\fboxrule}{0.8pt}
\fcolorbox{black}{white}{\includegraphics[width=\dimexpr\linewidth-1.6pt\relax]{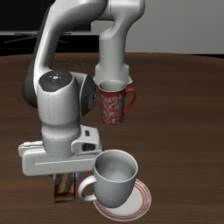}}\\[-2pt]
{\scriptsize \textbf{F298 / 29.8s}\strut}\\[-1pt]
{\scriptsize \textbf{Ref. $a$}\strut}\end{minipage}\hfill%
\begin{minipage}[t]{0.137\linewidth}\centering
\setlength{\fboxsep}{0pt}\setlength{\fboxrule}{0.8pt}
\fcolorbox{white}{white}{\includegraphics[width=\dimexpr\linewidth-1.6pt\relax]{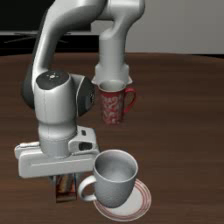}}\\[-2pt]
{\scriptsize F353 / 35.3s\strut}\\[-1pt]
{\scriptsize No advancement\strut}\end{minipage}\hfill%
\begin{minipage}[t]{0.137\linewidth}\centering
\setlength{\fboxsep}{0pt}\setlength{\fboxrule}{0.8pt}
\fcolorbox{white}{white}{\includegraphics[width=\dimexpr\linewidth-1.6pt\relax]{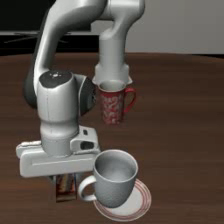}}\\[-2pt]
{\scriptsize F408 / 40.8s\strut}\\[-1pt]
{\scriptsize No advancement\strut}\end{minipage}\hfill%
\begin{minipage}[t]{0.137\linewidth}\centering
\setlength{\fboxsep}{0pt}\setlength{\fboxrule}{0.8pt}
\fcolorbox{white}{white}{\includegraphics[width=\dimexpr\linewidth-1.6pt\relax]{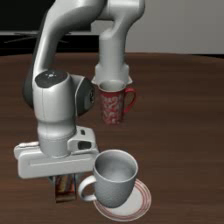}}\\[-2pt]
{\scriptsize F463 / 46.3s\strut}\\[-1pt]
{\scriptsize No advancement\strut}\end{minipage}\hfill%
\begin{minipage}[t]{0.137\linewidth}\centering
\setlength{\fboxsep}{0pt}\setlength{\fboxrule}{0.8pt}
\fcolorbox{LiberoFailure}{white}{\includegraphics[width=\dimexpr\linewidth-1.6pt\relax]{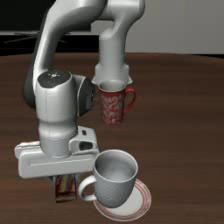}}\\[-2pt]
{\scriptsize F519 / 51.9s\strut}\\[-1pt]
{\scriptsize End $T$\strut}\\[-1pt]{\scriptsize\color{LiberoFailure}\textbf{Failure}\strut}\end{minipage}
\end{minipage}\par\vspace{10pt}
{\scriptsize Source: LIBERO; machine annotations, not human verified. Frames span F0 to the recording endpoint $T$ (10 fps). W: wrist-camera view. Black borders mark annotated event anchors; green/red indicates success/failure.\par}
\endgroup
\clearpage

\begingroup
\setlength{\parskip}{\eventExampleParskip}
\section*{LIBERO: Unrecovered regression}
\label{app:libero-event-examples-end}
{\normalsize An attained task state is lost and remains unrecovered. Black borders mark the reference $a$ and loss $\ell$. Each row starts at F0, shows the change from the attained state to the lost state, and ends at the failed recording endpoint $T$.\par}
\begin{minipage}{\linewidth}
\setlength{\parskip}{\eventExampleParskip}
{\normalsize \textbf{(a)} Put both moka pots on the stove. A held pot falls behind the stove and remains on its side.\par}
\noindent \begin{minipage}[t]{0.137\linewidth}\centering
\setlength{\fboxsep}{0pt}\setlength{\fboxrule}{0.8pt}
\fcolorbox{white}{white}{\includegraphics[width=\dimexpr\linewidth-1.6pt\relax]{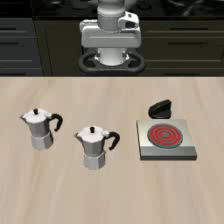}}\\[-2pt]
{\scriptsize F0 / 0.0s\strut}\\[-1pt]
{\scriptsize Context\strut}\end{minipage}\hfill%
\begin{minipage}[t]{0.137\linewidth}\centering
\setlength{\fboxsep}{0pt}\setlength{\fboxrule}{0.8pt}
\fcolorbox{black}{white}{\includegraphics[width=\dimexpr\linewidth-1.6pt\relax]{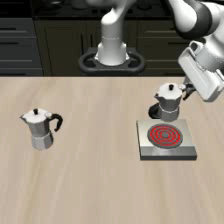}}\\[-2pt]
{\scriptsize \textbf{F180 / 18.0s}\strut}\\[-1pt]
{\scriptsize \textbf{Ref. $a$}\strut}\end{minipage}\hfill%
\begin{minipage}[t]{0.137\linewidth}\centering
\setlength{\fboxsep}{0pt}\setlength{\fboxrule}{0.8pt}
\fcolorbox{white}{white}{\includegraphics[width=\dimexpr\linewidth-1.6pt\relax]{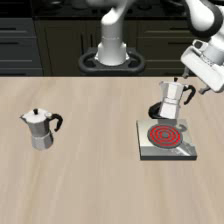}}\\[-2pt]
{\scriptsize F240 / 24.0s\strut}\\[-1pt]
{\scriptsize Before loss\strut}\end{minipage}\hfill%
\begin{minipage}[t]{0.137\linewidth}\centering
\setlength{\fboxsep}{0pt}\setlength{\fboxrule}{0.8pt}
\fcolorbox{black}{white}{\includegraphics[width=\dimexpr\linewidth-1.6pt\relax]{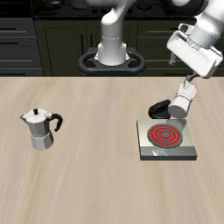}}\\[-2pt]
{\scriptsize \textbf{F250 / 25.0s}\strut}\\[-1pt]
{\scriptsize \textbf{Loss $\ell$}\strut}\end{minipage}\hfill%
\begin{minipage}[t]{0.137\linewidth}\centering
\setlength{\fboxsep}{0pt}\setlength{\fboxrule}{0.8pt}
\fcolorbox{white}{white}{\includegraphics[width=\dimexpr\linewidth-1.6pt\relax]{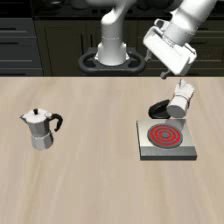}}\\[-2pt]
{\scriptsize F260 / 26.0s\strut}\\[-1pt]
{\scriptsize After loss\strut}\end{minipage}\hfill%
\begin{minipage}[t]{0.137\linewidth}\centering
\setlength{\fboxsep}{0pt}\setlength{\fboxrule}{0.8pt}
\fcolorbox{white}{white}{\includegraphics[width=\dimexpr\linewidth-1.6pt\relax]{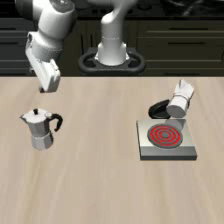}}\\[-2pt]
{\scriptsize F400 / 40.0s\strut}\\[-1pt]
{\scriptsize After loss\strut}\end{minipage}\hfill%
\begin{minipage}[t]{0.137\linewidth}\centering
\setlength{\fboxsep}{0pt}\setlength{\fboxrule}{0.8pt}
\fcolorbox{LiberoFailure}{white}{\includegraphics[width=\dimexpr\linewidth-1.6pt\relax]{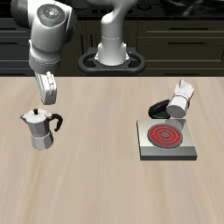}}\\[-2pt]
{\scriptsize F519 / 51.9s\strut}\\[-1pt]
{\scriptsize End $T$\strut}\\[-1pt]{\scriptsize\color{LiberoFailure}\textbf{Failure}\strut}\end{minipage}
\end{minipage}\par\vspace{10pt}
\begin{minipage}{\linewidth}
\setlength{\parskip}{\eventExampleParskip}
{\normalsize \textbf{(b)} Turn on the stove and put the moka pot on it. The carried pot falls beside the stove and stays on its side.\par}
\noindent \begin{minipage}[t]{0.137\linewidth}\centering
\setlength{\fboxsep}{0pt}\setlength{\fboxrule}{0.8pt}
\fcolorbox{white}{white}{\includegraphics[width=\dimexpr\linewidth-1.6pt\relax]{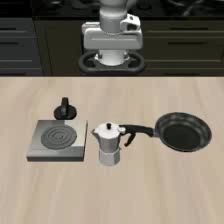}}\\[-2pt]
{\scriptsize F0 / 0.0s\strut}\\[-1pt]
{\scriptsize Context\strut}\end{minipage}\hfill%
\begin{minipage}[t]{0.137\linewidth}\centering
\setlength{\fboxsep}{0pt}\setlength{\fboxrule}{0.8pt}
\fcolorbox{white}{white}{\includegraphics[width=\dimexpr\linewidth-1.6pt\relax]{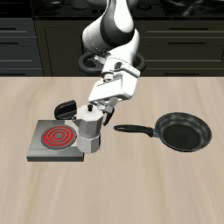}}\\[-2pt]
{\scriptsize F350 / 35.0s\strut}\\[-1pt]
{\scriptsize Before $a$\strut}\end{minipage}\hfill%
\begin{minipage}[t]{0.137\linewidth}\centering
\setlength{\fboxsep}{0pt}\setlength{\fboxrule}{0.8pt}
\fcolorbox{black}{white}{\includegraphics[width=\dimexpr\linewidth-1.6pt\relax]{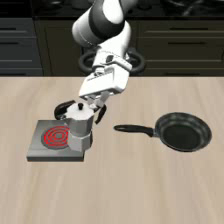}}\\[-2pt]
{\scriptsize \textbf{F360 / 36.0s}\strut}\\[-1pt]
{\scriptsize \textbf{Ref. $a$}\strut}\end{minipage}\hfill%
\begin{minipage}[t]{0.137\linewidth}\centering
\setlength{\fboxsep}{0pt}\setlength{\fboxrule}{0.8pt}
\fcolorbox{black}{white}{\includegraphics[width=\dimexpr\linewidth-1.6pt\relax]{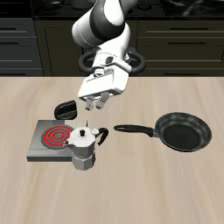}}\\[-2pt]
{\scriptsize \textbf{F362 / 36.2s}\strut}\\[-1pt]
{\scriptsize \textbf{Loss $\ell$}\strut}\end{minipage}\hfill%
\begin{minipage}[t]{0.137\linewidth}\centering
\setlength{\fboxsep}{0pt}\setlength{\fboxrule}{0.8pt}
\fcolorbox{white}{white}{\includegraphics[width=\dimexpr\linewidth-1.6pt\relax]{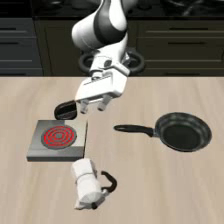}}\\[-2pt]
{\scriptsize F368 / 36.8s\strut}\\[-1pt]
{\scriptsize After loss\strut}\end{minipage}\hfill%
\begin{minipage}[t]{0.137\linewidth}\centering
\setlength{\fboxsep}{0pt}\setlength{\fboxrule}{0.8pt}
\fcolorbox{white}{white}{\includegraphics[width=\dimexpr\linewidth-1.6pt\relax]{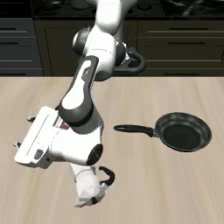}}\\[-2pt]
{\scriptsize F440 / 44.0s\strut}\\[-1pt]
{\scriptsize After loss\strut}\end{minipage}\hfill%
\begin{minipage}[t]{0.137\linewidth}\centering
\setlength{\fboxsep}{0pt}\setlength{\fboxrule}{0.8pt}
\fcolorbox{LiberoFailure}{white}{\includegraphics[width=\dimexpr\linewidth-1.6pt\relax]{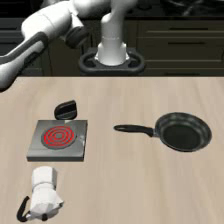}}\\[-2pt]
{\scriptsize F519 / 51.9s\strut}\\[-1pt]
{\scriptsize End $T$\strut}\\[-1pt]{\scriptsize\color{LiberoFailure}\textbf{Failure}\strut}\end{minipage}
\end{minipage}\par\vspace{10pt}
\begin{minipage}{\linewidth}
\setlength{\parskip}{\eventExampleParskip}
{\normalsize \textbf{(c)} Put the white mug on the plate and the pudding to its right. The placed mug loses stable support and stays tilted.\par}
\noindent \begin{minipage}[t]{0.137\linewidth}\centering
\setlength{\fboxsep}{0pt}\setlength{\fboxrule}{0.8pt}
\fcolorbox{white}{white}{\includegraphics[width=\dimexpr\linewidth-1.6pt\relax]{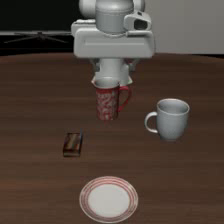}}\\[-2pt]
{\scriptsize F0 / 0.0s\strut}\\[-1pt]
{\scriptsize Context\strut}\end{minipage}\hfill%
\begin{minipage}[t]{0.137\linewidth}\centering
\setlength{\fboxsep}{0pt}\setlength{\fboxrule}{0.8pt}
\fcolorbox{white}{white}{\includegraphics[width=\dimexpr\linewidth-1.6pt\relax]{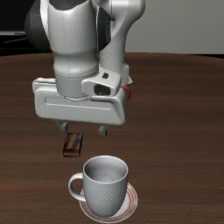}}\\[-2pt]
{\scriptsize F157 / 15.7s\strut}\\[-1pt]
{\scriptsize Before $a$\strut}\end{minipage}\hfill%
\begin{minipage}[t]{0.137\linewidth}\centering
\setlength{\fboxsep}{0pt}\setlength{\fboxrule}{0.8pt}
\fcolorbox{black}{white}{\includegraphics[width=\dimexpr\linewidth-1.6pt\relax]{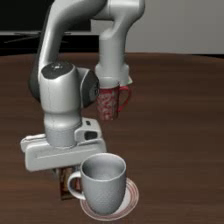}}\\[-2pt]
{\scriptsize \textbf{F291 / 29.1s}\strut}\\[-1pt]
{\scriptsize \textbf{Ref. $a$}\strut}\end{minipage}\hfill%
\begin{minipage}[t]{0.137\linewidth}\centering
\setlength{\fboxsep}{0pt}\setlength{\fboxrule}{0.8pt}
\fcolorbox{black}{white}{\includegraphics[width=\dimexpr\linewidth-1.6pt\relax]{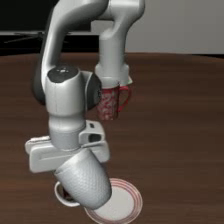}}\\[-2pt]
{\scriptsize \textbf{F298 / 29.8s}\strut}\\[-1pt]
{\scriptsize \textbf{Loss $\ell$}\strut}\end{minipage}\hfill%
\begin{minipage}[t]{0.137\linewidth}\centering
\setlength{\fboxsep}{0pt}\setlength{\fboxrule}{0.8pt}
\fcolorbox{white}{white}{\includegraphics[width=\dimexpr\linewidth-1.6pt\relax]{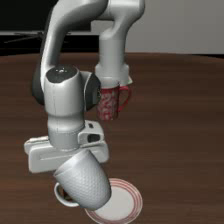}}\\[-2pt]
{\scriptsize F300 / 30.0s\strut}\\[-1pt]
{\scriptsize After loss\strut}\end{minipage}\hfill%
\begin{minipage}[t]{0.137\linewidth}\centering
\setlength{\fboxsep}{0pt}\setlength{\fboxrule}{0.8pt}
\fcolorbox{white}{white}{\includegraphics[width=\dimexpr\linewidth-1.6pt\relax]{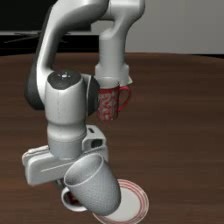}}\\[-2pt]
{\scriptsize F445 / 44.5s\strut}\\[-1pt]
{\scriptsize After loss\strut}\end{minipage}\hfill%
\begin{minipage}[t]{0.137\linewidth}\centering
\setlength{\fboxsep}{0pt}\setlength{\fboxrule}{0.8pt}
\fcolorbox{LiberoFailure}{white}{\includegraphics[width=\dimexpr\linewidth-1.6pt\relax]{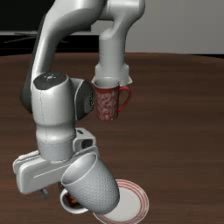}}\\[-2pt]
{\scriptsize F519 / 51.9s\strut}\\[-1pt]
{\scriptsize End $T$\strut}\\[-1pt]{\scriptsize\color{LiberoFailure}\textbf{Failure}\strut}\end{minipage}
\end{minipage}\par\vspace{10pt}
\begin{minipage}{\linewidth}
\setlength{\parskip}{\eventExampleParskip}
{\normalsize \textbf{(d)} Put the yellow-and-white mug in the microwave and close it. The held mug falls and remains upside down outside the microwave.\par}
\noindent \begin{minipage}[t]{0.137\linewidth}\centering
\setlength{\fboxsep}{0pt}\setlength{\fboxrule}{0.8pt}
\fcolorbox{white}{white}{\includegraphics[width=\dimexpr\linewidth-1.6pt\relax]{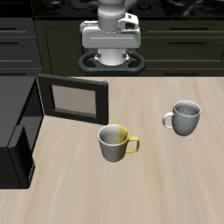}}\\[-2pt]
{\scriptsize F0 / 0.0s\strut}\\[-1pt]
{\scriptsize Context\strut}\end{minipage}\hfill%
\begin{minipage}[t]{0.137\linewidth}\centering
\setlength{\fboxsep}{0pt}\setlength{\fboxrule}{0.8pt}
\fcolorbox{white}{white}{\includegraphics[width=\dimexpr\linewidth-1.6pt\relax]{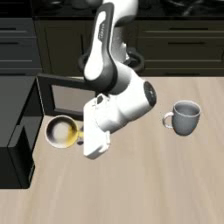}}\\[-2pt]
{\scriptsize F176 / 17.6s\strut}\\[-1pt]
{\scriptsize Before $a$\strut}\end{minipage}\hfill%
\begin{minipage}[t]{0.137\linewidth}\centering
\setlength{\fboxsep}{0pt}\setlength{\fboxrule}{0.8pt}
\fcolorbox{black}{white}{\includegraphics[width=\dimexpr\linewidth-1.6pt\relax]{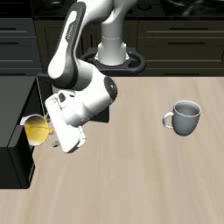}}\\[-2pt]
{\scriptsize \textbf{F284 / 28.4s}\strut}\\[-1pt]
{\scriptsize \textbf{Ref. $a$}\strut}\end{minipage}\hfill%
\begin{minipage}[t]{0.137\linewidth}\centering
\setlength{\fboxsep}{0pt}\setlength{\fboxrule}{0.8pt}
\fcolorbox{black}{white}{\includegraphics[width=\dimexpr\linewidth-1.6pt\relax]{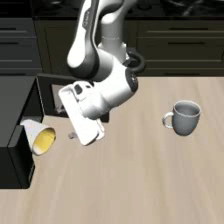}}\\[-2pt]
{\scriptsize \textbf{F294 / 29.4s}\strut}\\[-1pt]
{\scriptsize \textbf{Loss $\ell$}\strut}\end{minipage}\hfill%
\begin{minipage}[t]{0.137\linewidth}\centering
\setlength{\fboxsep}{0pt}\setlength{\fboxrule}{0.8pt}
\fcolorbox{white}{white}{\includegraphics[width=\dimexpr\linewidth-1.6pt\relax]{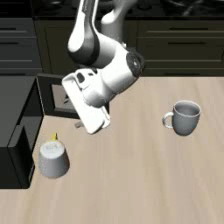}}\\[-2pt]
{\scriptsize F304 / 30.4s\strut}\\[-1pt]
{\scriptsize After loss\strut}\end{minipage}\hfill%
\begin{minipage}[t]{0.137\linewidth}\centering
\setlength{\fboxsep}{0pt}\setlength{\fboxrule}{0.8pt}
\fcolorbox{white}{white}{\includegraphics[width=\dimexpr\linewidth-1.6pt\relax]{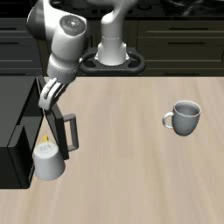}}\\[-2pt]
{\scriptsize F388 / 38.8s\strut}\\[-1pt]
{\scriptsize After loss\strut}\end{minipage}\hfill%
\begin{minipage}[t]{0.137\linewidth}\centering
\setlength{\fboxsep}{0pt}\setlength{\fboxrule}{0.8pt}
\fcolorbox{LiberoFailure}{white}{\includegraphics[width=\dimexpr\linewidth-1.6pt\relax]{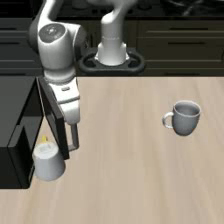}}\\[-2pt]
{\scriptsize F519 / 51.9s\strut}\\[-1pt]
{\scriptsize End $T$\strut}\\[-1pt]{\scriptsize\color{LiberoFailure}\textbf{Failure}\strut}\end{minipage}
\end{minipage}\par\vspace{10pt}
{\scriptsize Source: LIBERO; machine annotations, not human verified. Frames span F0 to the recording endpoint $T$ (10 fps). W: wrist-camera view. Black borders mark annotated event anchors; green/red indicates success/failure.\par}
\endgroup

\clearpage
\endgroup

\clearpage
\begingroup
\definecolor{endpointgreen}{HTML}{15803D}
\definecolor{endpointred}{HTML}{B91C1C}
\begingroup
\setlength{\parskip}{\eventExampleParskip}
\vbox to \textheight\bgroup
\hsize=\linewidth
\section*{Real robots: Progress stall}
\label{app:real-event-examples}
{\normalsize Task progress remains approximately unchanged within the annotated interval ($p_t\approx p_a$). Later frames show renewed advancement.\par}
\begin{minipage}{\linewidth}
\setlength{\parskip}{\eventExampleParskip}
{\normalsize \textbf{(a)} Move all task objects into the white box. The fork remains on the table during repeated grasp attempts. Grasping resumes, and all objects are in the box at the endpoint. \mbox{\textbf{Stall: 26.4\,s}}\par}
\noindent \begin{minipage}[t]{0.137\linewidth}\centering
\setlength{\fboxsep}{0pt}\setlength{\fboxrule}{0.8pt}
\fcolorbox{white}{white}{\includegraphics[width=\dimexpr\linewidth-1.6pt\relax]{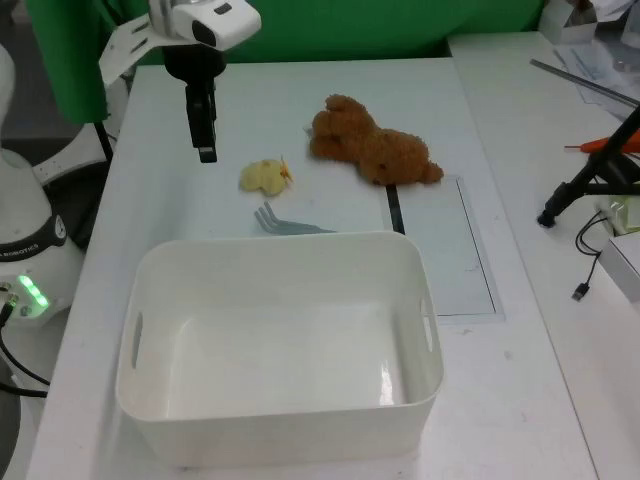}}\\[-2pt]
{\scriptsize F0 / 0.0s\strut}\\[-1pt]
{\scriptsize Context\strut}\end{minipage}\hfill%
\begin{minipage}[t]{0.137\linewidth}\centering
\setlength{\fboxsep}{0pt}\setlength{\fboxrule}{0.8pt}
\fcolorbox{black}{white}{\includegraphics[width=\dimexpr\linewidth-1.6pt\relax,trim=173.3333bp 40.0000bp 0.0000bp 90.0000bp,clip]{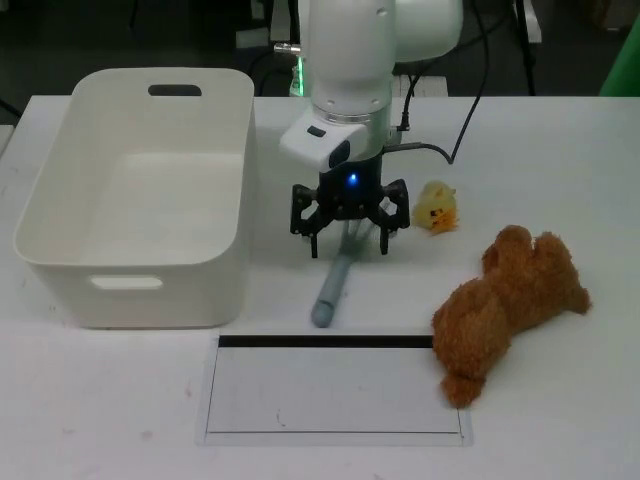}}\\[-2pt]
{\scriptsize \textbf{F45 / 4.5s}\strut}\\[-1pt]
{\scriptsize \textbf{Start $a$}\strut}\end{minipage}\hfill%
\begin{minipage}[t]{0.137\linewidth}\centering
\setlength{\fboxsep}{0pt}\setlength{\fboxrule}{0.8pt}
\fcolorbox{white}{white}{\includegraphics[width=\dimexpr\linewidth-1.6pt\relax,trim=173.3333bp 40.0000bp 0.0000bp 90.0000bp,clip]{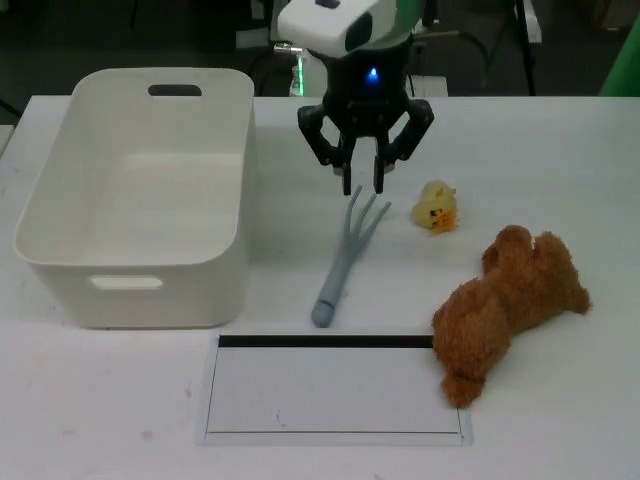}}\\[-2pt]
{\scriptsize F98 / 9.8s\strut}\\[-1pt]
{\scriptsize Stall\strut}\end{minipage}\hfill%
\begin{minipage}[t]{0.137\linewidth}\centering
\setlength{\fboxsep}{0pt}\setlength{\fboxrule}{0.8pt}
\fcolorbox{white}{white}{\includegraphics[width=\dimexpr\linewidth-1.6pt\relax,trim=173.3333bp 40.0000bp 0.0000bp 90.0000bp,clip]{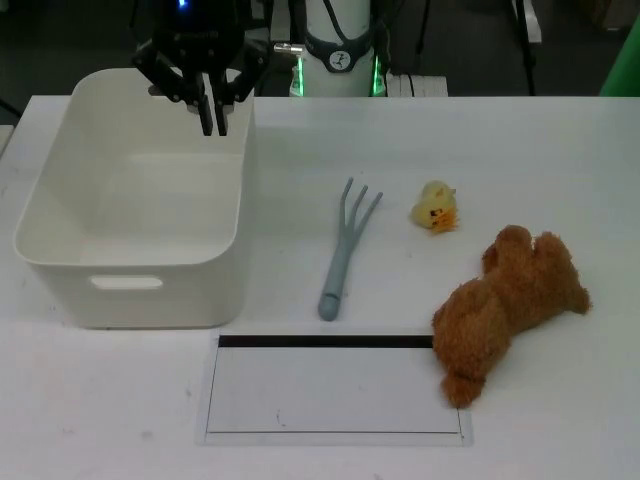}}\\[-2pt]
{\scriptsize F164 / 16.4s\strut}\\[-1pt]
{\scriptsize Stall\strut}\end{minipage}\hfill%
\begin{minipage}[t]{0.137\linewidth}\centering
\setlength{\fboxsep}{0pt}\setlength{\fboxrule}{0.8pt}
\fcolorbox{black}{white}{\includegraphics[width=\dimexpr\linewidth-1.6pt\relax,trim=173.3333bp 40.0000bp 0.0000bp 90.0000bp,clip]{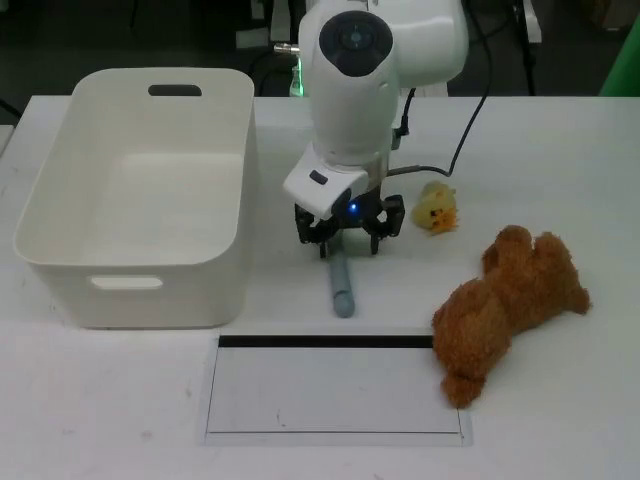}}\\[-2pt]
{\scriptsize \textbf{F309 / 30.9s}\strut}\\[-1pt]
{\scriptsize \textbf{Resume $r$}\strut}\end{minipage}\hfill%
\begin{minipage}[t]{0.137\linewidth}\centering
\setlength{\fboxsep}{0pt}\setlength{\fboxrule}{0.8pt}
\fcolorbox{white}{white}{\includegraphics[width=\dimexpr\linewidth-1.6pt\relax]{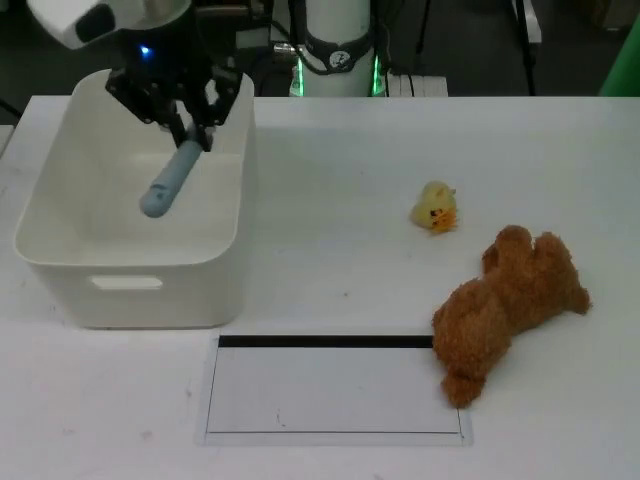}}\\[-2pt]
{\scriptsize F339 / 33.9s\strut}\\[-1pt]
{\scriptsize Later progress\strut}\end{minipage}\hfill%
\begin{minipage}[t]{0.137\linewidth}\centering
\setlength{\fboxsep}{0pt}\setlength{\fboxrule}{0.8pt}
\fcolorbox{endpointgreen}{white}{\includegraphics[width=\dimexpr\linewidth-1.6pt\relax]{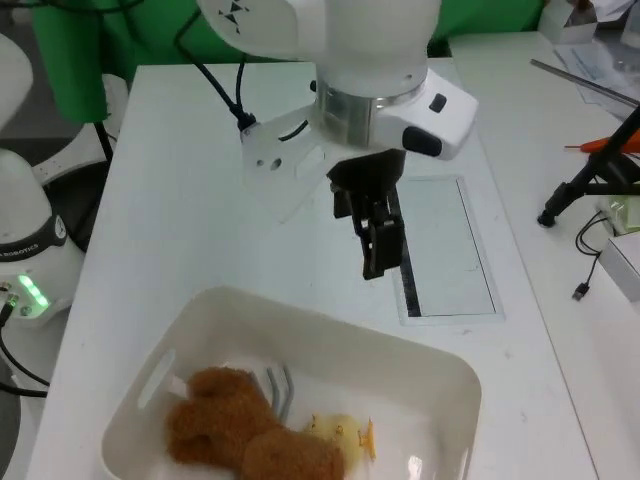}}\\[-2pt]
{\scriptsize \textcolor{endpointgreen}{\textbf{F1002 / 100.2s}}\strut}\\[-1pt]
{\scriptsize \textcolor{endpointgreen}{\textbf{Success / $T$}}\strut}\end{minipage}
\end{minipage}\par
\vfill
\begin{minipage}{\linewidth}
\setlength{\parskip}{\eventExampleParskip}
{\normalsize \textbf{(b)} Move all task objects into the white box. Repeated attempts leave the yellow brush on the table. Manipulation resumes, and the brush, fork and tape roll end inside the box. \mbox{\textbf{Stall: 19.5\,s}}\par}
\noindent \begin{minipage}[t]{0.137\linewidth}\centering
\setlength{\fboxsep}{0pt}\setlength{\fboxrule}{0.8pt}
\fcolorbox{white}{white}{\includegraphics[width=\dimexpr\linewidth-1.6pt\relax]{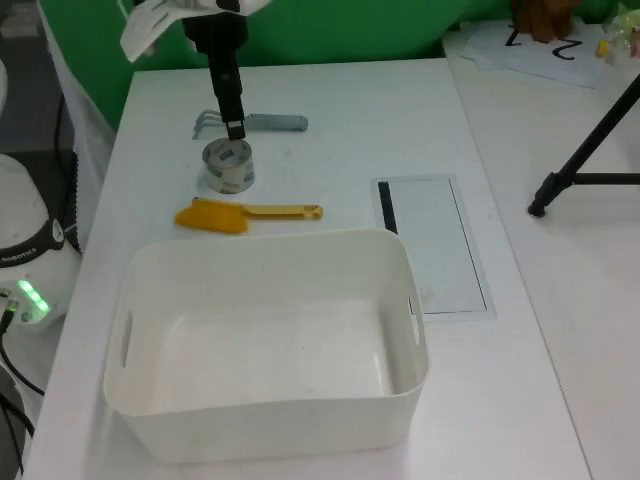}}\\[-2pt]
{\scriptsize F0 / 0.0s\strut}\\[-1pt]
{\scriptsize Context\strut}\end{minipage}\hfill%
\begin{minipage}[t]{0.137\linewidth}\centering
\setlength{\fboxsep}{0pt}\setlength{\fboxrule}{0.8pt}
\fcolorbox{black}{white}{\includegraphics[width=\dimexpr\linewidth-1.6pt\relax,trim=180.8333bp 45.0000bp 5.8333bp 95.0000bp,clip]{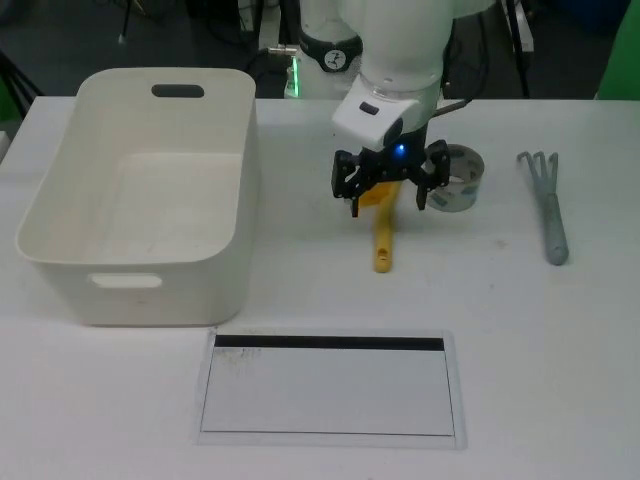}}\\[-2pt]
{\scriptsize \textbf{F39 / 3.9s}\strut}\\[-1pt]
{\scriptsize \textbf{Start $a$}\strut}\end{minipage}\hfill%
\begin{minipage}[t]{0.137\linewidth}\centering
\setlength{\fboxsep}{0pt}\setlength{\fboxrule}{0.8pt}
\fcolorbox{white}{white}{\includegraphics[width=\dimexpr\linewidth-1.6pt\relax,trim=180.8333bp 45.0000bp 5.8333bp 95.0000bp,clip]{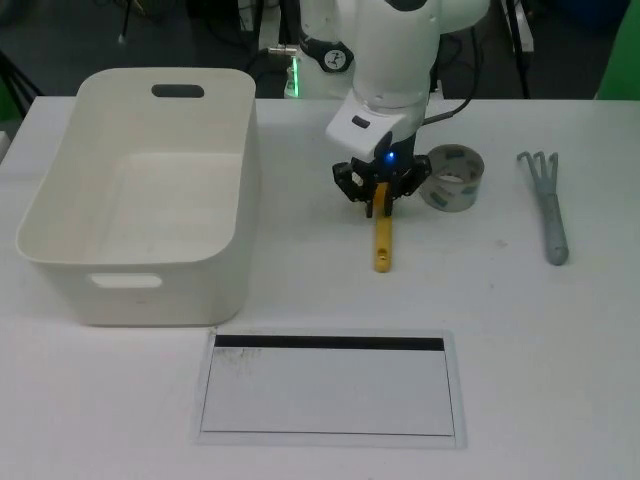}}\\[-2pt]
{\scriptsize F78 / 7.8s\strut}\\[-1pt]
{\scriptsize Stall\strut}\end{minipage}\hfill%
\begin{minipage}[t]{0.137\linewidth}\centering
\setlength{\fboxsep}{0pt}\setlength{\fboxrule}{0.8pt}
\fcolorbox{white}{white}{\includegraphics[width=\dimexpr\linewidth-1.6pt\relax,trim=180.8333bp 45.0000bp 5.8333bp 95.0000bp,clip]{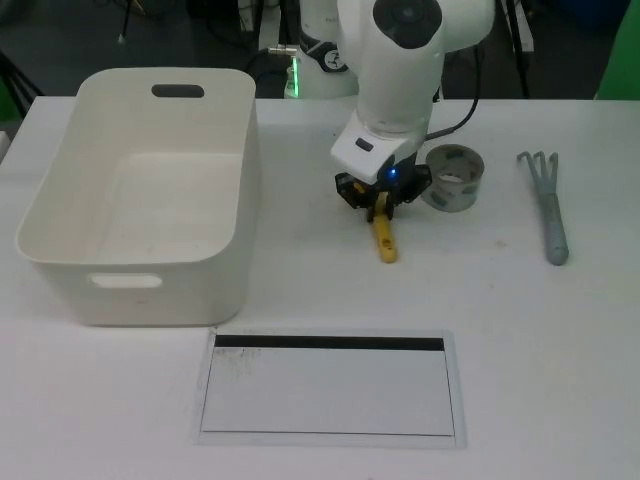}}\\[-2pt]
{\scriptsize F127 / 12.7s\strut}\\[-1pt]
{\scriptsize Stall\strut}\end{minipage}\hfill%
\begin{minipage}[t]{0.137\linewidth}\centering
\setlength{\fboxsep}{0pt}\setlength{\fboxrule}{0.8pt}
\fcolorbox{black}{white}{\includegraphics[width=\dimexpr\linewidth-1.6pt\relax,trim=180.8333bp 45.0000bp 5.8333bp 95.0000bp,clip]{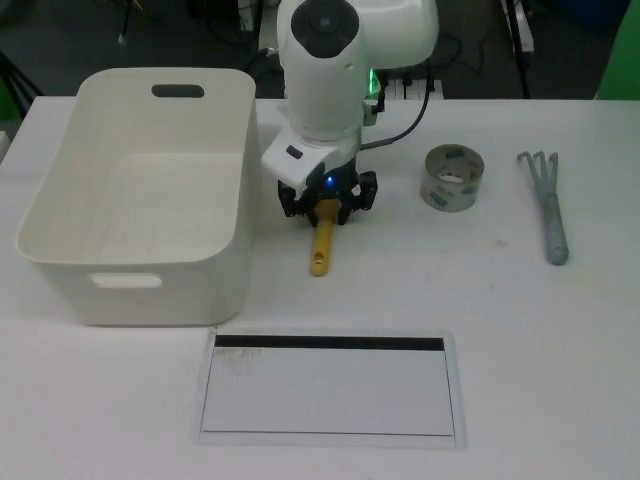}}\\[-2pt]
{\scriptsize \textbf{F234 / 23.4s}\strut}\\[-1pt]
{\scriptsize \textbf{Resume $r$}\strut}\end{minipage}\hfill%
\begin{minipage}[t]{0.137\linewidth}\centering
\setlength{\fboxsep}{0pt}\setlength{\fboxrule}{0.8pt}
\fcolorbox{white}{white}{\includegraphics[width=\dimexpr\linewidth-1.6pt\relax]{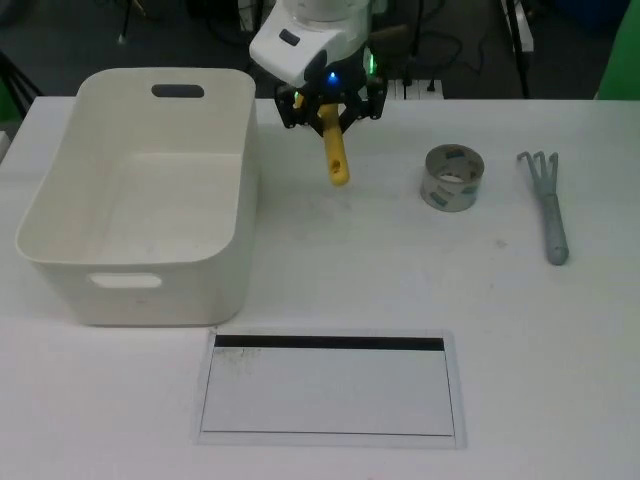}}\\[-2pt]
{\scriptsize F264 / 26.4s\strut}\\[-1pt]
{\scriptsize Later progress\strut}\end{minipage}\hfill%
\begin{minipage}[t]{0.137\linewidth}\centering
\setlength{\fboxsep}{0pt}\setlength{\fboxrule}{0.8pt}
\fcolorbox{endpointgreen}{white}{\includegraphics[width=\dimexpr\linewidth-1.6pt\relax]{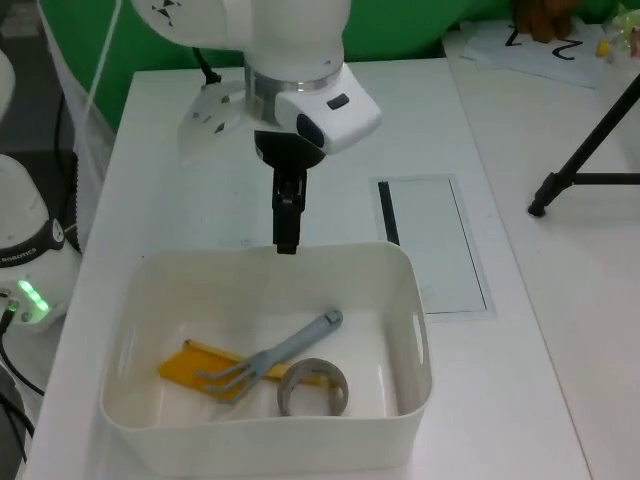}}\\[-2pt]
{\scriptsize \textcolor{endpointgreen}{\textbf{F900 / 90.0s}}\strut}\\[-1pt]
{\scriptsize \textcolor{endpointgreen}{\textbf{Success / $T$}}\strut}\end{minipage}
\end{minipage}\par
\vfill
\begin{minipage}{\linewidth}
\setlength{\parskip}{\eventExampleParskip}
{\normalsize \textbf{(c)} Nest the three bowls into one stack. The blue bowl remains on the table during repeated approach attempts. The final frame shows all three bowls nested together. \mbox{\textbf{Stall: 26.6\,s}}\par}
\noindent \begin{minipage}[t]{0.137\linewidth}\centering
\setlength{\fboxsep}{0pt}\setlength{\fboxrule}{0.8pt}
\fcolorbox{white}{white}{\includegraphics[width=\dimexpr\linewidth-1.6pt\relax,trim=145.0000bp 0.0000bp 75.0000bp 165.0000bp,clip]{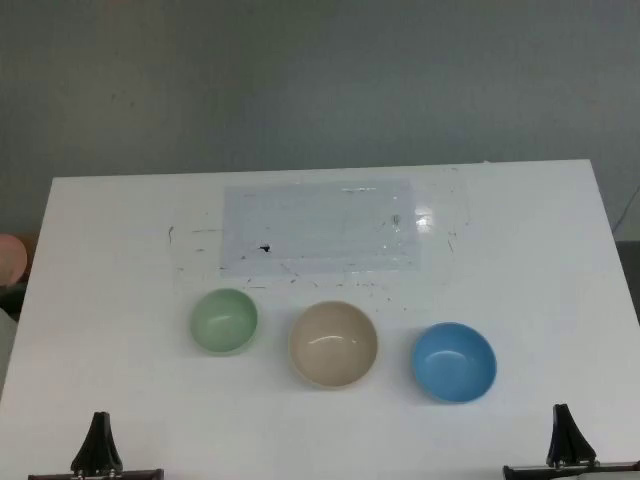}}\\[-2pt]
{\scriptsize F0 / 0.0s\strut}\\[-1pt]
{\scriptsize Context\strut}\end{minipage}\hfill%
\begin{minipage}[t]{0.137\linewidth}\centering
\setlength{\fboxsep}{0pt}\setlength{\fboxrule}{0.8pt}
\fcolorbox{black}{white}{\includegraphics[width=\dimexpr\linewidth-1.6pt\relax,trim=145.0000bp 0.0000bp 75.0000bp 165.0000bp,clip]{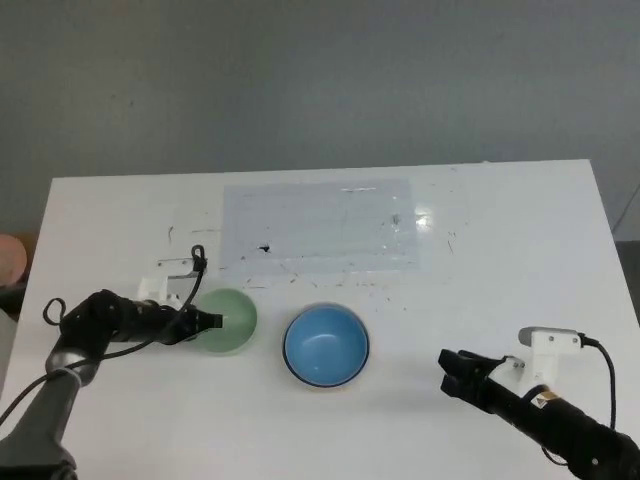}}\\[-2pt]
{\scriptsize \textbf{F160 / 16.0s}\strut}\\[-1pt]
{\scriptsize \textbf{Start $a$}\strut}\end{minipage}\hfill%
\begin{minipage}[t]{0.137\linewidth}\centering
\setlength{\fboxsep}{0pt}\setlength{\fboxrule}{0.8pt}
\fcolorbox{white}{white}{\includegraphics[width=\dimexpr\linewidth-1.6pt\relax,trim=145.0000bp 0.0000bp 75.0000bp 165.0000bp,clip]{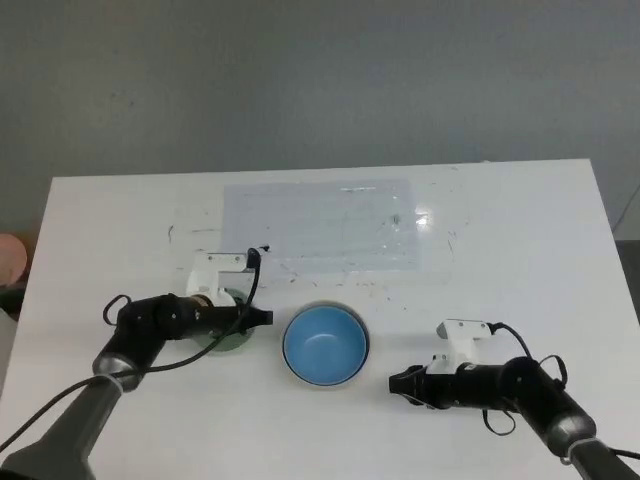}}\\[-2pt]
{\scriptsize F213 / 21.3s\strut}\\[-1pt]
{\scriptsize Stall\strut}\end{minipage}\hfill%
\begin{minipage}[t]{0.137\linewidth}\centering
\setlength{\fboxsep}{0pt}\setlength{\fboxrule}{0.8pt}
\fcolorbox{white}{white}{\includegraphics[width=\dimexpr\linewidth-1.6pt\relax,trim=145.0000bp 0.0000bp 75.0000bp 165.0000bp,clip]{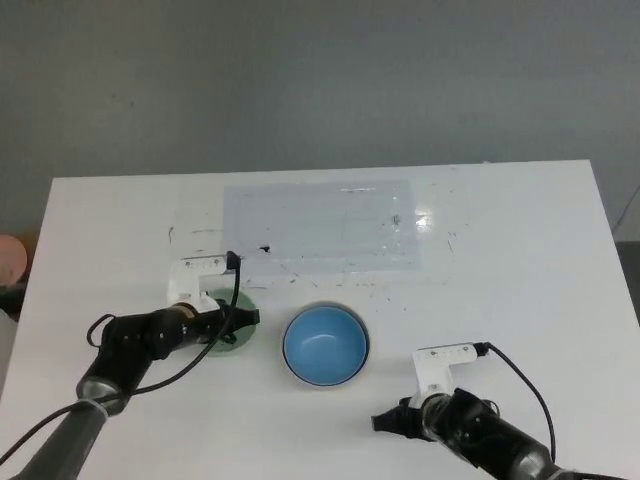}}\\[-2pt]
{\scriptsize F280 / 28.0s\strut}\\[-1pt]
{\scriptsize Stall\strut}\end{minipage}\hfill%
\begin{minipage}[t]{0.137\linewidth}\centering
\setlength{\fboxsep}{0pt}\setlength{\fboxrule}{0.8pt}
\fcolorbox{black}{white}{\includegraphics[width=\dimexpr\linewidth-1.6pt\relax,trim=145.0000bp 0.0000bp 75.0000bp 165.0000bp,clip]{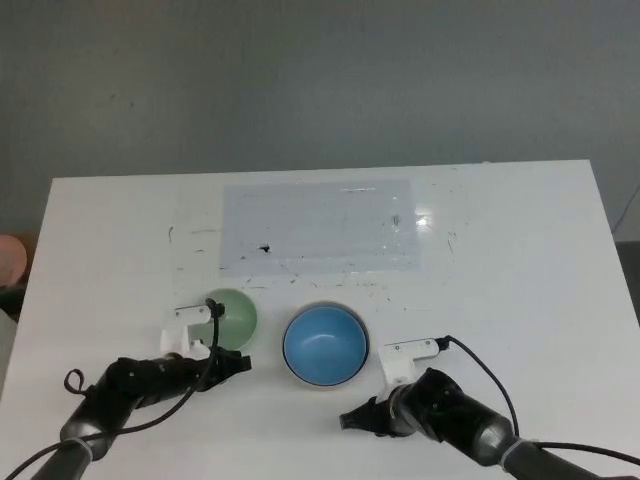}}\\[-2pt]
{\scriptsize \textbf{F426 / 42.6s}\strut}\\[-1pt]
{\scriptsize \textbf{Resume $r$}\strut}\end{minipage}\hfill%
\begin{minipage}[t]{0.137\linewidth}\centering
\setlength{\fboxsep}{0pt}\setlength{\fboxrule}{0.8pt}
\fcolorbox{white}{white}{\includegraphics[width=\dimexpr\linewidth-1.6pt\relax]{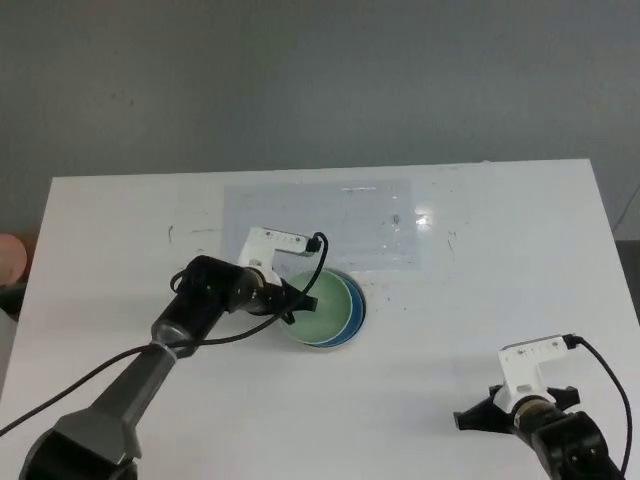}}\\[-2pt]
{\scriptsize F869 / 86.9s\strut}\\[-1pt]
{\scriptsize Later progress\strut}\end{minipage}\hfill%
\begin{minipage}[t]{0.137\linewidth}\centering
\setlength{\fboxsep}{0pt}\setlength{\fboxrule}{0.8pt}
\fcolorbox{endpointgreen}{white}{\includegraphics[width=\dimexpr\linewidth-1.6pt\relax]{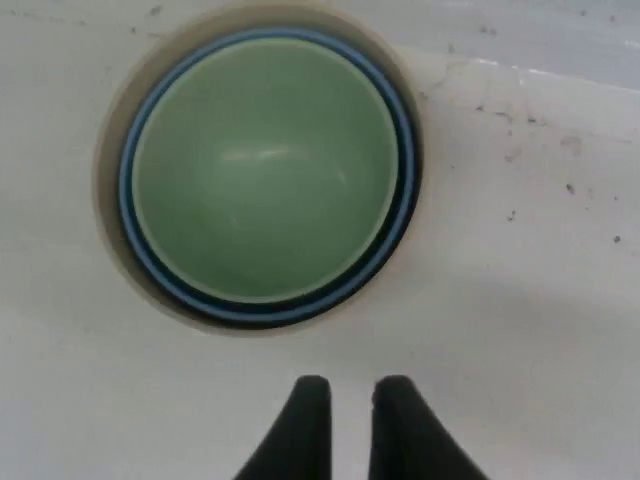}}\\[-2pt]
{\scriptsize \textcolor{endpointgreen}{\textbf{F1023 / 102.3s}}\strut}\\[-1pt]
{\scriptsize \textcolor{endpointgreen}{\textbf{Success / $T$}}\strut}\end{minipage}
\end{minipage}\par
\vfill
\begin{minipage}{\linewidth}
\setlength{\parskip}{\eventExampleParskip}
{\normalsize \textbf{(d)} Nest the three bowls into one stack. Repeated attempts leave the green bowl outside the stack. The final frame shows the three bowls nested together. \mbox{\textbf{Stall: 24.9\,s}}\par}
\noindent \begin{minipage}[t]{0.137\linewidth}\centering
\setlength{\fboxsep}{0pt}\setlength{\fboxrule}{0.8pt}
\fcolorbox{white}{white}{\includegraphics[width=\dimexpr\linewidth-1.6pt\relax,trim=140.0000bp 0.0000bp 80.0000bp 165.0000bp,clip]{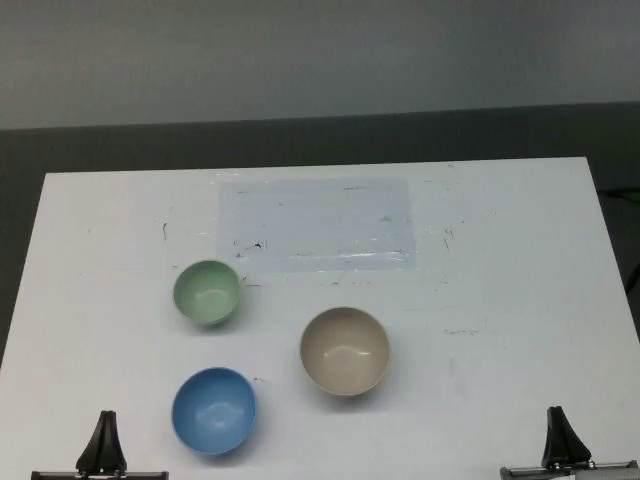}}\\[-2pt]
{\scriptsize F0 / 0.0s\strut}\\[-1pt]
{\scriptsize Context\strut}\end{minipage}\hfill%
\begin{minipage}[t]{0.137\linewidth}\centering
\setlength{\fboxsep}{0pt}\setlength{\fboxrule}{0.8pt}
\fcolorbox{black}{white}{\includegraphics[width=\dimexpr\linewidth-1.6pt\relax,trim=140.0000bp 0.0000bp 80.0000bp 165.0000bp,clip]{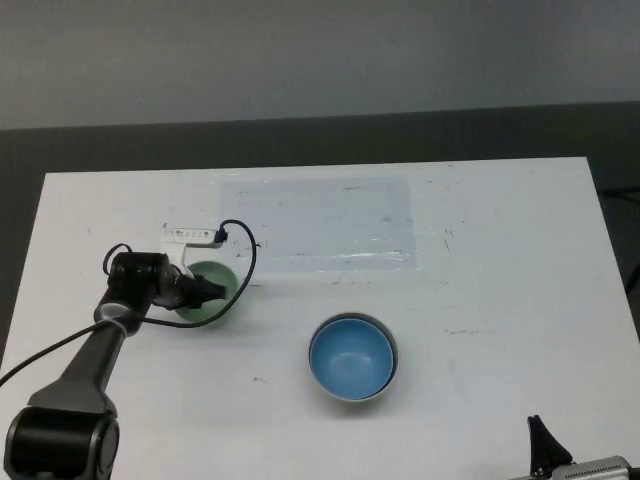}}\\[-2pt]
{\scriptsize \textbf{F286 / 28.6s}\strut}\\[-1pt]
{\scriptsize \textbf{Start $a$}\strut}\end{minipage}\hfill%
\begin{minipage}[t]{0.137\linewidth}\centering
\setlength{\fboxsep}{0pt}\setlength{\fboxrule}{0.8pt}
\fcolorbox{white}{white}{\includegraphics[width=\dimexpr\linewidth-1.6pt\relax,trim=140.0000bp 0.0000bp 80.0000bp 165.0000bp,clip]{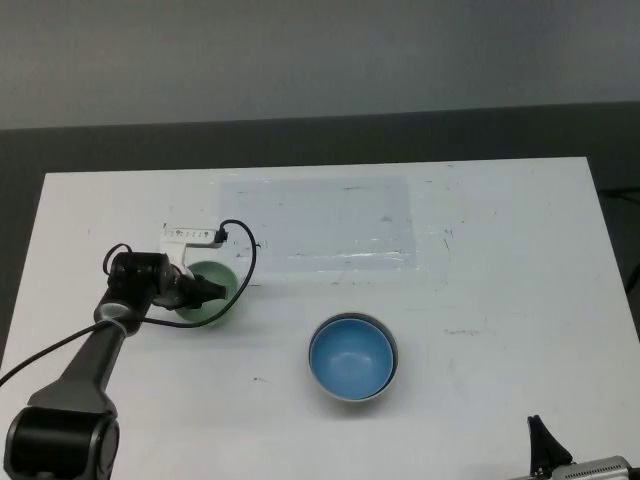}}\\[-2pt]
{\scriptsize F336 / 33.6s\strut}\\[-1pt]
{\scriptsize Stall\strut}\end{minipage}\hfill%
\begin{minipage}[t]{0.137\linewidth}\centering
\setlength{\fboxsep}{0pt}\setlength{\fboxrule}{0.8pt}
\fcolorbox{white}{white}{\includegraphics[width=\dimexpr\linewidth-1.6pt\relax,trim=140.0000bp 0.0000bp 80.0000bp 165.0000bp,clip]{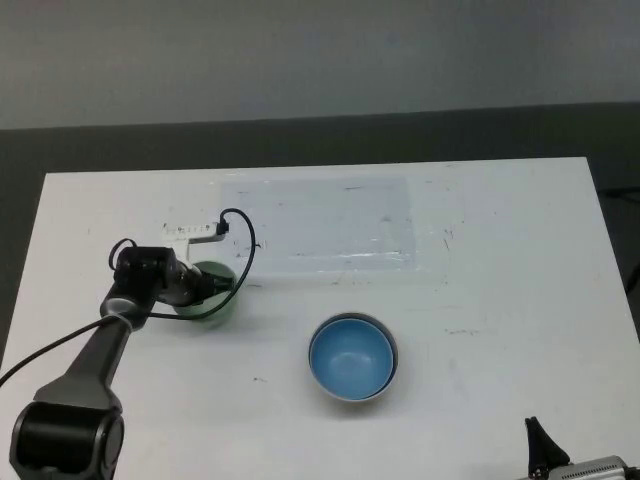}}\\[-2pt]
{\scriptsize F398 / 39.8s\strut}\\[-1pt]
{\scriptsize Stall\strut}\end{minipage}\hfill%
\begin{minipage}[t]{0.137\linewidth}\centering
\setlength{\fboxsep}{0pt}\setlength{\fboxrule}{0.8pt}
\fcolorbox{black}{white}{\includegraphics[width=\dimexpr\linewidth-1.6pt\relax,trim=140.0000bp 0.0000bp 80.0000bp 165.0000bp,clip]{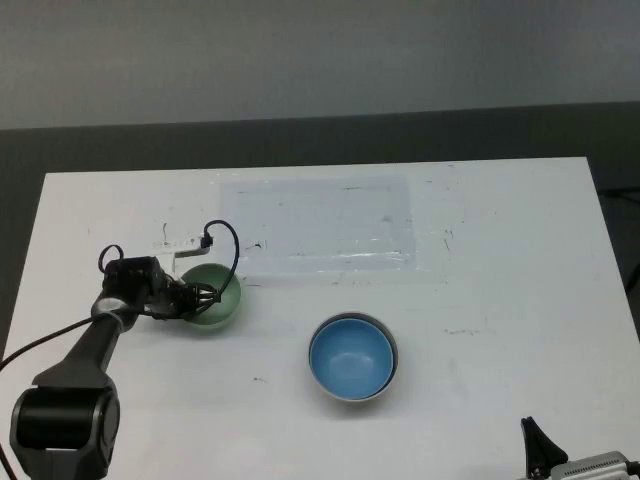}}\\[-2pt]
{\scriptsize \textbf{F535 / 53.5s}\strut}\\[-1pt]
{\scriptsize \textbf{Resume $r$}\strut}\end{minipage}\hfill%
\begin{minipage}[t]{0.137\linewidth}\centering
\setlength{\fboxsep}{0pt}\setlength{\fboxrule}{0.8pt}
\fcolorbox{white}{white}{\includegraphics[width=\dimexpr\linewidth-1.6pt\relax]{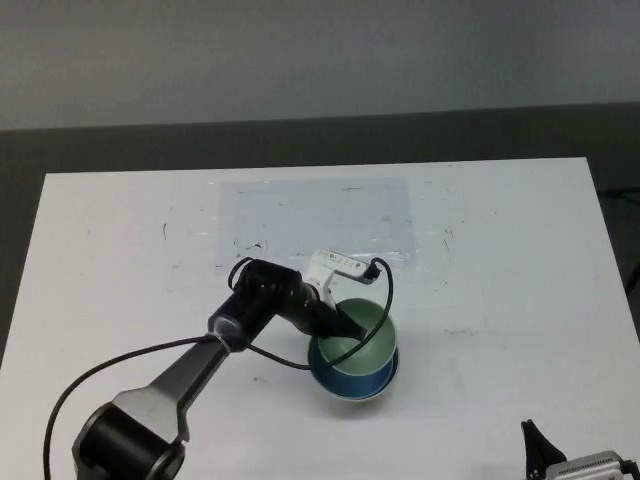}}\\[-2pt]
{\scriptsize F565 / 56.5s\strut}\\[-1pt]
{\scriptsize Later progress\strut}\end{minipage}\hfill%
\begin{minipage}[t]{0.137\linewidth}\centering
\setlength{\fboxsep}{0pt}\setlength{\fboxrule}{0.8pt}
\fcolorbox{endpointgreen}{white}{\includegraphics[width=\dimexpr\linewidth-1.6pt\relax,trim=140.0000bp 0.0000bp 80.0000bp 165.0000bp,clip]{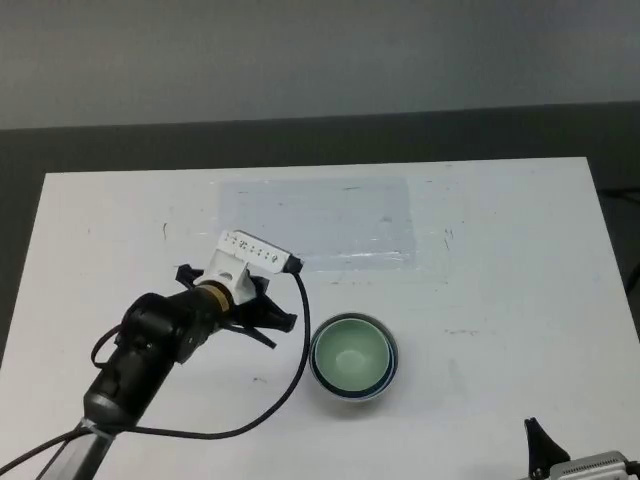}}\\[-2pt]
{\scriptsize \textcolor{endpointgreen}{\textbf{F596 / 59.6s}}\strut}\\[-1pt]
{\scriptsize \textcolor{endpointgreen}{\textbf{Success / $T$}}\strut}\end{minipage}
\end{minipage}\par
\vspace{14pt}
{\scriptsize Source: RoboChallenge; existing human annotations. Black borders mark annotated event anchors. Green/red endpoints indicate success/failure. Frame numbers and times identify positions in each video. Some views are cropped to highlight task-relevant regions.\par}
\egroup
\endgroup
\clearpage

\begingroup
\setlength{\parskip}{\eventExampleParskip}
\vbox to \textheight\bgroup
\hsize=\linewidth
\section*{Real robots: Regression and recovery}
{\normalsize An achieved task state is lost and then restored. Supported loss states have lower progress than both references ($p_t<p_a$, $p_t<p_r$).\par}
\begin{minipage}{\linewidth}
\setlength{\parskip}{\eventExampleParskip}
{\normalsize \textbf{(a)} Place all loose fruits in the black wire basket. The green fruit is held, falls back onto the table, and is grasped again. All fruits are in the basket at the endpoint.\par}
\noindent \begin{minipage}[t]{0.137\linewidth}\centering
\setlength{\fboxsep}{0pt}\setlength{\fboxrule}{0.8pt}
\fcolorbox{white}{white}{\includegraphics[width=\dimexpr\linewidth-1.6pt\relax]{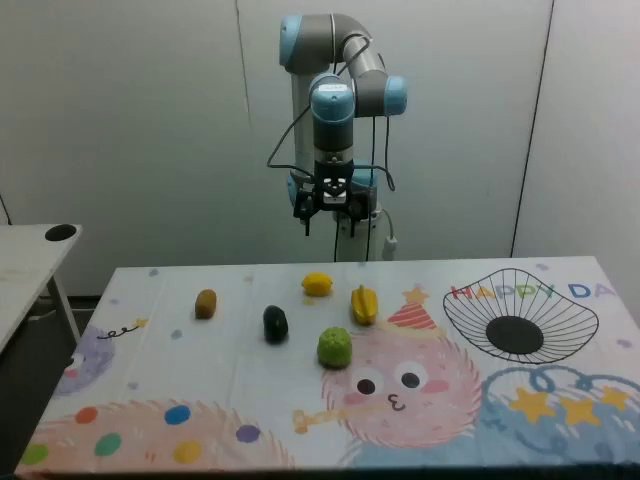}}\\[-2pt]
{\scriptsize F0 / 0.0s\strut}\\[-1pt]
{\scriptsize Context\strut}\end{minipage}\hfill%
\begin{minipage}[t]{0.137\linewidth}\centering
\setlength{\fboxsep}{0pt}\setlength{\fboxrule}{0.8pt}
\fcolorbox{black}{white}{\includegraphics[width=\dimexpr\linewidth-1.6pt\relax]{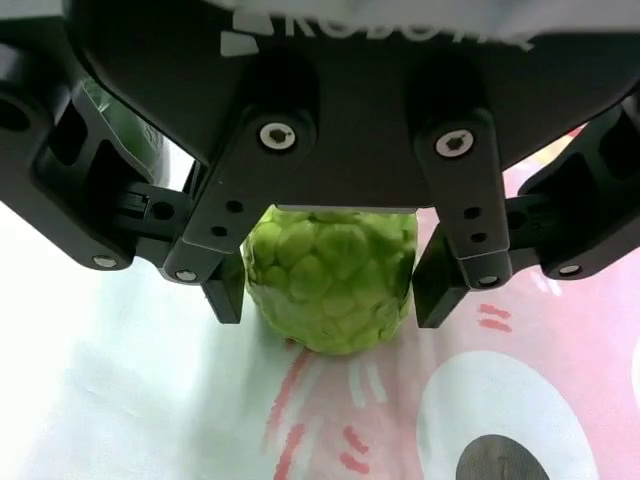}}\\[-2pt]
{\scriptsize \textbf{F327 / 32.7s}\strut}\\[-1pt]
{\scriptsize \textbf{Ref. $a$}\strut}\end{minipage}\hfill%
\begin{minipage}[t]{0.137\linewidth}\centering
\setlength{\fboxsep}{0pt}\setlength{\fboxrule}{0.8pt}
\fcolorbox{white}{white}{\includegraphics[width=\dimexpr\linewidth-1.6pt\relax]{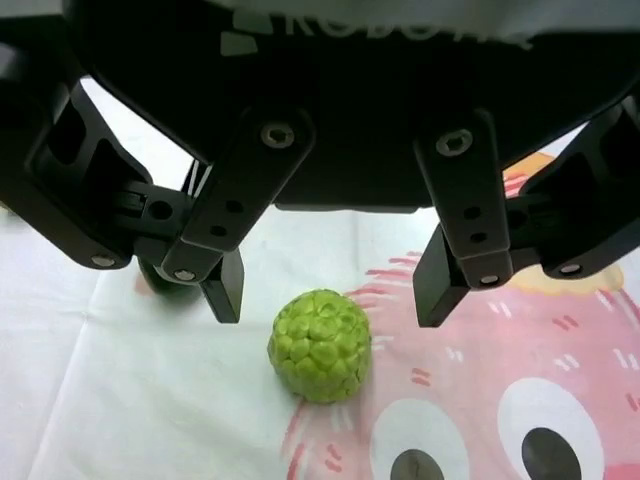}}\\[-2pt]
{\scriptsize F351 / 35.1s\strut}\\[-1pt]
{\scriptsize Loss evidence\strut}\end{minipage}\hfill%
\begin{minipage}[t]{0.137\linewidth}\centering
\setlength{\fboxsep}{0pt}\setlength{\fboxrule}{0.8pt}
\fcolorbox{white}{white}{\includegraphics[width=\dimexpr\linewidth-1.6pt\relax]{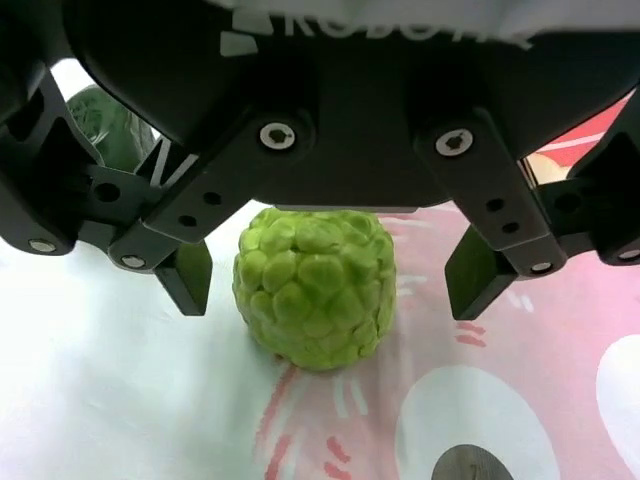}}\\[-2pt]
{\scriptsize F366 / 36.6s\strut}\\[-1pt]
{\scriptsize Reapproach\strut}\end{minipage}\hfill%
\begin{minipage}[t]{0.137\linewidth}\centering
\setlength{\fboxsep}{0pt}\setlength{\fboxrule}{0.8pt}
\fcolorbox{black}{white}{\includegraphics[width=\dimexpr\linewidth-1.6pt\relax]{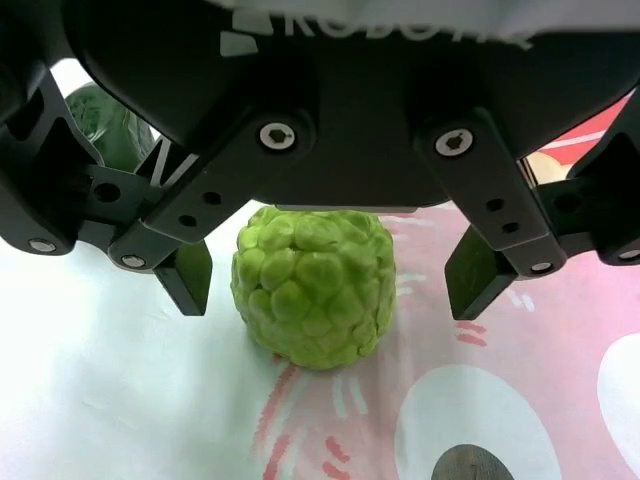}}\\[-2pt]
{\scriptsize \textbf{F376 / 37.6s}\strut}\\[-1pt]
{\scriptsize \textbf{Recovery $r$}\strut}\end{minipage}\hfill%
\begin{minipage}[t]{0.137\linewidth}\centering
\setlength{\fboxsep}{0pt}\setlength{\fboxrule}{0.8pt}
\fcolorbox{white}{white}{\includegraphics[width=\dimexpr\linewidth-1.6pt\relax]{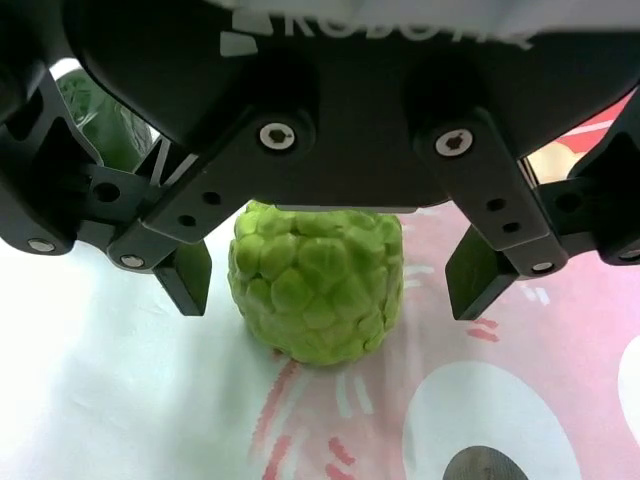}}\\[-2pt]
{\scriptsize F388 / 38.8s\strut}\\[-1pt]
{\scriptsize After recovery\strut}\end{minipage}\hfill%
\begin{minipage}[t]{0.137\linewidth}\centering
\setlength{\fboxsep}{0pt}\setlength{\fboxrule}{0.8pt}
\fcolorbox{endpointgreen}{white}{\includegraphics[width=\dimexpr\linewidth-1.6pt\relax,trim=346.6667bp 0.0000bp 0.0000bp 260.0000bp,clip]{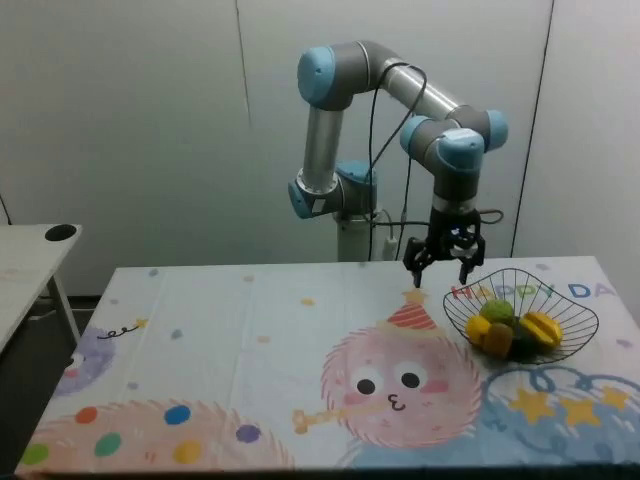}}\\[-2pt]
{\scriptsize \textcolor{endpointgreen}{\textbf{F1804 / 180.4s}}\strut}\\[-1pt]
{\scriptsize \textcolor{endpointgreen}{\textbf{Success / $T$}}\strut}\end{minipage}
\end{minipage}\par
\vfill
\begin{minipage}{\linewidth}
\setlength{\parskip}{\eventExampleParskip}
{\normalsize \textbf{(b)} Place all loose fruits in the black wire basket. The brown fruit slips from the grasp and is recovered. The final overview shows the fruits in the basket and the gripper withdrawn.\par}
\noindent \begin{minipage}[t]{0.137\linewidth}\centering
\setlength{\fboxsep}{0pt}\setlength{\fboxrule}{0.8pt}
\fcolorbox{white}{white}{\includegraphics[width=\dimexpr\linewidth-1.6pt\relax]{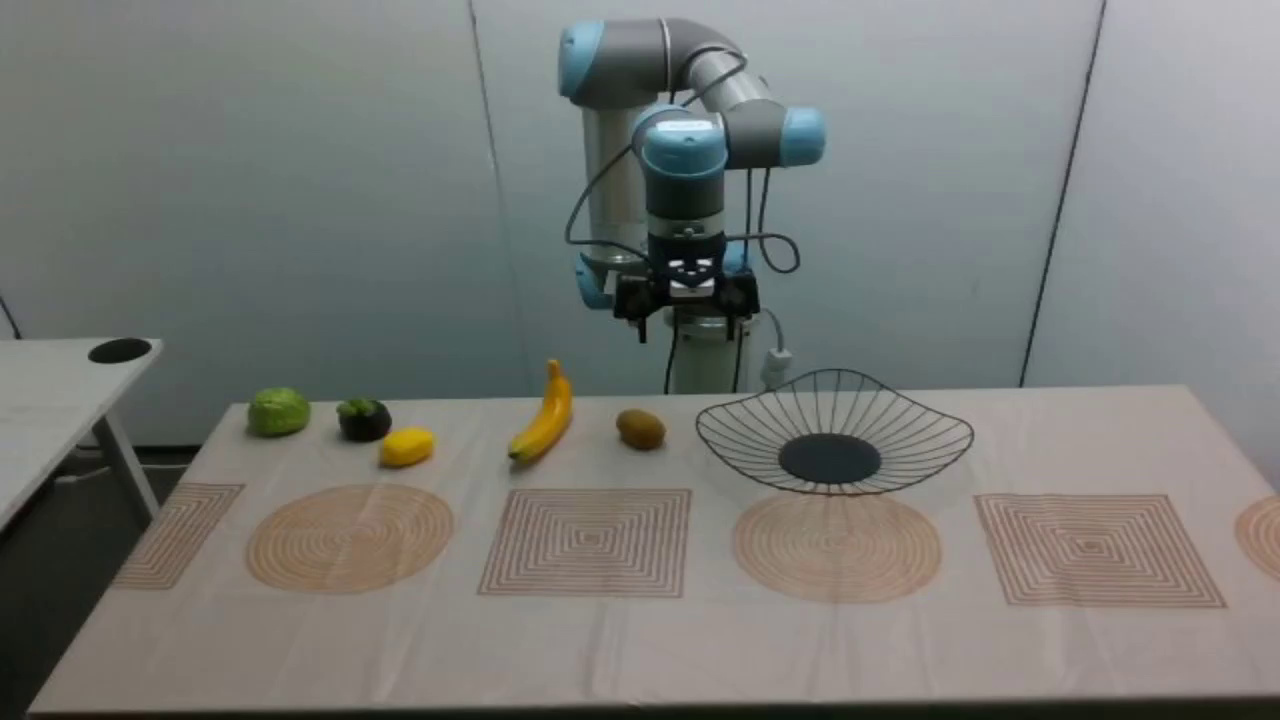}}\\[-2pt]
{\scriptsize F0 / 0.0s\strut}\\[-1pt]
{\scriptsize Context\strut}\end{minipage}\hfill%
\begin{minipage}[t]{0.137\linewidth}\centering
\setlength{\fboxsep}{0pt}\setlength{\fboxrule}{0.8pt}
\fcolorbox{black}{white}{\includegraphics[width=\dimexpr\linewidth-1.6pt\relax]{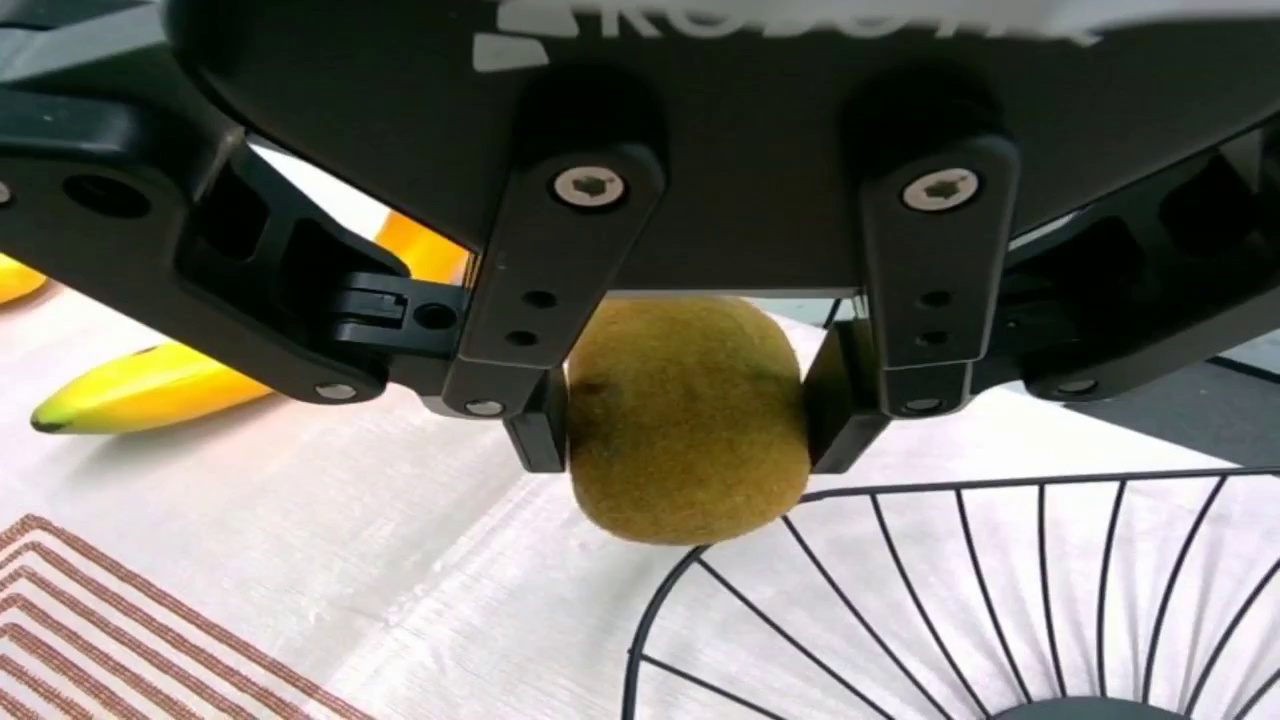}}\\[-2pt]
{\scriptsize \textbf{F282 / 28.2s}\strut}\\[-1pt]
{\scriptsize \textbf{Ref. $a$}\strut}\end{minipage}\hfill%
\begin{minipage}[t]{0.137\linewidth}\centering
\setlength{\fboxsep}{0pt}\setlength{\fboxrule}{0.8pt}
\fcolorbox{white}{white}{\includegraphics[width=\dimexpr\linewidth-1.6pt\relax]{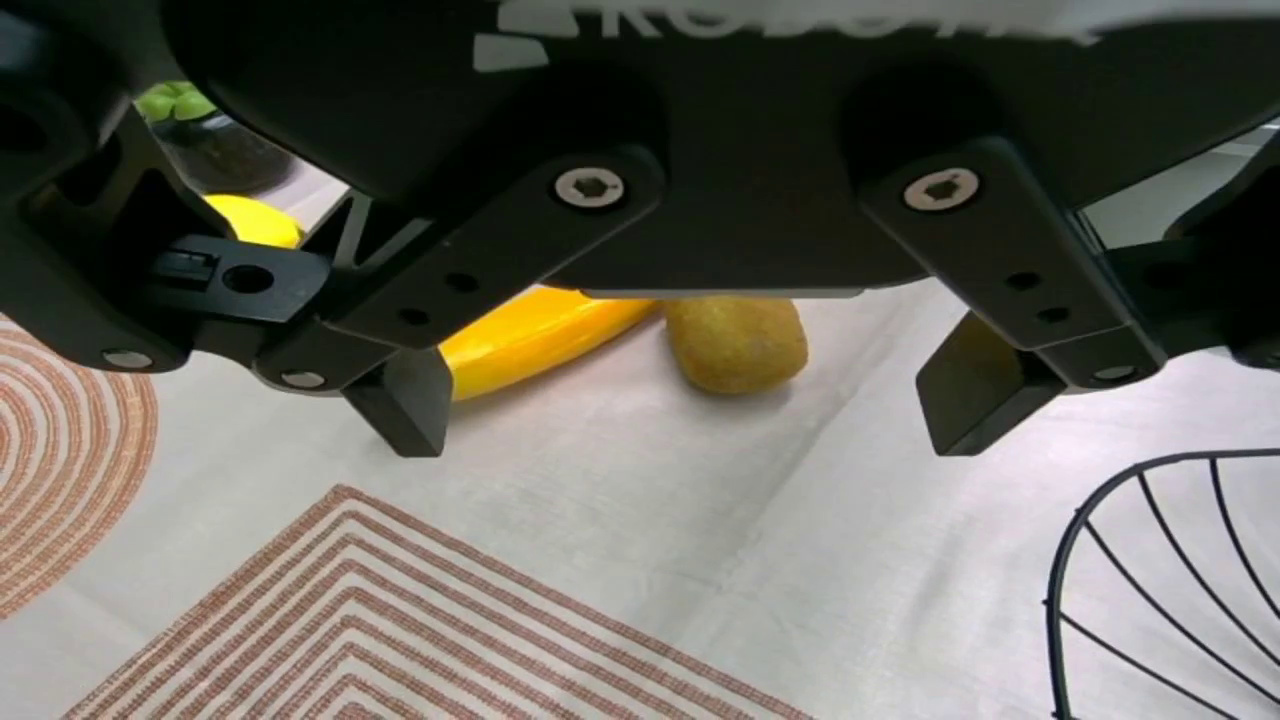}}\\[-2pt]
{\scriptsize F476 / 47.6s\strut}\\[-1pt]
{\scriptsize Loss evidence\strut}\end{minipage}\hfill%
\begin{minipage}[t]{0.137\linewidth}\centering
\setlength{\fboxsep}{0pt}\setlength{\fboxrule}{0.8pt}
\fcolorbox{white}{white}{\includegraphics[width=\dimexpr\linewidth-1.6pt\relax]{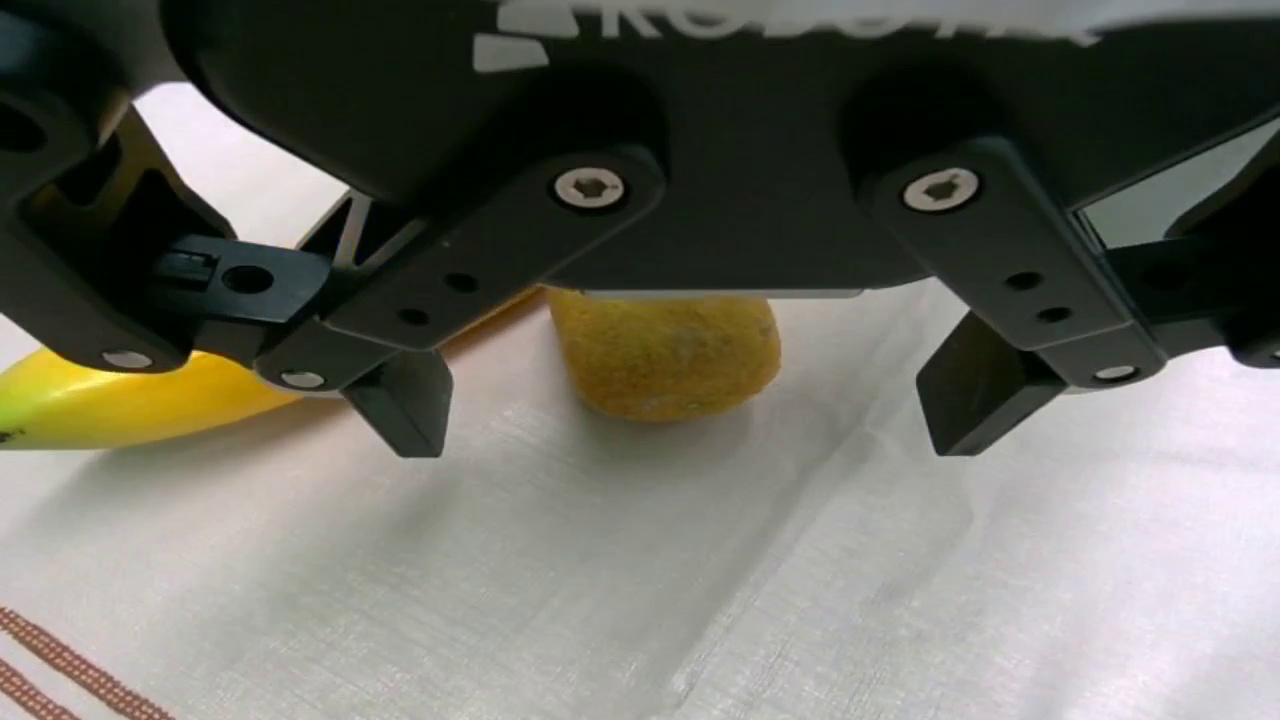}}\\[-2pt]
{\scriptsize F593 / 59.3s\strut}\\[-1pt]
{\scriptsize Reapproach\strut}\end{minipage}\hfill%
\begin{minipage}[t]{0.137\linewidth}\centering
\setlength{\fboxsep}{0pt}\setlength{\fboxrule}{0.8pt}
\fcolorbox{black}{white}{\includegraphics[width=\dimexpr\linewidth-1.6pt\relax]{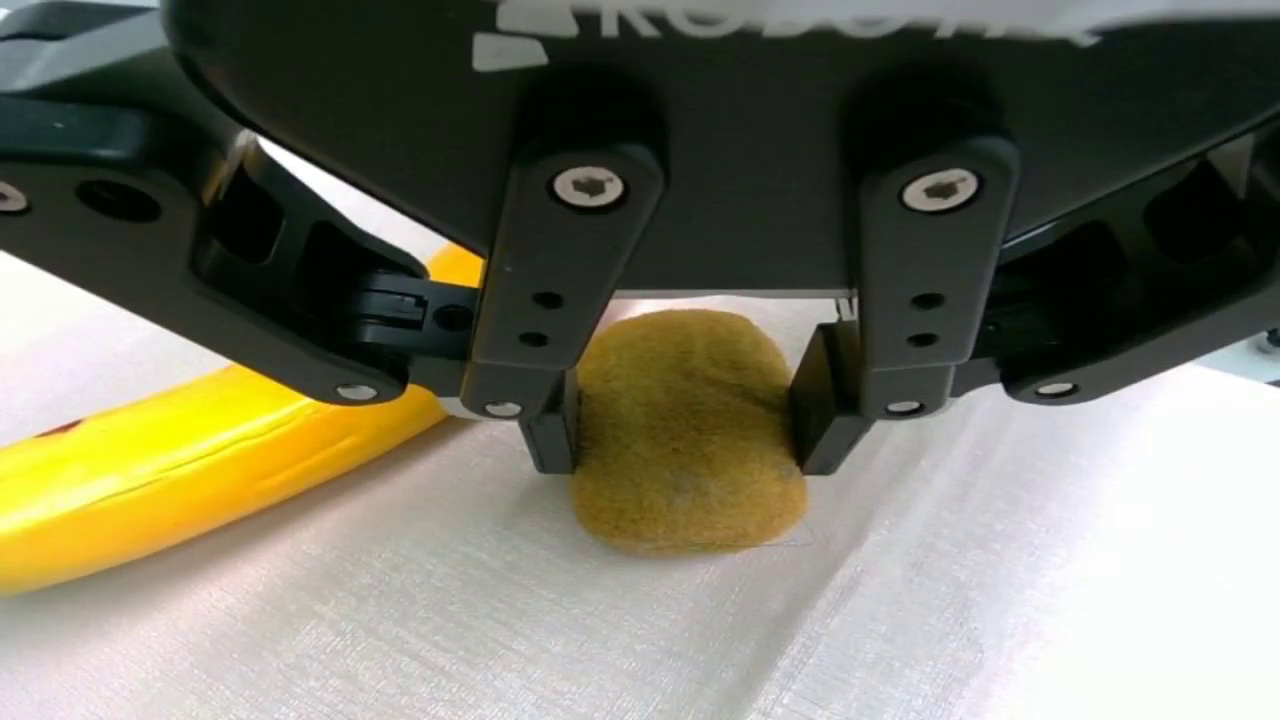}}\\[-2pt]
{\scriptsize \textbf{F671 / 67.1s}\strut}\\[-1pt]
{\scriptsize \textbf{Recovery $r$}\strut}\end{minipage}\hfill%
\begin{minipage}[t]{0.137\linewidth}\centering
\setlength{\fboxsep}{0pt}\setlength{\fboxrule}{0.8pt}
\fcolorbox{white}{white}{\includegraphics[width=\dimexpr\linewidth-1.6pt\relax]{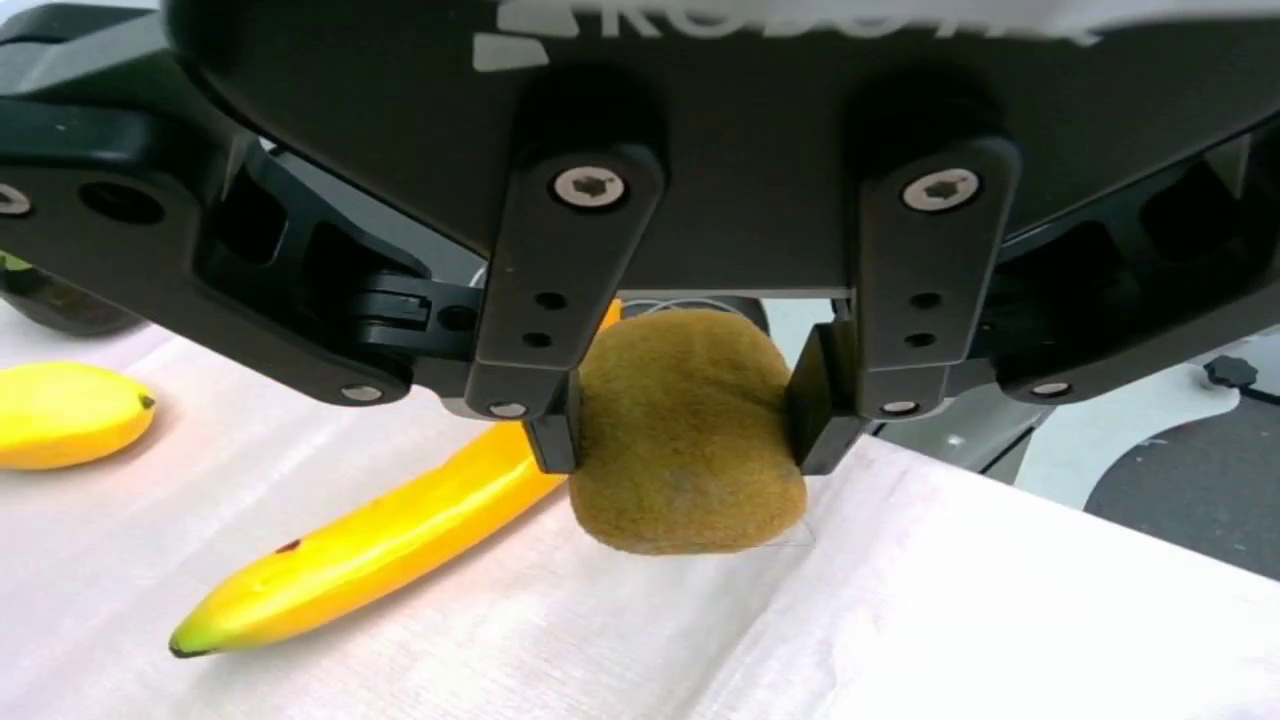}}\\[-2pt]
{\scriptsize F701 / 70.1s\strut}\\[-1pt]
{\scriptsize After recovery\strut}\end{minipage}\hfill%
\begin{minipage}[t]{0.137\linewidth}\centering
\setlength{\fboxsep}{0pt}\setlength{\fboxrule}{0.8pt}
\fcolorbox{endpointgreen}{white}{\includegraphics[width=\dimexpr\linewidth-1.6pt\relax,trim=541.1111bp 0.0000bp 81.1111bp 350.0000bp,clip]{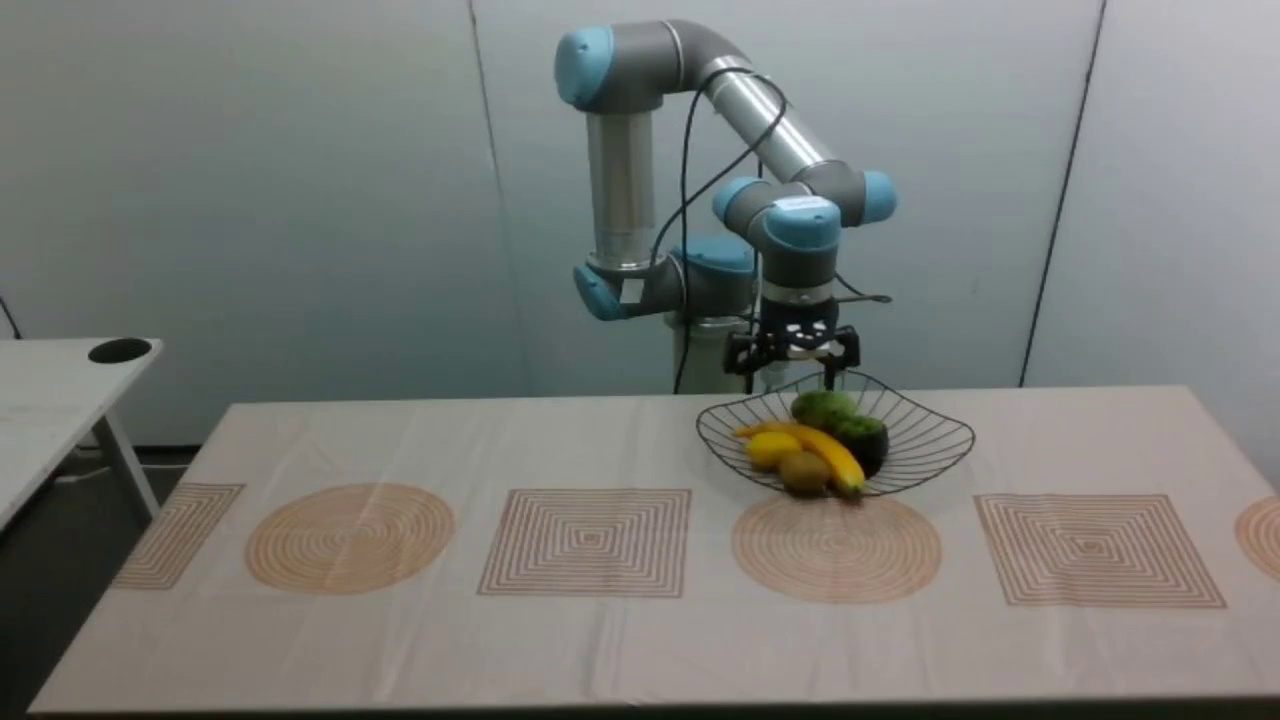}}\\[-2pt]
{\scriptsize \textcolor{endpointgreen}{\textbf{F3717 / 371.7s}}\strut}\\[-1pt]
{\scriptsize \textcolor{endpointgreen}{\textbf{Success / $T$}}\strut}\end{minipage}
\end{minipage}\par
\vfill
\begin{minipage}{\linewidth}
\setlength{\parskip}{\eventExampleParskip}
{\normalsize \textbf{(c)} Place all loose fruits in the rectangular wicker tray. The yellow fruit falls to the table and is regrasped. By the final frame, the fruits are collected in the tray.\par}
\noindent \begin{minipage}[t]{0.137\linewidth}\centering
\setlength{\fboxsep}{0pt}\setlength{\fboxrule}{0.8pt}
\fcolorbox{white}{white}{\includegraphics[width=\dimexpr\linewidth-1.6pt\relax]{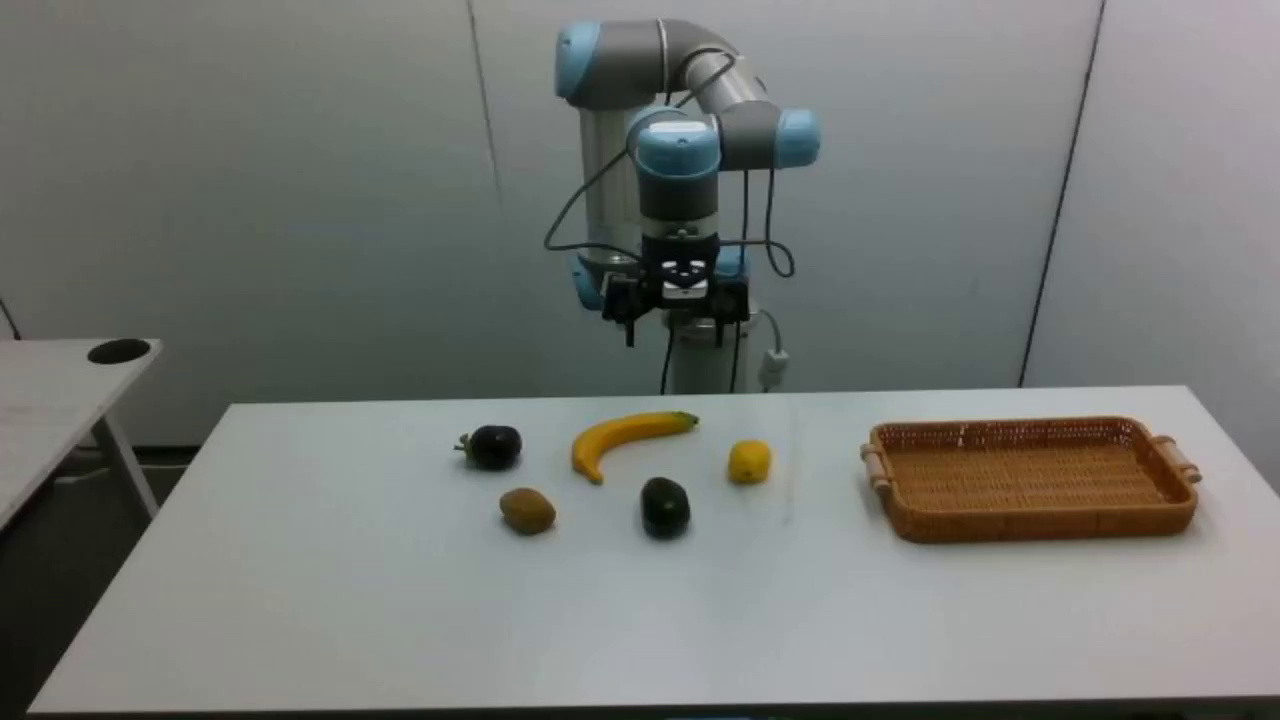}}\\[-2pt]
{\scriptsize F0 / 0.0s\strut}\\[-1pt]
{\scriptsize Context\strut}\end{minipage}\hfill%
\begin{minipage}[t]{0.137\linewidth}\centering
\setlength{\fboxsep}{0pt}\setlength{\fboxrule}{0.8pt}
\fcolorbox{black}{white}{\includegraphics[width=\dimexpr\linewidth-1.6pt\relax]{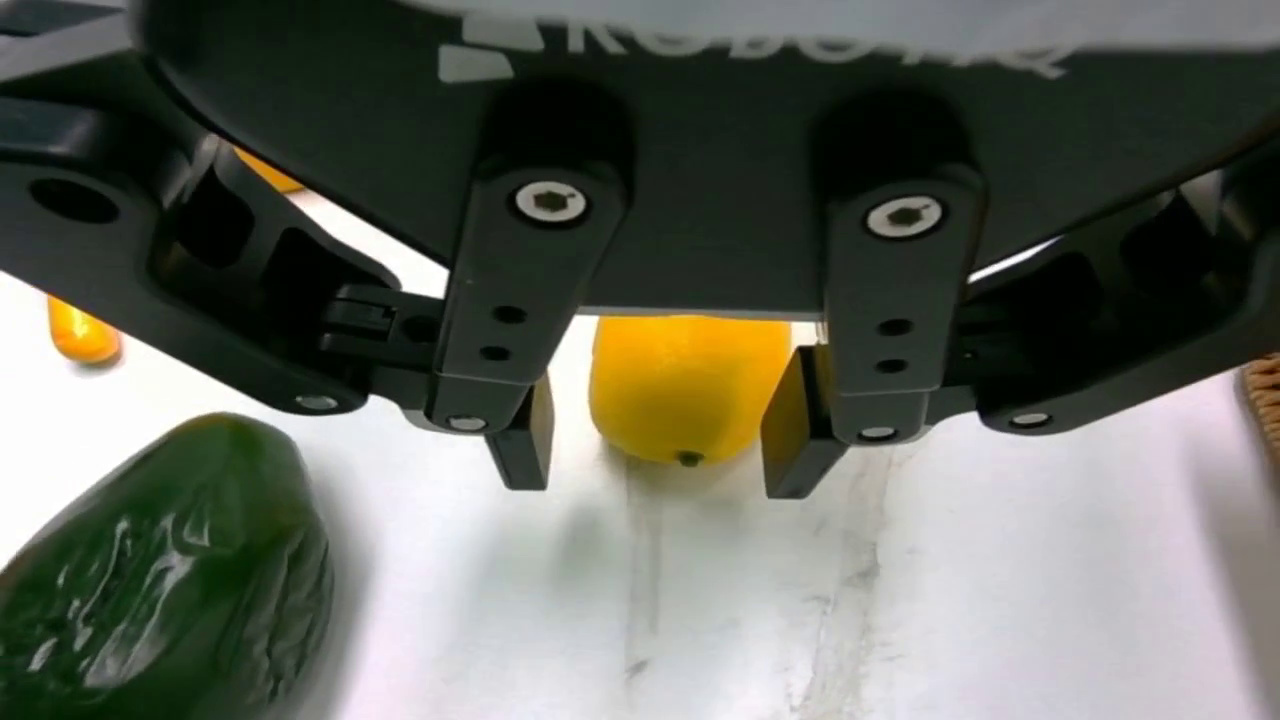}}\\[-2pt]
{\scriptsize \textbf{F185 / 18.5s}\strut}\\[-1pt]
{\scriptsize \textbf{Ref. $a$}\strut}\end{minipage}\hfill%
\begin{minipage}[t]{0.137\linewidth}\centering
\setlength{\fboxsep}{0pt}\setlength{\fboxrule}{0.8pt}
\fcolorbox{white}{white}{\includegraphics[width=\dimexpr\linewidth-1.6pt\relax]{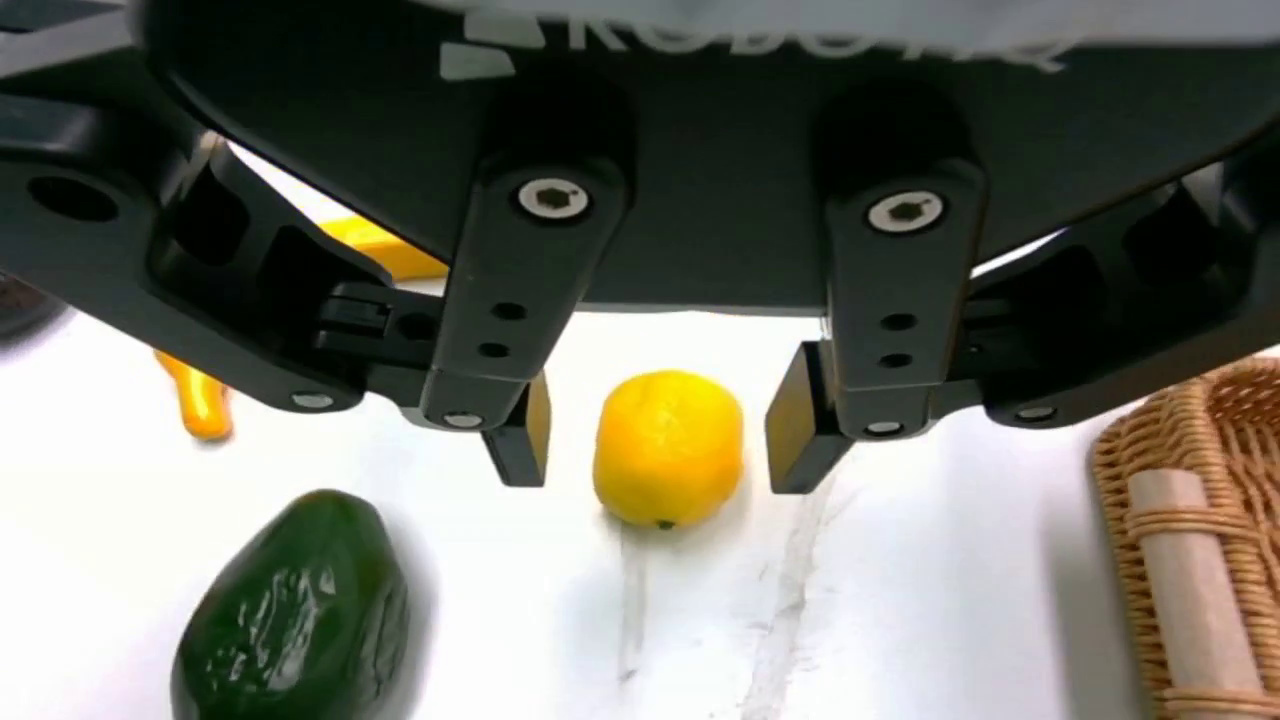}}\\[-2pt]
{\scriptsize F197 / 19.7s\strut}\\[-1pt]
{\scriptsize Loss evidence\strut}\end{minipage}\hfill%
\begin{minipage}[t]{0.137\linewidth}\centering
\setlength{\fboxsep}{0pt}\setlength{\fboxrule}{0.8pt}
\fcolorbox{white}{white}{\includegraphics[width=\dimexpr\linewidth-1.6pt\relax]{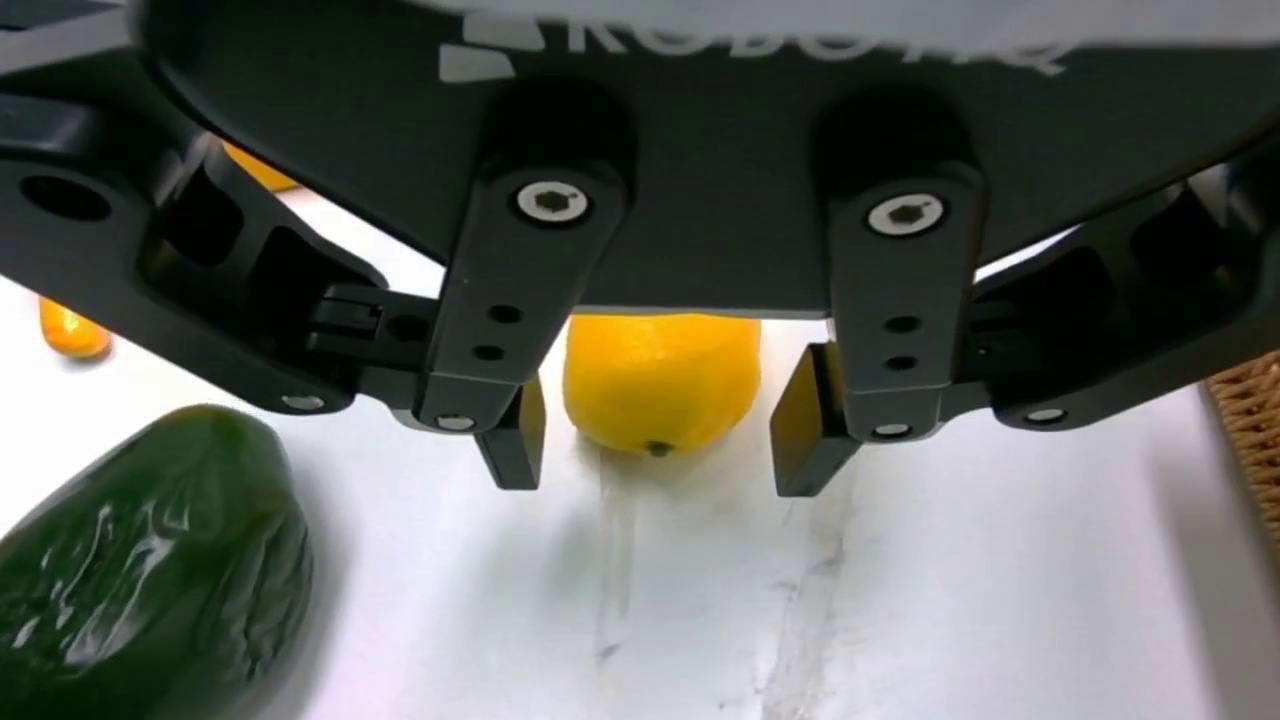}}\\[-2pt]
{\scriptsize F234 / 23.4s\strut}\\[-1pt]
{\scriptsize Reapproach\strut}\end{minipage}\hfill%
\begin{minipage}[t]{0.137\linewidth}\centering
\setlength{\fboxsep}{0pt}\setlength{\fboxrule}{0.8pt}
\fcolorbox{black}{white}{\includegraphics[width=\dimexpr\linewidth-1.6pt\relax]{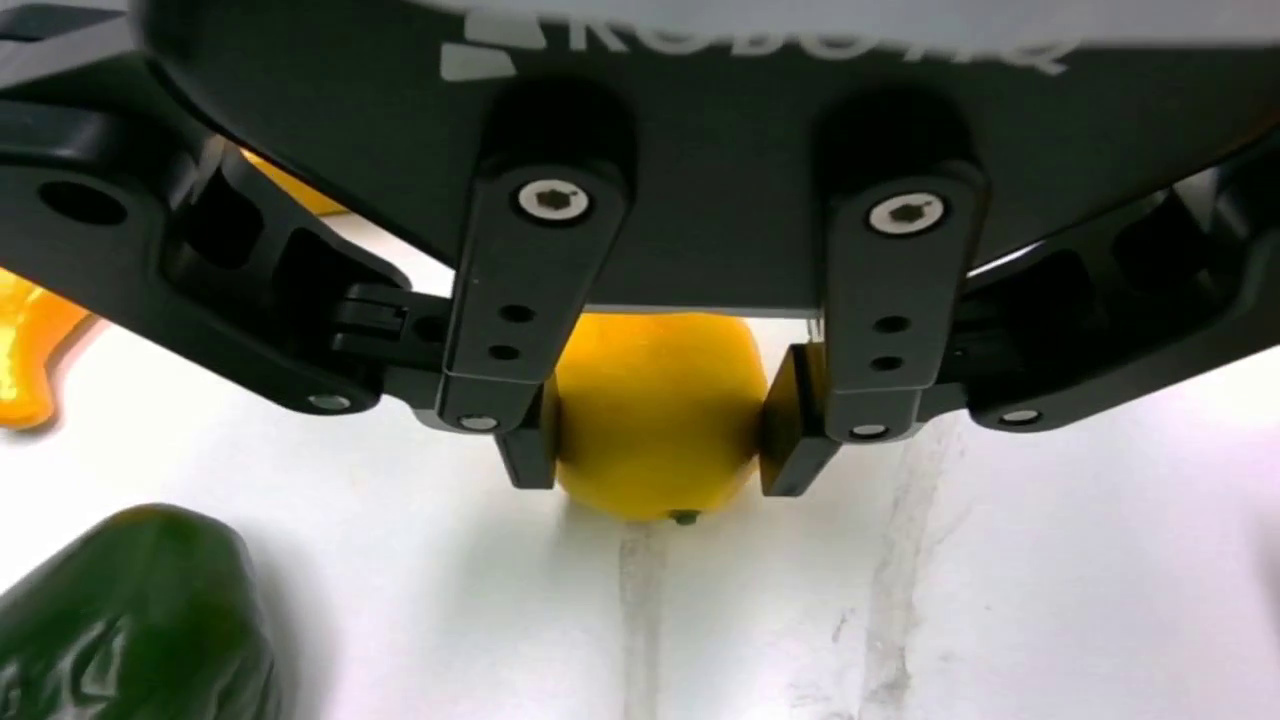}}\\[-2pt]
{\scriptsize \textbf{F246 / 24.6s}\strut}\\[-1pt]
{\scriptsize \textbf{Recovery $r$}\strut}\end{minipage}\hfill%
\begin{minipage}[t]{0.137\linewidth}\centering
\setlength{\fboxsep}{0pt}\setlength{\fboxrule}{0.8pt}
\fcolorbox{white}{white}{\includegraphics[width=\dimexpr\linewidth-1.6pt\relax]{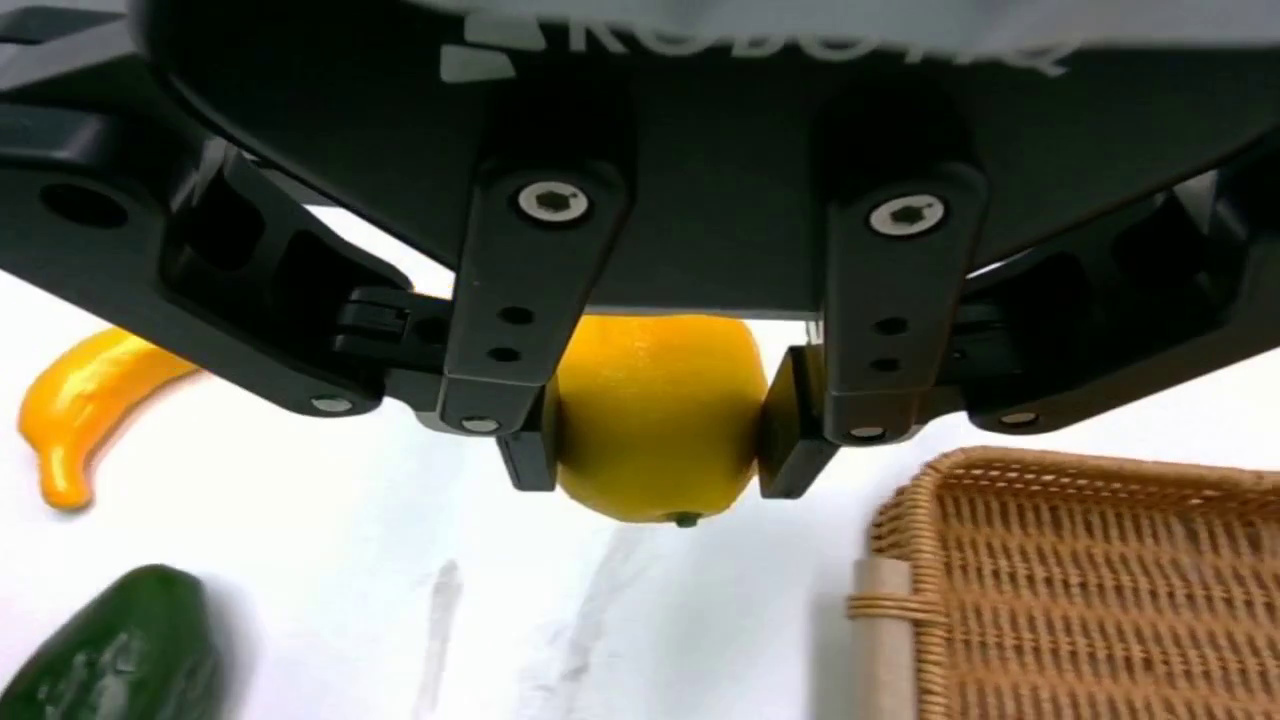}}\\[-2pt]
{\scriptsize F261 / 26.1s\strut}\\[-1pt]
{\scriptsize After recovery\strut}\end{minipage}\hfill%
\begin{minipage}[t]{0.137\linewidth}\centering
\setlength{\fboxsep}{0pt}\setlength{\fboxrule}{0.8pt}
\fcolorbox{endpointgreen}{white}{\includegraphics[width=\dimexpr\linewidth-1.6pt\relax,trim=640.0000bp 0.0000bp 0.0000bp 360.0000bp,clip]{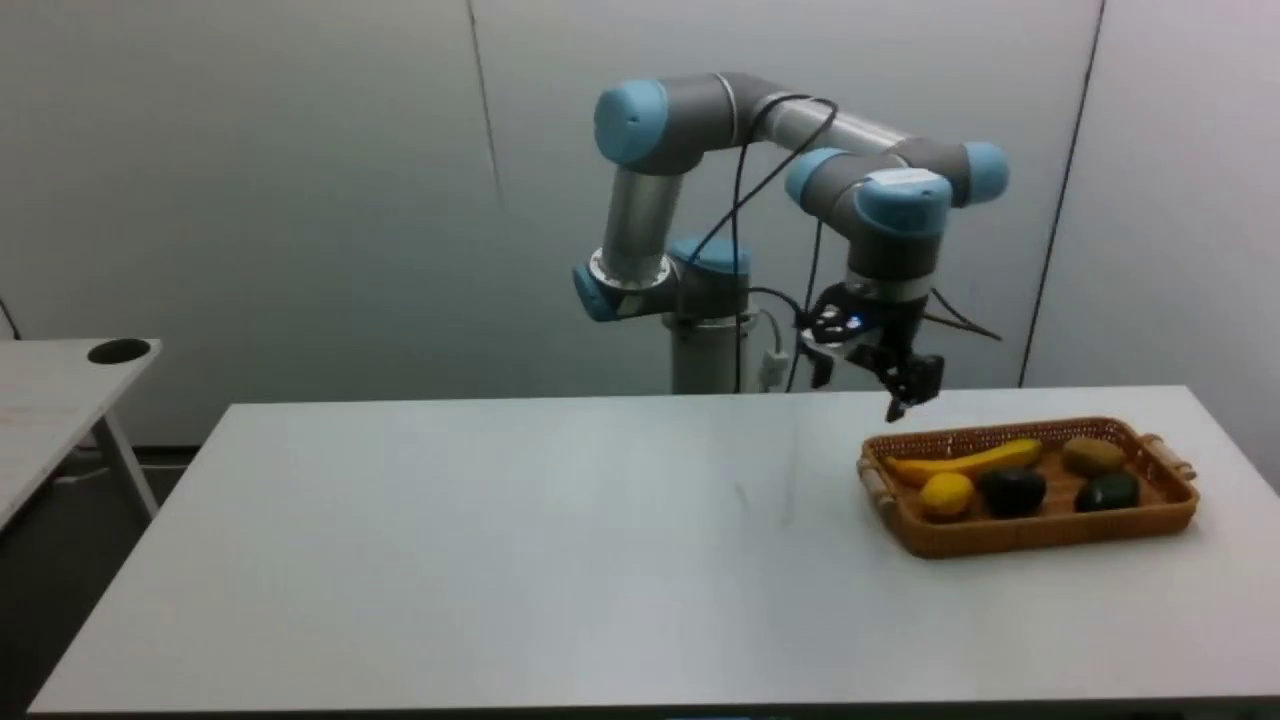}}\\[-2pt]
{\scriptsize \textcolor{endpointgreen}{\textbf{F1527 / 152.7s}}\strut}\\[-1pt]
{\scriptsize \textcolor{endpointgreen}{\textbf{Success / $T$}}\strut}\end{minipage}
\end{minipage}\par
\vfill
\begin{minipage}{\linewidth}
\setlength{\parskip}{\eventExampleParskip}
{\normalsize \textbf{(d)} Put the banana into the cup. The lifted banana drops onto the counter, is grasped and lifted again, and is released into the cup.\par}
\noindent \begin{minipage}[t]{0.137\linewidth}\centering
\setlength{\fboxsep}{0pt}\setlength{\fboxrule}{0.8pt}
\fcolorbox{white}{white}{\includegraphics[width=\dimexpr\linewidth-1.6pt\relax,trim=132.2222bp 18.0000bp 24.2222bp 70.0000bp,clip]{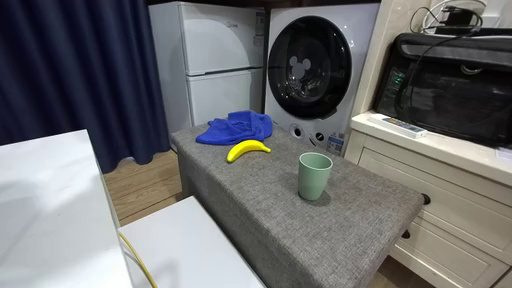}}\\[-2pt]
{\scriptsize F0 / 0.0s\strut}\\[-1pt]
{\scriptsize Context\strut}\end{minipage}\hfill%
\begin{minipage}[t]{0.137\linewidth}\centering
\setlength{\fboxsep}{0pt}\setlength{\fboxrule}{0.8pt}
\fcolorbox{black}{white}{\includegraphics[width=\dimexpr\linewidth-1.6pt\relax,trim=132.2222bp 18.0000bp 24.2222bp 70.0000bp,clip]{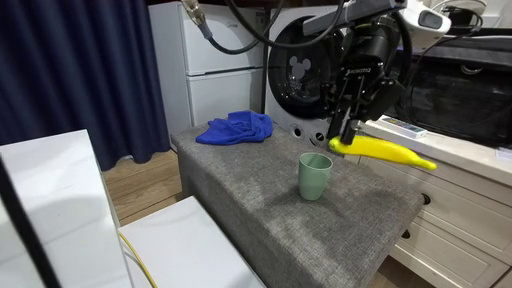}}\\[-2pt]
{\scriptsize \textbf{F153 / 15.3s}\strut}\\[-1pt]
{\scriptsize \textbf{Ref. $a$}\strut}\end{minipage}\hfill%
\begin{minipage}[t]{0.137\linewidth}\centering
\setlength{\fboxsep}{0pt}\setlength{\fboxrule}{0.8pt}
\fcolorbox{white}{white}{\includegraphics[width=\dimexpr\linewidth-1.6pt\relax,trim=132.2222bp 18.0000bp 24.2222bp 70.0000bp,clip]{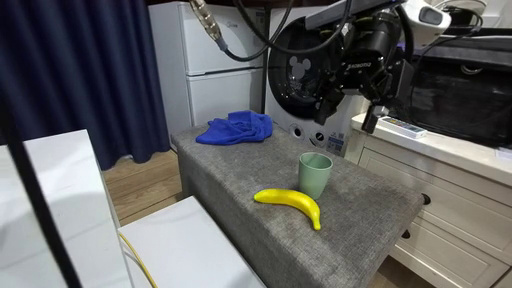}}\\[-2pt]
{\scriptsize F165 / 16.5s\strut}\\[-1pt]
{\scriptsize Loss evidence\strut}\end{minipage}\hfill%
\begin{minipage}[t]{0.137\linewidth}\centering
\setlength{\fboxsep}{0pt}\setlength{\fboxrule}{0.8pt}
\fcolorbox{white}{white}{\includegraphics[width=\dimexpr\linewidth-1.6pt\relax,trim=132.2222bp 18.0000bp 24.2222bp 70.0000bp,clip]{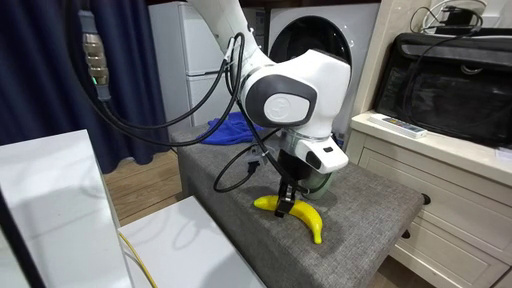}}\\[-2pt]
{\scriptsize F241 / 24.1s\strut}\\[-1pt]
{\scriptsize Reapproach\strut}\end{minipage}\hfill%
\begin{minipage}[t]{0.137\linewidth}\centering
\setlength{\fboxsep}{0pt}\setlength{\fboxrule}{0.8pt}
\fcolorbox{black}{white}{\includegraphics[width=\dimexpr\linewidth-1.6pt\relax,trim=132.2222bp 18.0000bp 24.2222bp 70.0000bp,clip]{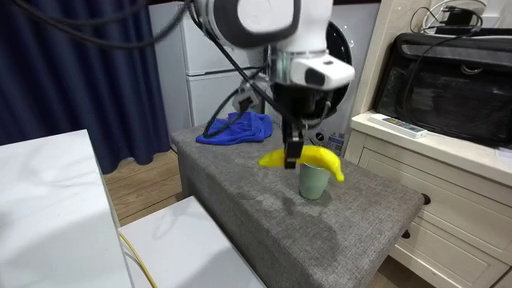}}\\[-2pt]
{\scriptsize \textbf{F259 / 25.9s}\strut}\\[-1pt]
{\scriptsize \textbf{Recovery $r$}\strut}\end{minipage}\hfill%
\begin{minipage}[t]{0.137\linewidth}\centering
\setlength{\fboxsep}{0pt}\setlength{\fboxrule}{0.8pt}
\fcolorbox{white}{white}{\includegraphics[width=\dimexpr\linewidth-1.6pt\relax,trim=132.2222bp 18.0000bp 24.2222bp 70.0000bp,clip]{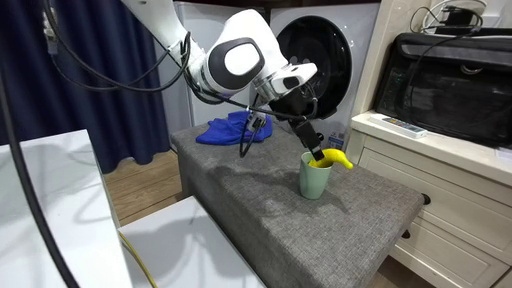}}\\[-2pt]
{\scriptsize F335 / 33.5s\strut}\\[-1pt]
{\scriptsize After recovery\strut}\end{minipage}\hfill%
\begin{minipage}[t]{0.137\linewidth}\centering
\setlength{\fboxsep}{0pt}\setlength{\fboxrule}{0.8pt}
\fcolorbox{endpointgreen}{white}{\includegraphics[width=\dimexpr\linewidth-1.6pt\relax,trim=132.2222bp 18.0000bp 24.2222bp 70.0000bp,clip]{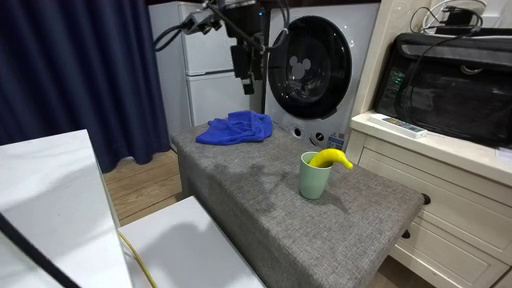}}\\[-2pt]
{\scriptsize \textcolor{endpointgreen}{\textbf{F400 / 40.0s}}\strut}\\[-1pt]
{\scriptsize \textcolor{endpointgreen}{\textbf{Success / $T$}}\strut}\end{minipage}
\end{minipage}\par
\vspace{14pt}
{\scriptsize Source: RoboChallenge (a--c), RoboArena (d). Examples use final annotation records from the procedure in Appendix~\ref{app:annotation-procedure}. Black borders mark annotated event anchors. Green/red endpoints indicate success/failure. Frame numbers and times identify positions in each video. Some views are cropped to highlight task-relevant regions.\par}
\egroup
\endgroup
\clearpage

\begingroup
\setlength{\parskip}{\eventExampleParskip}
\vbox to \textheight\bgroup
\hsize=\linewidth
\section*{Real robots: No further progress}
{\normalsize No later state advances beyond the annotated reference before recording ends ($p_t\leq p_a$). Images around $a$ show the last advancement boundary.\par}
\begin{minipage}{\linewidth}
\setlength{\parskip}{\eventExampleParskip}
{\normalsize \textbf{(a)} Place the tape roll inside the drawer. The roll is lifted to the drawer opening, then remains gripped outside without insertion or release.\par}
\noindent \begin{minipage}[t]{0.137\linewidth}\centering
\setlength{\fboxsep}{0pt}\setlength{\fboxrule}{0.8pt}
\fcolorbox{white}{white}{\includegraphics[width=\dimexpr\linewidth-1.6pt\relax,trim=105.0000bp 51.7500bp 127.0000bp 78.7500bp,clip]{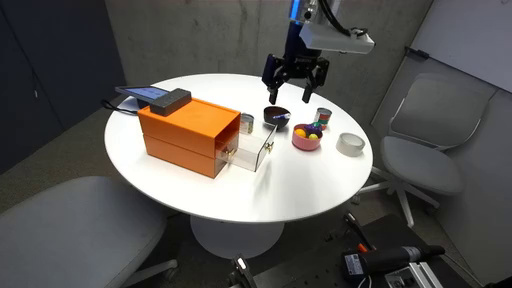}}\\[-2pt]
{\scriptsize F0 / 0.0s\strut}\\[-1pt]
{\scriptsize Start\strut}\end{minipage}\hfill%
\begin{minipage}[t]{0.137\linewidth}\centering
\setlength{\fboxsep}{0pt}\setlength{\fboxrule}{0.8pt}
\fcolorbox{white}{white}{\includegraphics[width=\dimexpr\linewidth-1.6pt\relax]{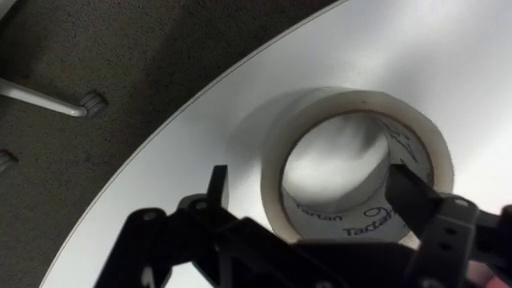}}\\[-2pt]
{\scriptsize F120 / 12.0s\strut}\\[-1pt]
{\scriptsize Before\strut}\end{minipage}\hfill%
\begin{minipage}[t]{0.137\linewidth}\centering
\setlength{\fboxsep}{0pt}\setlength{\fboxrule}{0.8pt}
\fcolorbox{white}{white}{\includegraphics[width=\dimexpr\linewidth-1.6pt\relax]{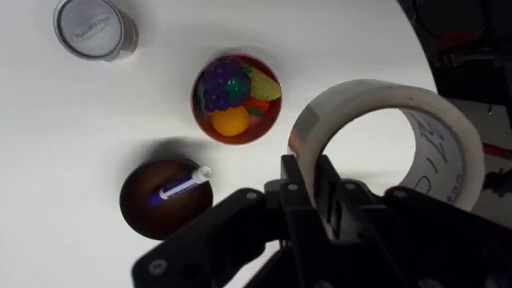}}\\[-2pt]
{\scriptsize F164 / 16.4s\strut}\\[-1pt]
{\scriptsize Before $a$\strut}\end{minipage}\hfill%
\begin{minipage}[t]{0.137\linewidth}\centering
\setlength{\fboxsep}{0pt}\setlength{\fboxrule}{0.8pt}
\fcolorbox{black}{white}{\includegraphics[width=\dimexpr\linewidth-1.6pt\relax]{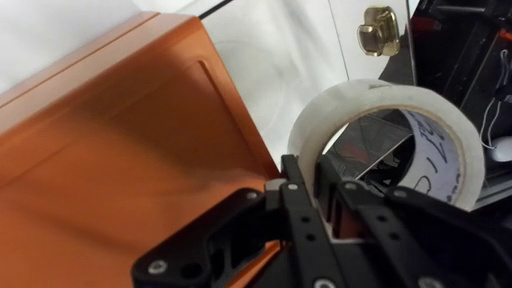}}\\[-2pt]
{\scriptsize \textbf{F214 / 21.4s}\strut}\\[-1pt]
{\scriptsize \textbf{Ref. $a$}\strut}\end{minipage}\hfill%
\begin{minipage}[t]{0.137\linewidth}\centering
\setlength{\fboxsep}{0pt}\setlength{\fboxrule}{0.8pt}
\fcolorbox{white}{white}{\includegraphics[width=\dimexpr\linewidth-1.6pt\relax]{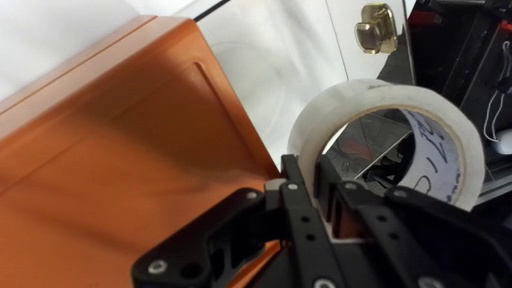}}\\[-2pt]
{\scriptsize F250 / 25.0s\strut}\\[-1pt]
{\scriptsize Plateau\strut}\end{minipage}\hfill%
\begin{minipage}[t]{0.137\linewidth}\centering
\setlength{\fboxsep}{0pt}\setlength{\fboxrule}{0.8pt}
\fcolorbox{white}{white}{\includegraphics[width=\dimexpr\linewidth-1.6pt\relax]{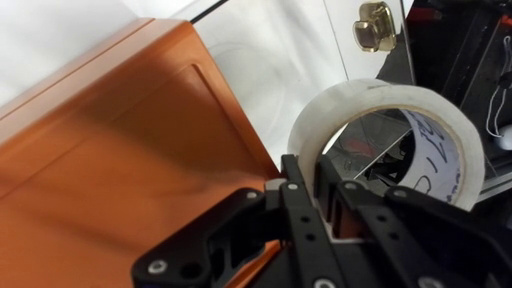}}\\[-2pt]
{\scriptsize F350 / 35.0s\strut}\\[-1pt]
{\scriptsize Plateau\strut}\end{minipage}\hfill%
\begin{minipage}[t]{0.137\linewidth}\centering
\setlength{\fboxsep}{0pt}\setlength{\fboxrule}{0.8pt}
\fcolorbox{endpointred}{white}{\includegraphics[width=\dimexpr\linewidth-1.6pt\relax,trim=105.0000bp 51.7500bp 127.0000bp 78.7500bp,clip]{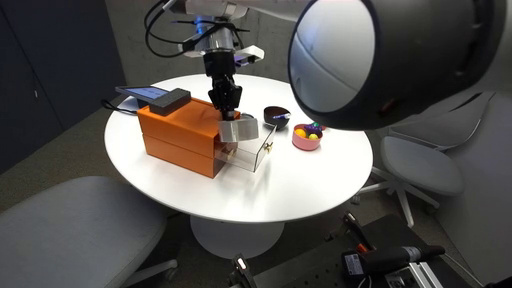}}\\[-2pt]
{\scriptsize \textcolor{endpointred}{\textbf{F400 / 40.0s}}\strut}\\[-1pt]
{\scriptsize \textcolor{endpointred}{\textbf{Failure / $T$}}\strut}\end{minipage}
\end{minipage}\par
\vfill
\begin{minipage}{\linewidth}
\setlength{\parskip}{\eventExampleParskip}
{\normalsize \textbf{(b)} Place the red cup inside the cabinet. The cup leaves the table and reaches the cabinet, then remains gripped outside through T.\par}
\noindent \begin{minipage}[t]{0.137\linewidth}\centering
\setlength{\fboxsep}{0pt}\setlength{\fboxrule}{0.8pt}
\fcolorbox{white}{white}{\includegraphics[width=\dimexpr\linewidth-1.6pt\relax,trim=95.5556bp 13.0000bp 7.5556bp 45.0000bp,clip]{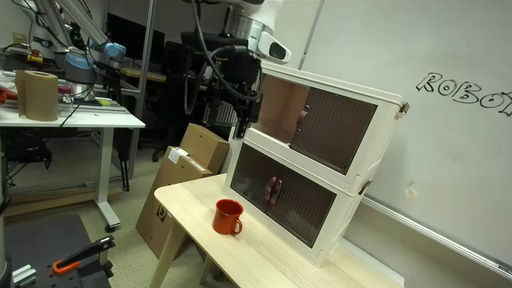}}\\[-2pt]
{\scriptsize F0 / 0.0s\strut}\\[-1pt]
{\scriptsize Start\strut}\end{minipage}\hfill%
\begin{minipage}[t]{0.137\linewidth}\centering
\setlength{\fboxsep}{0pt}\setlength{\fboxrule}{0.8pt}
\fcolorbox{white}{white}{\includegraphics[width=\dimexpr\linewidth-1.6pt\relax]{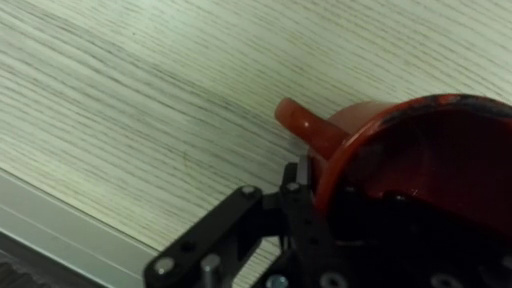}}\\[-2pt]
{\scriptsize F50 / 5.0s\strut}\\[-1pt]
{\scriptsize Before\strut}\end{minipage}\hfill%
\begin{minipage}[t]{0.137\linewidth}\centering
\setlength{\fboxsep}{0pt}\setlength{\fboxrule}{0.8pt}
\fcolorbox{white}{white}{\includegraphics[width=\dimexpr\linewidth-1.6pt\relax]{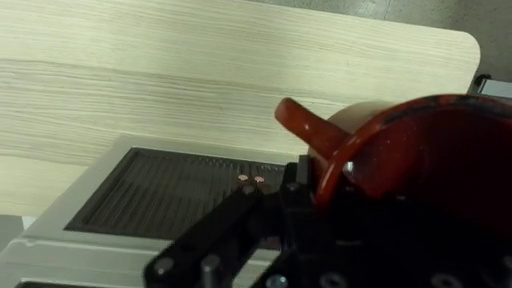}}\\[-2pt]
{\scriptsize F80 / 8.0s\strut}\\[-1pt]
{\scriptsize Before $a$\strut}\end{minipage}\hfill%
\begin{minipage}[t]{0.137\linewidth}\centering
\setlength{\fboxsep}{0pt}\setlength{\fboxrule}{0.8pt}
\fcolorbox{black}{white}{\includegraphics[width=\dimexpr\linewidth-1.6pt\relax]{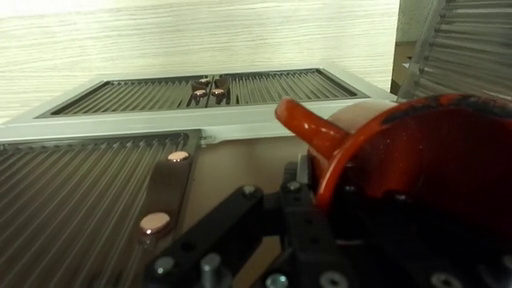}}\\[-2pt]
{\scriptsize \textbf{F110 / 11.0s}\strut}\\[-1pt]
{\scriptsize \textbf{Ref. $a$}\strut}\end{minipage}\hfill%
\begin{minipage}[t]{0.137\linewidth}\centering
\setlength{\fboxsep}{0pt}\setlength{\fboxrule}{0.8pt}
\fcolorbox{white}{white}{\includegraphics[width=\dimexpr\linewidth-1.6pt\relax]{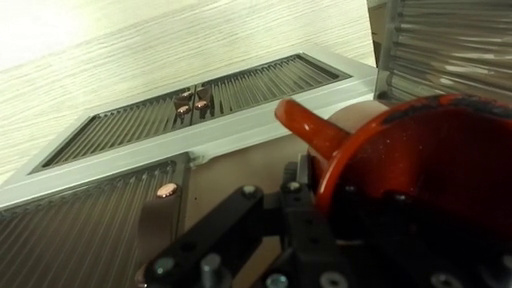}}\\[-2pt]
{\scriptsize F150 / 15.0s\strut}\\[-1pt]
{\scriptsize Plateau\strut}\end{minipage}\hfill%
\begin{minipage}[t]{0.137\linewidth}\centering
\setlength{\fboxsep}{0pt}\setlength{\fboxrule}{0.8pt}
\fcolorbox{white}{white}{\includegraphics[width=\dimexpr\linewidth-1.6pt\relax]{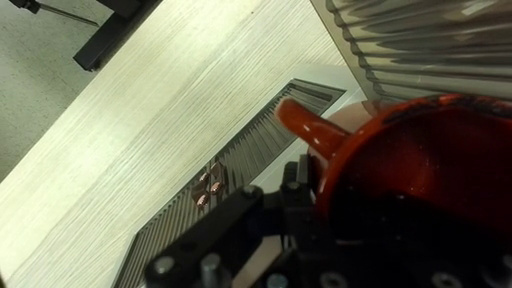}}\\[-2pt]
{\scriptsize F215 / 21.5s\strut}\\[-1pt]
{\scriptsize Plateau\strut}\end{minipage}\hfill%
\begin{minipage}[t]{0.137\linewidth}\centering
\setlength{\fboxsep}{0pt}\setlength{\fboxrule}{0.8pt}
\fcolorbox{endpointred}{white}{\includegraphics[width=\dimexpr\linewidth-1.6pt\relax,trim=95.5556bp 13.0000bp 7.5556bp 45.0000bp,clip]{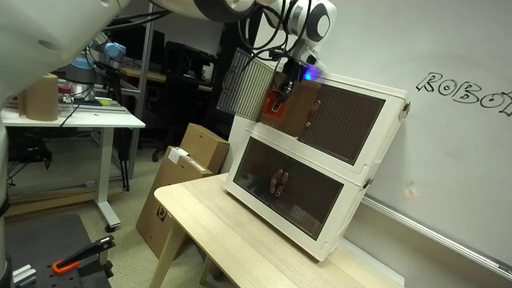}}\\[-2pt]
{\scriptsize \textcolor{endpointred}{\textbf{F265 / 26.5s}}\strut}\\[-1pt]
{\scriptsize \textcolor{endpointred}{\textbf{Failure / $T$}}\strut}\end{minipage}
\end{minipage}\par
\vfill
\begin{minipage}{\linewidth}
\setlength{\parskip}{\eventExampleParskip}
{\normalsize \textbf{(c)} Place the cup inside the white storage box. The cup is lifted, then stays held outside the white box without placement through T.\par}
\noindent \begin{minipage}[t]{0.137\linewidth}\centering
\setlength{\fboxsep}{0pt}\setlength{\fboxrule}{0.8pt}
\fcolorbox{white}{white}{\includegraphics[width=\dimexpr\linewidth-1.6pt\relax,trim=0.0000bp 33.0000bp 165.3333bp 60.0000bp,clip]{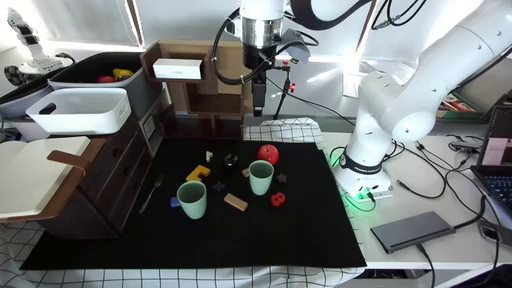}}\\[-2pt]
{\scriptsize F0 / 0.0s\strut}\\[-1pt]
{\scriptsize Start\strut}\end{minipage}\hfill%
\begin{minipage}[t]{0.137\linewidth}\centering
\setlength{\fboxsep}{0pt}\setlength{\fboxrule}{0.8pt}
\fcolorbox{white}{white}{\includegraphics[width=\dimexpr\linewidth-1.6pt\relax]{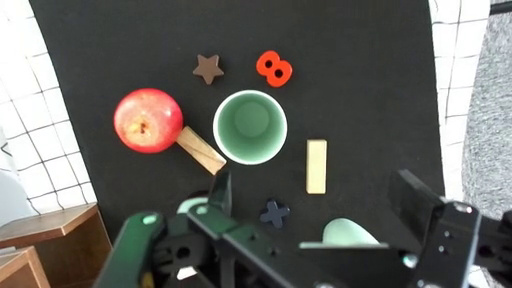}}\\[-2pt]
{\scriptsize F250 / 25.0s\strut}\\[-1pt]
{\scriptsize Before\strut}\end{minipage}\hfill%
\begin{minipage}[t]{0.137\linewidth}\centering
\setlength{\fboxsep}{0pt}\setlength{\fboxrule}{0.8pt}
\fcolorbox{white}{white}{\includegraphics[width=\dimexpr\linewidth-1.6pt\relax]{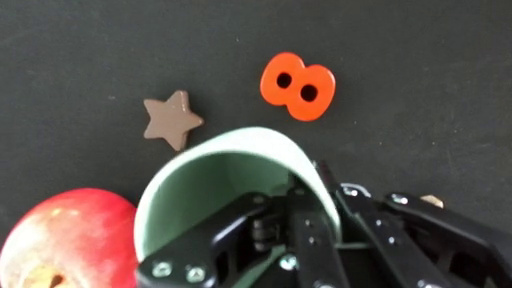}}\\[-2pt]
{\scriptsize F300 / 30.0s\strut}\\[-1pt]
{\scriptsize Before $a$\strut}\end{minipage}\hfill%
\begin{minipage}[t]{0.137\linewidth}\centering
\setlength{\fboxsep}{0pt}\setlength{\fboxrule}{0.8pt}
\fcolorbox{black}{white}{\includegraphics[width=\dimexpr\linewidth-1.6pt\relax]{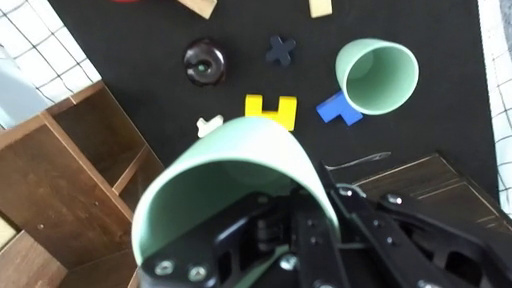}}\\[-2pt]
{\scriptsize \textbf{F330 / 33.0s}\strut}\\[-1pt]
{\scriptsize \textbf{Ref. $a$}\strut}\end{minipage}\hfill%
\begin{minipage}[t]{0.137\linewidth}\centering
\setlength{\fboxsep}{0pt}\setlength{\fboxrule}{0.8pt}
\fcolorbox{white}{white}{\includegraphics[width=\dimexpr\linewidth-1.6pt\relax]{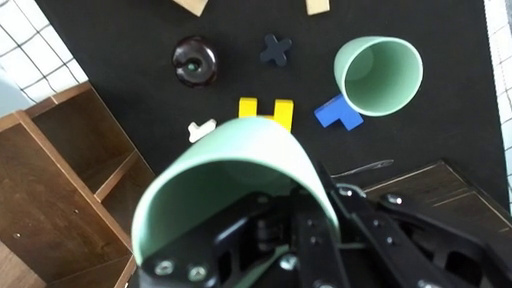}}\\[-2pt]
{\scriptsize F350 / 35.0s\strut}\\[-1pt]
{\scriptsize Plateau\strut}\end{minipage}\hfill%
\begin{minipage}[t]{0.137\linewidth}\centering
\setlength{\fboxsep}{0pt}\setlength{\fboxrule}{0.8pt}
\fcolorbox{white}{white}{\includegraphics[width=\dimexpr\linewidth-1.6pt\relax]{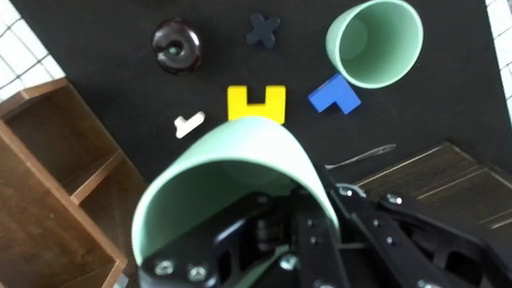}}\\[-2pt]
{\scriptsize F375 / 37.5s\strut}\\[-1pt]
{\scriptsize Plateau\strut}\end{minipage}\hfill%
\begin{minipage}[t]{0.137\linewidth}\centering
\setlength{\fboxsep}{0pt}\setlength{\fboxrule}{0.8pt}
\fcolorbox{endpointred}{white}{\includegraphics[width=\dimexpr\linewidth-1.6pt\relax,trim=0.0000bp 33.0000bp 165.3333bp 60.0000bp,clip]{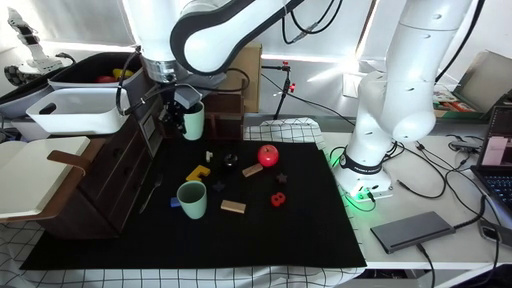}}\\[-2pt]
{\scriptsize \textcolor{endpointred}{\textbf{F400 / 40.0s}}\strut}\\[-1pt]
{\scriptsize \textcolor{endpointred}{\textbf{Failure / $T$}}\strut}\end{minipage}
\end{minipage}\par
\vfill
\begin{minipage}{\linewidth}
\setlength{\parskip}{\eventExampleParskip}
{\normalsize \textbf{(d)} Place the plush toy inside the fabric bag. The toy is lifted toward the bag but remains gripped outside it while the arm repositions it.\par}
\noindent \begin{minipage}[t]{0.137\linewidth}\centering
\setlength{\fboxsep}{0pt}\setlength{\fboxrule}{0.8pt}
\fcolorbox{white}{white}{\includegraphics[width=\dimexpr\linewidth-1.6pt\relax,trim=85.0000bp 108.0000bp 107.0000bp 0.0000bp,clip]{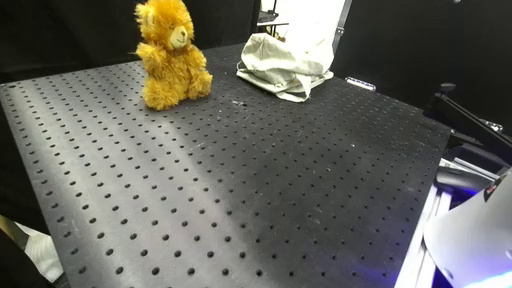}}\\[-2pt]
{\scriptsize F0 / 0.0s\strut}\\[-1pt]
{\scriptsize Start\strut}\end{minipage}\hfill%
\begin{minipage}[t]{0.137\linewidth}\centering
\setlength{\fboxsep}{0pt}\setlength{\fboxrule}{0.8pt}
\fcolorbox{white}{white}{\includegraphics[width=\dimexpr\linewidth-1.6pt\relax]{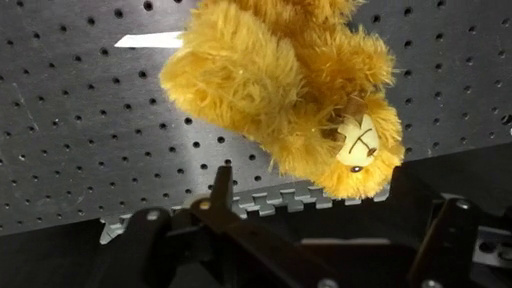}}\\[-2pt]
{\scriptsize F100 / 10.0s\strut}\\[-1pt]
{\scriptsize Before\strut}\end{minipage}\hfill%
\begin{minipage}[t]{0.137\linewidth}\centering
\setlength{\fboxsep}{0pt}\setlength{\fboxrule}{0.8pt}
\fcolorbox{white}{white}{\includegraphics[width=\dimexpr\linewidth-1.6pt\relax]{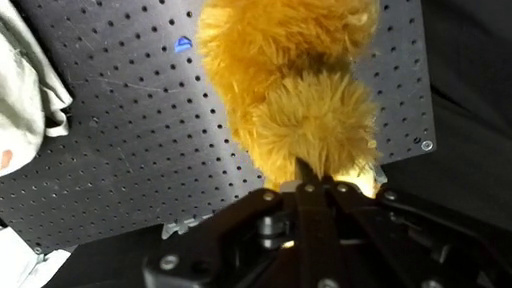}}\\[-2pt]
{\scriptsize F160 / 16.0s\strut}\\[-1pt]
{\scriptsize Before $a$\strut}\end{minipage}\hfill%
\begin{minipage}[t]{0.137\linewidth}\centering
\setlength{\fboxsep}{0pt}\setlength{\fboxrule}{0.8pt}
\fcolorbox{black}{white}{\includegraphics[width=\dimexpr\linewidth-1.6pt\relax]{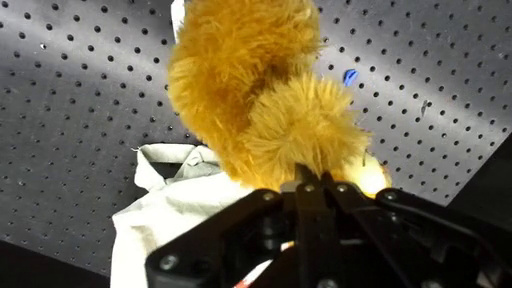}}\\[-2pt]
{\scriptsize \textbf{F200 / 20.0s}\strut}\\[-1pt]
{\scriptsize \textbf{Ref. $a$}\strut}\end{minipage}\hfill%
\begin{minipage}[t]{0.137\linewidth}\centering
\setlength{\fboxsep}{0pt}\setlength{\fboxrule}{0.8pt}
\fcolorbox{white}{white}{\includegraphics[width=\dimexpr\linewidth-1.6pt\relax]{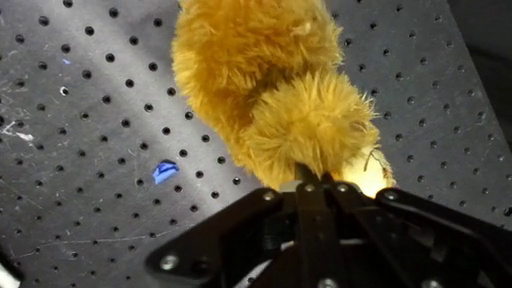}}\\[-2pt]
{\scriptsize F240 / 24.0s\strut}\\[-1pt]
{\scriptsize Plateau\strut}\end{minipage}\hfill%
\begin{minipage}[t]{0.137\linewidth}\centering
\setlength{\fboxsep}{0pt}\setlength{\fboxrule}{0.8pt}
\fcolorbox{white}{white}{\includegraphics[width=\dimexpr\linewidth-1.6pt\relax]{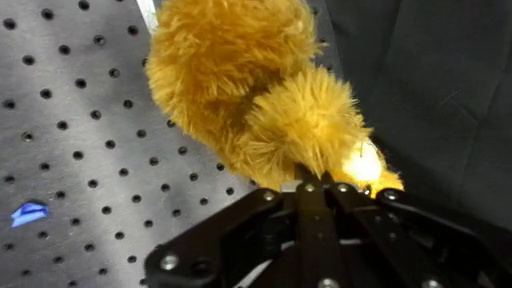}}\\[-2pt]
{\scriptsize F330 / 33.0s\strut}\\[-1pt]
{\scriptsize Plateau\strut}\end{minipage}\hfill%
\begin{minipage}[t]{0.137\linewidth}\centering
\setlength{\fboxsep}{0pt}\setlength{\fboxrule}{0.8pt}
\fcolorbox{endpointred}{white}{\includegraphics[width=\dimexpr\linewidth-1.6pt\relax,trim=85.0000bp 108.0000bp 107.0000bp 0.0000bp,clip]{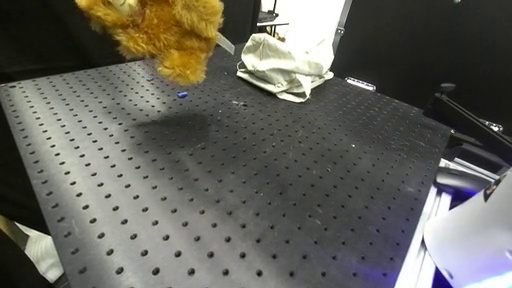}}\\[-2pt]
{\scriptsize \textcolor{endpointred}{\textbf{F400 / 40.0s}}\strut}\\[-1pt]
{\scriptsize \textcolor{endpointred}{\textbf{Failure / $T$}}\strut}\end{minipage}
\end{minipage}\par
\vspace{14pt}
{\scriptsize Source: RoboArena. Examples use final annotation records from the procedure in Appendix~\ref{app:annotation-procedure}. Black borders mark annotated event anchors. Green/red endpoints indicate success/failure. Frame numbers and times identify positions in each video. Some views are cropped to highlight task-relevant regions.\par}
\egroup
\endgroup
\clearpage

\begingroup
\setlength{\parskip}{\eventExampleParskip}
\vbox to \textheight\bgroup
\hsize=\linewidth
\section*{Real robots: Unrecovered regression}
\label{app:real-event-examples-end}
{\normalsize An achieved task state is lost and remains unrecovered at the recording endpoint. Supported post-loss states have lower progress ($p_t<p_a$).\par}
\begin{minipage}{\linewidth}
\setlength{\parskip}{\eventExampleParskip}
{\normalsize \textbf{(a)} Place both shoes inside the box. After one shoe is placed, the second slips from the gripper and stays outside the box at T.\par}
\noindent \begin{minipage}[t]{0.137\linewidth}\centering
\setlength{\fboxsep}{0pt}\setlength{\fboxrule}{0.8pt}
\fcolorbox{white}{white}{\includegraphics[width=\dimexpr\linewidth-1.6pt\relax,trim=35.5556bp 0.0000bp 0.0000bp 20.0000bp,clip]{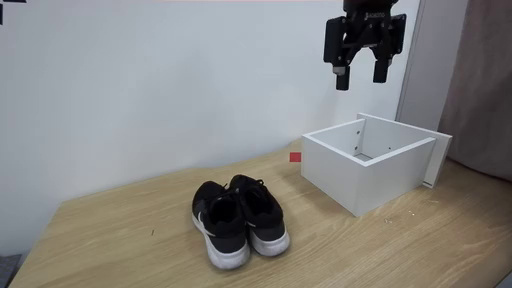}}\\[-2pt]
{\scriptsize F0 / 0.0s\strut}\\[-1pt]
{\scriptsize Start\strut}\end{minipage}\hfill%
\begin{minipage}[t]{0.137\linewidth}\centering
\setlength{\fboxsep}{0pt}\setlength{\fboxrule}{0.8pt}
\fcolorbox{white}{white}{\includegraphics[width=\dimexpr\linewidth-1.6pt\relax,trim=35.5556bp 0.0000bp 0.0000bp 20.0000bp,clip]{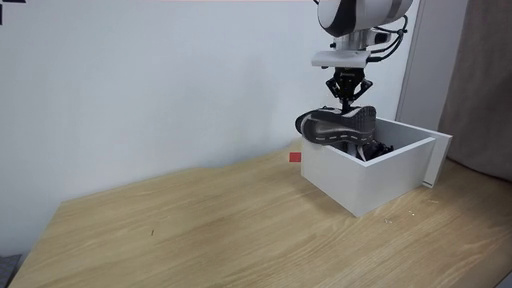}}\\[-2pt]
{\scriptsize F361 / 36.1s\strut}\\[-1pt]
{\scriptsize Before loss\strut}\end{minipage}\hfill%
\begin{minipage}[t]{0.137\linewidth}\centering
\setlength{\fboxsep}{0pt}\setlength{\fboxrule}{0.8pt}
\fcolorbox{black}{white}{\includegraphics[width=\dimexpr\linewidth-1.6pt\relax,trim=35.5556bp 0.0000bp 0.0000bp 20.0000bp,clip]{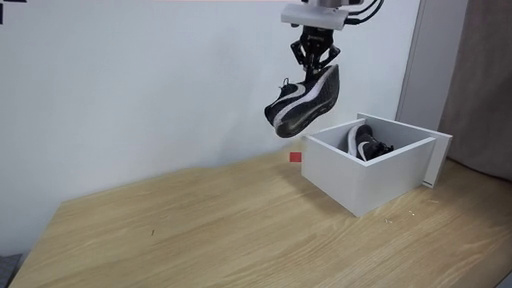}}\\[-2pt]
{\scriptsize \textbf{F373 / 37.3s}\strut}\\[-1pt]
{\scriptsize \textbf{Ref. $a$}\strut}\end{minipage}\hfill%
\begin{minipage}[t]{0.137\linewidth}\centering
\setlength{\fboxsep}{0pt}\setlength{\fboxrule}{0.8pt}
\fcolorbox{white}{white}{\includegraphics[width=\dimexpr\linewidth-1.6pt\relax,trim=35.5556bp 0.0000bp 0.0000bp 20.0000bp,clip]{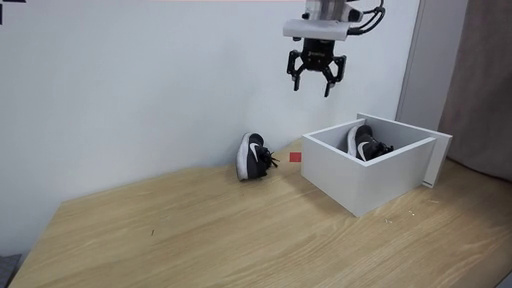}}\\[-2pt]
{\scriptsize F377 / 37.7s\strut}\\[-1pt]
{\scriptsize Loss evidence\strut}\end{minipage}\hfill%
\begin{minipage}[t]{0.137\linewidth}\centering
\setlength{\fboxsep}{0pt}\setlength{\fboxrule}{0.8pt}
\fcolorbox{white}{white}{\includegraphics[width=\dimexpr\linewidth-1.6pt\relax,trim=35.5556bp 0.0000bp 0.0000bp 20.0000bp,clip]{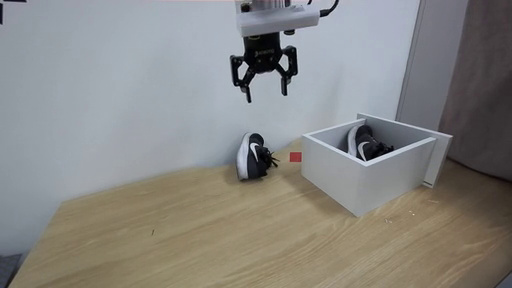}}\\[-2pt]
{\scriptsize F433 / 43.3s\strut}\\[-1pt]
{\scriptsize After loss\strut}\end{minipage}\hfill%
\begin{minipage}[t]{0.137\linewidth}\centering
\setlength{\fboxsep}{0pt}\setlength{\fboxrule}{0.8pt}
\fcolorbox{white}{white}{\includegraphics[width=\dimexpr\linewidth-1.6pt\relax,trim=35.5556bp 0.0000bp 0.0000bp 20.0000bp,clip]{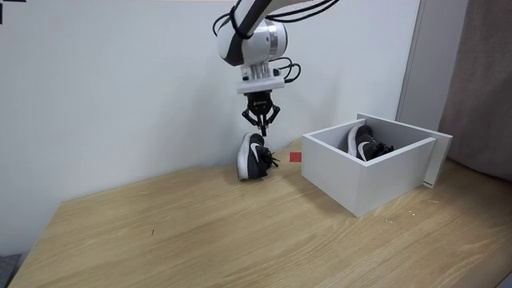}}\\[-2pt]
{\scriptsize F518 / 51.8s\strut}\\[-1pt]
{\scriptsize Later\strut}\end{minipage}\hfill%
\begin{minipage}[t]{0.137\linewidth}\centering
\setlength{\fboxsep}{0pt}\setlength{\fboxrule}{0.8pt}
\fcolorbox{endpointred}{white}{\includegraphics[width=\dimexpr\linewidth-1.6pt\relax,trim=35.5556bp 0.0000bp 0.0000bp 20.0000bp,clip]{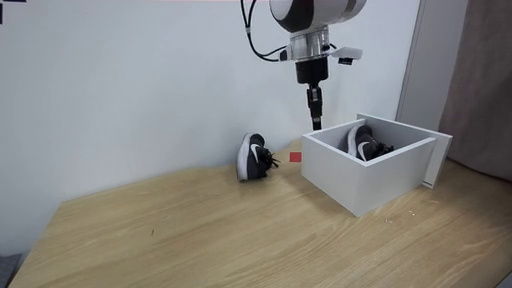}}\\[-2pt]
{\scriptsize \textcolor{endpointred}{\textbf{F799 / 79.9s}}\strut}\\[-1pt]
{\scriptsize \textcolor{endpointred}{\textbf{Failure / $T$}}\strut}\end{minipage}
\end{minipage}\par
\vfill
\begin{minipage}{\linewidth}
\setlength{\parskip}{\eventExampleParskip}
{\normalsize \textbf{(b)} Place the banana on the plate. The banana is lifted, falls beside the empty plate, and remains on the table at T.\par}
\noindent \begin{minipage}[t]{0.137\linewidth}\centering
\setlength{\fboxsep}{0pt}\setlength{\fboxrule}{0.8pt}
\fcolorbox{white}{white}{\includegraphics[width=\dimexpr\linewidth-1.6pt\relax,trim=81.2222bp 0.0000bp 43.2222bp 70.0000bp,clip]{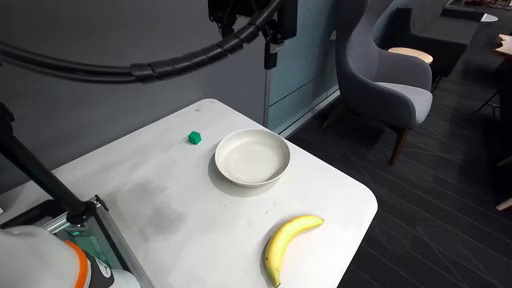}}\\[-2pt]
{\scriptsize F0 / 0.0s\strut}\\[-1pt]
{\scriptsize Start\strut}\end{minipage}\hfill%
\begin{minipage}[t]{0.137\linewidth}\centering
\setlength{\fboxsep}{0pt}\setlength{\fboxrule}{0.8pt}
\fcolorbox{white}{white}{\includegraphics[width=\dimexpr\linewidth-1.6pt\relax,trim=81.2222bp 0.0000bp 43.2222bp 70.0000bp,clip]{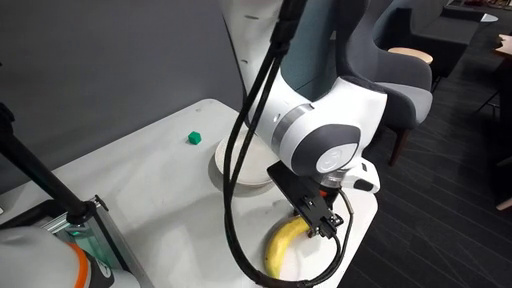}}\\[-2pt]
{\scriptsize F57 / 5.7s\strut}\\[-1pt]
{\scriptsize Before loss\strut}\end{minipage}\hfill%
\begin{minipage}[t]{0.137\linewidth}\centering
\setlength{\fboxsep}{0pt}\setlength{\fboxrule}{0.8pt}
\fcolorbox{black}{white}{\includegraphics[width=\dimexpr\linewidth-1.6pt\relax,trim=81.2222bp 0.0000bp 43.2222bp 70.0000bp,clip]{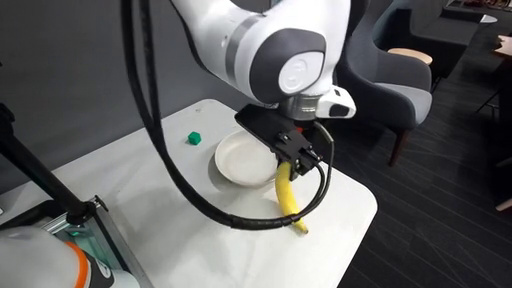}}\\[-2pt]
{\scriptsize \textbf{F69 / 6.9s}\strut}\\[-1pt]
{\scriptsize \textbf{Ref. $a$}\strut}\end{minipage}\hfill%
\begin{minipage}[t]{0.137\linewidth}\centering
\setlength{\fboxsep}{0pt}\setlength{\fboxrule}{0.8pt}
\fcolorbox{white}{white}{\includegraphics[width=\dimexpr\linewidth-1.6pt\relax,trim=81.2222bp 0.0000bp 43.2222bp 70.0000bp,clip]{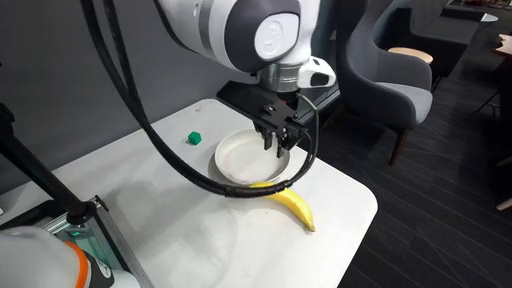}}\\[-2pt]
{\scriptsize F72 / 7.2s\strut}\\[-1pt]
{\scriptsize Loss evidence\strut}\end{minipage}\hfill%
\begin{minipage}[t]{0.137\linewidth}\centering
\setlength{\fboxsep}{0pt}\setlength{\fboxrule}{0.8pt}
\fcolorbox{white}{white}{\includegraphics[width=\dimexpr\linewidth-1.6pt\relax,trim=81.2222bp 0.0000bp 43.2222bp 70.0000bp,clip]{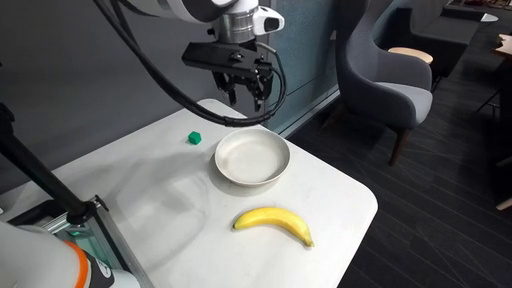}}\\[-2pt]
{\scriptsize F81 / 8.1s\strut}\\[-1pt]
{\scriptsize After loss\strut}\end{minipage}\hfill%
\begin{minipage}[t]{0.137\linewidth}\centering
\setlength{\fboxsep}{0pt}\setlength{\fboxrule}{0.8pt}
\fcolorbox{white}{white}{\includegraphics[width=\dimexpr\linewidth-1.6pt\relax,trim=81.2222bp 0.0000bp 43.2222bp 70.0000bp,clip]{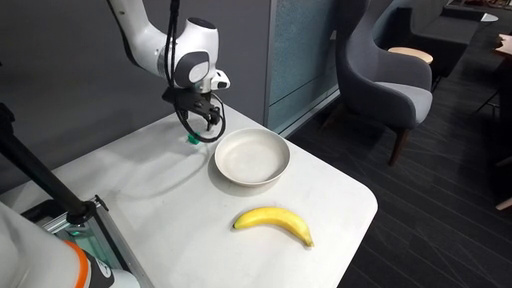}}\\[-2pt]
{\scriptsize F129 / 12.9s\strut}\\[-1pt]
{\scriptsize Later\strut}\end{minipage}\hfill%
\begin{minipage}[t]{0.137\linewidth}\centering
\setlength{\fboxsep}{0pt}\setlength{\fboxrule}{0.8pt}
\fcolorbox{endpointred}{white}{\includegraphics[width=\dimexpr\linewidth-1.6pt\relax,trim=81.2222bp 0.0000bp 43.2222bp 70.0000bp,clip]{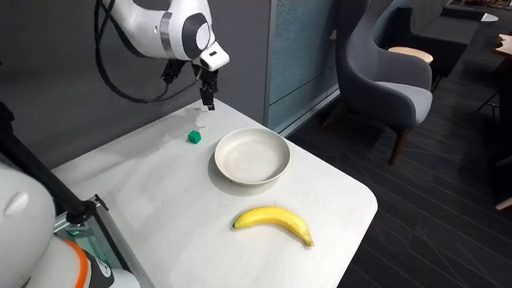}}\\[-2pt]
{\scriptsize \textcolor{endpointred}{\textbf{F399 / 39.9s}}\strut}\\[-1pt]
{\scriptsize \textcolor{endpointred}{\textbf{Failure / $T$}}\strut}\end{minipage}
\end{minipage}\par
\vfill
\begin{minipage}{\linewidth}
\setlength{\parskip}{\eventExampleParskip}
{\normalsize \textbf{(c)} Place the carrot inside the paper cup. The carrot falls from the raised gripper and is still on the table outside the cup at T.\par}
\noindent \begin{minipage}[t]{0.137\linewidth}\centering
\setlength{\fboxsep}{0pt}\setlength{\fboxrule}{0.8pt}
\fcolorbox{white}{white}{\includegraphics[width=\dimexpr\linewidth-1.6pt\relax,trim=39.4444bp 13.0000bp 81.4444bp 55.0000bp,clip]{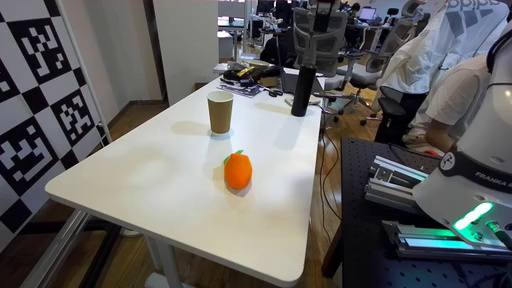}}\\[-2pt]
{\scriptsize F0 / 0.0s\strut}\\[-1pt]
{\scriptsize Start\strut}\end{minipage}\hfill%
\begin{minipage}[t]{0.137\linewidth}\centering
\setlength{\fboxsep}{0pt}\setlength{\fboxrule}{0.8pt}
\fcolorbox{white}{white}{\includegraphics[width=\dimexpr\linewidth-1.6pt\relax,trim=39.4444bp 13.0000bp 81.4444bp 55.0000bp,clip]{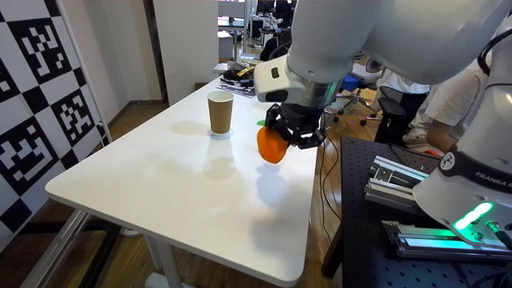}}\\[-2pt]
{\scriptsize F104 / 10.4s\strut}\\[-1pt]
{\scriptsize Before loss\strut}\end{minipage}\hfill%
\begin{minipage}[t]{0.137\linewidth}\centering
\setlength{\fboxsep}{0pt}\setlength{\fboxrule}{0.8pt}
\fcolorbox{black}{white}{\includegraphics[width=\dimexpr\linewidth-1.6pt\relax,trim=39.4444bp 13.0000bp 81.4444bp 55.0000bp,clip]{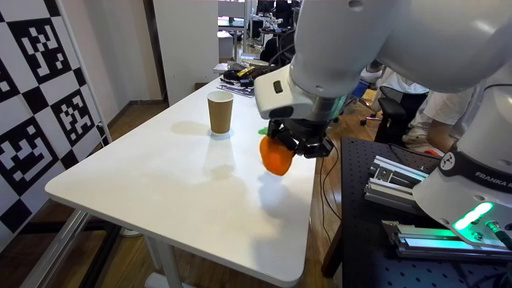}}\\[-2pt]
{\scriptsize \textbf{F116 / 11.6s}\strut}\\[-1pt]
{\scriptsize \textbf{Ref. $a$}\strut}\end{minipage}\hfill%
\begin{minipage}[t]{0.137\linewidth}\centering
\setlength{\fboxsep}{0pt}\setlength{\fboxrule}{0.8pt}
\fcolorbox{white}{white}{\includegraphics[width=\dimexpr\linewidth-1.6pt\relax,trim=39.4444bp 13.0000bp 81.4444bp 55.0000bp,clip]{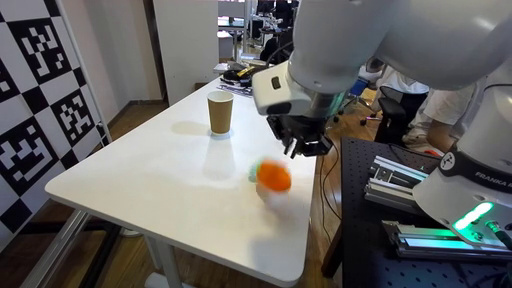}}\\[-2pt]
{\scriptsize F118 / 11.8s\strut}\\[-1pt]
{\scriptsize Loss evidence\strut}\end{minipage}\hfill%
\begin{minipage}[t]{0.137\linewidth}\centering
\setlength{\fboxsep}{0pt}\setlength{\fboxrule}{0.8pt}
\fcolorbox{white}{white}{\includegraphics[width=\dimexpr\linewidth-1.6pt\relax,trim=39.4444bp 13.0000bp 81.4444bp 55.0000bp,clip]{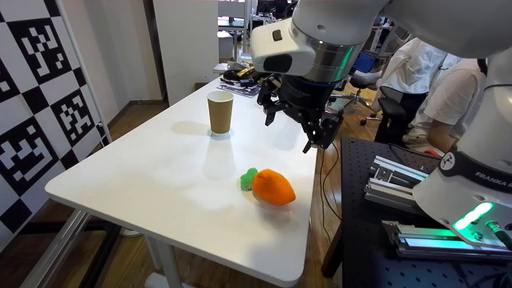}}\\[-2pt]
{\scriptsize F121 / 12.1s\strut}\\[-1pt]
{\scriptsize After loss\strut}\end{minipage}\hfill%
\begin{minipage}[t]{0.137\linewidth}\centering
\setlength{\fboxsep}{0pt}\setlength{\fboxrule}{0.8pt}
\fcolorbox{white}{white}{\includegraphics[width=\dimexpr\linewidth-1.6pt\relax,trim=39.4444bp 13.0000bp 81.4444bp 55.0000bp,clip]{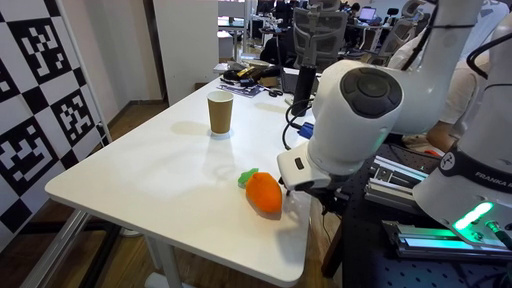}}\\[-2pt]
{\scriptsize F176 / 17.6s\strut}\\[-1pt]
{\scriptsize Later\strut}\end{minipage}\hfill%
\begin{minipage}[t]{0.137\linewidth}\centering
\setlength{\fboxsep}{0pt}\setlength{\fboxrule}{0.8pt}
\fcolorbox{endpointred}{white}{\includegraphics[width=\dimexpr\linewidth-1.6pt\relax,trim=39.4444bp 13.0000bp 81.4444bp 55.0000bp,clip]{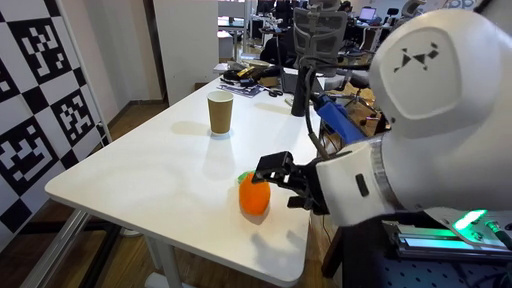}}\\[-2pt]
{\scriptsize \textcolor{endpointred}{\textbf{F400 / 40.0s}}\strut}\\[-1pt]
{\scriptsize \textcolor{endpointred}{\textbf{Failure / $T$}}\strut}\end{minipage}
\end{minipage}\par
\vfill
\begin{minipage}{\linewidth}
\setlength{\parskip}{\eventExampleParskip}
{\normalsize \textbf{(d)} Lift the green frog from the table. The frog is lifted, slips out and rolls onto the mat, then remains ungrasped at T.\par}
\noindent \begin{minipage}[t]{0.137\linewidth}\centering
\setlength{\fboxsep}{0pt}\setlength{\fboxrule}{0.8pt}
\fcolorbox{white}{white}{\includegraphics[width=\dimexpr\linewidth-1.6pt\relax,trim=40.8333bp 78.0000bp 97.8333bp 0.0000bp,clip]{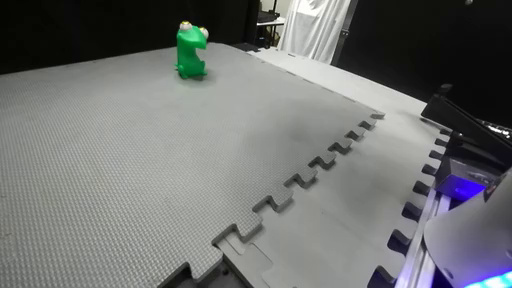}}\\[-2pt]
{\scriptsize F0 / 0.0s\strut}\\[-1pt]
{\scriptsize Start\strut}\end{minipage}\hfill%
\begin{minipage}[t]{0.137\linewidth}\centering
\setlength{\fboxsep}{0pt}\setlength{\fboxrule}{0.8pt}
\fcolorbox{white}{white}{\includegraphics[width=\dimexpr\linewidth-1.6pt\relax,trim=40.8333bp 78.0000bp 97.8333bp 0.0000bp,clip]{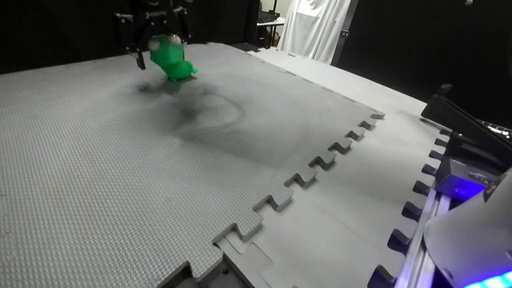}}\\[-2pt]
{\scriptsize F156 / 15.6s\strut}\\[-1pt]
{\scriptsize Before loss\strut}\end{minipage}\hfill%
\begin{minipage}[t]{0.137\linewidth}\centering
\setlength{\fboxsep}{0pt}\setlength{\fboxrule}{0.8pt}
\fcolorbox{black}{white}{\includegraphics[width=\dimexpr\linewidth-1.6pt\relax,trim=40.8333bp 78.0000bp 97.8333bp 0.0000bp,clip]{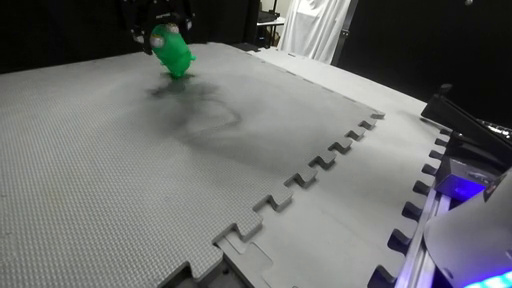}}\\[-2pt]
{\scriptsize \textbf{F168 / 16.8s}\strut}\\[-1pt]
{\scriptsize \textbf{Ref. $a$}\strut}\end{minipage}\hfill%
\begin{minipage}[t]{0.137\linewidth}\centering
\setlength{\fboxsep}{0pt}\setlength{\fboxrule}{0.8pt}
\fcolorbox{white}{white}{\includegraphics[width=\dimexpr\linewidth-1.6pt\relax,trim=40.8333bp 78.0000bp 97.8333bp 0.0000bp,clip]{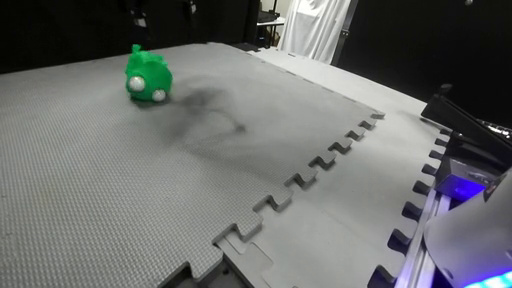}}\\[-2pt]
{\scriptsize F169 / 16.9s\strut}\\[-1pt]
{\scriptsize Loss evidence\strut}\end{minipage}\hfill%
\begin{minipage}[t]{0.137\linewidth}\centering
\setlength{\fboxsep}{0pt}\setlength{\fboxrule}{0.8pt}
\fcolorbox{white}{white}{\includegraphics[width=\dimexpr\linewidth-1.6pt\relax,trim=40.8333bp 78.0000bp 97.8333bp 0.0000bp,clip]{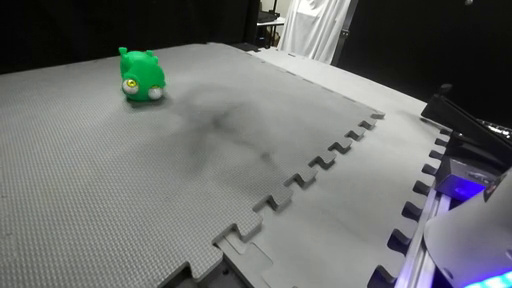}}\\[-2pt]
{\scriptsize F180 / 18.0s\strut}\\[-1pt]
{\scriptsize After loss\strut}\end{minipage}\hfill%
\begin{minipage}[t]{0.137\linewidth}\centering
\setlength{\fboxsep}{0pt}\setlength{\fboxrule}{0.8pt}
\fcolorbox{white}{white}{\includegraphics[width=\dimexpr\linewidth-1.6pt\relax,trim=40.8333bp 78.0000bp 97.8333bp 0.0000bp,clip]{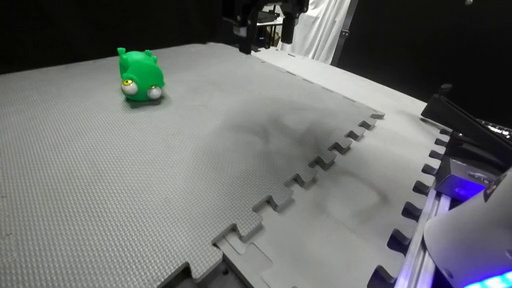}}\\[-2pt]
{\scriptsize F228 / 22.8s\strut}\\[-1pt]
{\scriptsize Later\strut}\end{minipage}\hfill%
\begin{minipage}[t]{0.137\linewidth}\centering
\setlength{\fboxsep}{0pt}\setlength{\fboxrule}{0.8pt}
\fcolorbox{endpointred}{white}{\includegraphics[width=\dimexpr\linewidth-1.6pt\relax,trim=40.8333bp 78.0000bp 97.8333bp 0.0000bp,clip]{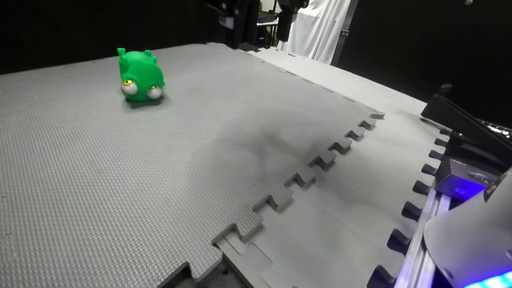}}\\[-2pt]
{\scriptsize \textcolor{endpointred}{\textbf{F400 / 40.0s}}\strut}\\[-1pt]
{\scriptsize \textcolor{endpointred}{\textbf{Failure / $T$}}\strut}\end{minipage}
\end{minipage}\par
\vspace{14pt}
{\scriptsize Source: RoboArena. Examples use final annotation records from the procedure in Appendix~\ref{app:annotation-procedure}. Black borders mark annotated event anchors. Green/red endpoints indicate success/failure. Frame numbers and times identify positions in each video. Some views are cropped to highlight task-relevant regions.\par}
\egroup
\endgroup
\clearpage

\clearpage
\endgroup

\endgroup

\section{Model and Training Details}
\label{app:method-training}

\subsection{Model Inputs and Prediction Heads}
\label{app:model-inputs}

\paragraph{Inputs and visual history.}
Using the input components defined in Section~\ref{sec:progress-model},
we process images at $224\times224$ resolution and take the start
observation from the beginning of the trajectory or training segment.
The instruction and images form a single multimodal input, with text
markers indicating image roles and temporal order. The visual history
contains five observations. For chronological inputs, it ends at the
queried observation and contains no later observations. Videos in the
event-annotated dataset use approximately $0.1$-second sampling intervals,
with each requested timestamp mapped to the latest available frame at or
before it. Stage-1 pretraining and replay use $0.25$-second intervals.
AutoEval uses the latest five logged observations without assuming a
fixed frame rate. When history is insufficient at the beginning of a
trajectory or segment, we repeat the first eligible observation.
Temporal transformations for trend supervision are described in
Appendix~\ref{app:progress-trend}.

\paragraph{Goal conditioning.}
For the event-annotated dataset, goal images are selected from accepted
completion frames of successful training trajectories with compatible
task, scene, and camera context. Successful trajectories containing events
may also supply goal images. When a compatible goal image is unavailable
or omitted during training, we use a dedicated no-goal token.
Goal-image dropout settings are given in
Appendix~\ref{app:training-configuration}.

\paragraph{Backbone and prediction heads.}
With Qwen3.5-4B as the backbone, we append separate progress and trend
query tokens to the multimodal input and apply a prediction head to each
token's final-layer hidden state. Each head applies layer normalization
followed by a two-layer MLP
with hidden width $d/2$, GELU activation, and dropout of $0.1$ between
the linear layers; $d$ denotes the backbone hidden dimension. The progress
head applies softmax to 101 logits, giving
probabilities $\pi_{\theta,k}(x_t)$ over support values $k/100$ for
$k=0,\ldots,100$. Its scalar prediction is the distribution's expectation:
\begin{equation}
    p_{\theta}(x_t)
    = \sum_{k=0}^{100}\frac{k}{100}\,\pi_{\theta,k}(x_t).
    \label{eq:progress-expectation}
\end{equation}
The trend head applies softmax to three logits for increasing, stable,
and decreasing progress.

\subsection{Progress and Trend Supervision}
\label{app:progress-trend}

\paragraph{Temporal labels.}
Successful demonstration segments used for pretraining and replay, and
normal successful rollouts in the event-annotated dataset, provide
temporal labels. Between the first eligible observation $b$ and the
accepted completion boundary $S$ of the segment or rollout, we assign
$y_t=(u_t-u_b)/(u_S-u_b)$, where $u$ denotes time on the processed video
timeline or the ordered observation index of a logged sequence.
These labels serve as a numerical proxy for progress toward completing
the accompanying instruction; observations after completion are excluded.
Segment boundaries and timeline processing are described in
Appendix~\ref{app:training-configuration} for pretraining data and
Appendix~\ref{app:annotation-procedure} for the event-annotated rollouts.
Replay retains the original pretraining segments and temporal labels.
A zero-length interval receives only the completion label $y=1$,
without trend supervision.

\paragraph{Mean-preserving soft targets.}
To supervise the progress distribution in Appendix~\ref{app:model-inputs}
while preserving the scalar target, we encode $y\in[0,1]$ using symmetric
discrete Gaussian kernels over the 101 support values. With $K=100$, define
$\mathcal S_j=\{k\in\{0,\ldots,K\}:|k-j|\leq\min(j,K-j)\}$ and
\begin{equation}
    G_j(k)=
    \frac{\mathbf{1}[k\in\mathcal S_j]
    \exp\!\left(-\frac{(k-j)^2}{2\sigma^2}\right)}
    {\sum_{r\in\mathcal S_j}
    \exp\!\left(-\frac{(r-j)^2}{2\sigma^2}\right)},
    \label{eq:app-target-kernel}
\end{equation}
where $\sigma=2$ bins. For $j_-=\lfloor Ky\rfloor$,
$j_+=\lceil Ky\rceil$, and $\alpha=Ky-j_-$, the soft target is
\begin{equation}
    \widetilde\pi_k(y)=(1-\alpha)G_{j_-}(k)+\alpha G_{j_+}(k).
    \label{eq:app-soft-progress-target}
\end{equation}
The symmetric support and interpolation preserve expectation $y$; targets
at zero and one remain concentrated at their respective endpoints.

\paragraph{Auxiliary trend targets.}
During pretraining, successful segments provide synthetic increasing,
stable, and decreasing trend labels through the following sequences:
\begin{equation}
\begin{aligned}
    H_t^{\uparrow}&=(o_{t-4\Delta},\ldots,o_{t-\Delta},o_t),\\
    H_t^{=}&=(o_t,o_t,o_t,o_t,o_t),\\
    H_t^{\downarrow}&=(o_{t+4\Delta},\ldots,o_{t+\Delta},o_t),
\end{aligned}
    \label{eq:app-trend-histories}
\end{equation}
where $\Delta$ is the sampling interval. The decreasing sequence uses
later observations from the source segment in reverse order as a training
transformation. All three sequences end at $o_t$ and retain the same
progress target $y_t$; their trend labels follow the displayed order under
the successful-segment progress proxy. Replay retains this construction.
Pretraining endpoint queries and sequences lacking the temporal support
required by a transformation are excluded from trend supervision.

For new normal successes in the event-annotated dataset, we use
chronological histories and assign increasing labels only when all five
observations come from distinct sampling positions, with no repetitions
introduced by prefix padding. Histories containing repeated observations
from prefix padding receive no trend label.

\paragraph{Losses and supervision scope.}
The progress loss $\mathcal L_{\mathrm{progress}}$ is cross-entropy
between the soft target $\widetilde\pi(y)$ and the predicted progress
distribution. The trend loss $\mathcal L_{\mathrm{trend}}$ is
cross-entropy with the corresponding three-class label. Each loss is
averaged over its own valid queries and is zero when none are available.
Queries used for relation supervision, including reference and comparison observations,
receive neither temporal labels nor trend labels. They supervise
scalar progress through the relation losses in
Appendix~\ref{app:event-losses}.

\subsection{Progress-Relation Losses}
\label{app:event-losses}

\paragraph{Reference predictions and supported comparisons.}
For each event $e$, $\mathcal M_e$ is a nonempty set of sampled comparison observations
supported by its annotation (Appendix~\ref{app:event-definitions}), with
reference $a$ and, for regression and recovery, recovery reference $r$.
For regression-and-recovery events, $a$ retains the attained state at loss
onset and $r$ restores it, so the anchors represent approximately equal
task progress. Reference and comparison predictions share the task instruction,
start reference, and optional goal. With $p_t=p_\theta(x_t)$, we use
$\bar p_a=\operatorname{sg}(p_a)$ and, where applicable,
$\bar p_r=\operatorname{sg}(p_r)$, where $\operatorname{sg}$ stops gradients
within the relation loss. Only comparison predictions receive these
gradients; references are recomputed with the current model.

\paragraph{Relation penalties and margins.}
For $t\in\mathcal M_e$, we use the following squared hinge penalties,
where $[z]_+=\max(z,0)$:
\begin{equation}
\begin{aligned}
    \ell_{\mathrm{stall}}(t)
    &=\bigl[|p_t-\bar p_a|-\epsilon\bigr]_+^2,\\
    \ell_{\mathrm{regression}}(t)
    &=\bigl[p_t-\min(\bar p_a,\bar p_r)+m_e\bigr]_+^2,\\
    \ell_{\mathrm{no\text{-}progress}}(t)
    &=[p_t-\bar p_a]_+^2,\\
    \ell_{\mathrm{unrecovered}}(t)
    &=[p_t-\bar p_a+m]_+^2.
\end{aligned}
    \label{eq:app-event-penalties}
\end{equation}
Progress stall permits variation within $\epsilon$ of the reference score.
Regression and recovery use the lower of the two reference scores,
without requiring them to be equal.

We set $\epsilon=0.02$ and the base decrease margin $m=0.1$.
For regression and recovery, the effective margin is
$m_e=\min\{m,\tfrac12\min(\bar p_a,\bar p_r)\}$, reducing the required
separation when reference scores approach zero in the bounded progress
range $[0,1]$. Unrecovered regression retains the fixed margin $m$.
These tolerances and margins are optimization parameters, not annotated
completion percentages or measured amounts of progress loss.

\paragraph{Loss aggregation.}
\begin{samepage}
For the set $\mathcal B_e$ of valid event groups in a batch, we average
comparison penalties within each event and then average over events:
\begin{equation}
    \mathcal L_{\mathrm{rel}}
    =\frac{1}{\max(1,|\mathcal B_e|)}
    \sum_{e\in\mathcal B_e}\frac{1}{|\mathcal M_e|}
    \sum_{t\in\mathcal M_e}\ell_e(t),
    \label{eq:app-event-aggregation}
\end{equation}
where $\ell_e$
is the event-type penalty. Each event has equal weight regardless of its
comparison count, and a batch with no valid events has zero loss.
Sampling proportions are given in Appendix~\ref{app:training-configuration};
the progress and trend losses in Appendix~\ref{app:progress-trend} are combined
with this term through the joint objective in Section~\ref{sec:joint-learning}.
\par\end{samepage}

\subsection{Pretraining and Rollout Joint Training}
\label{app:training-configuration}

This section describes pretraining and rollout joint training of the full
Event2Reward model. The controlled supervision ablations in
Section~\ref{sec:ablation} use a separate training configuration,
detailed in Appendix~\ref{app:ablation}.

\paragraph{Pretraining corpus.}
Pretraining combines public robot demonstrations, additional robot expert
demonstrations, and private human manipulation videos
(Table~\ref{tab:pretraining-corpus}). DROID uses the processed release
provided by DreamZero~\citep{ye2026wam}.

\begin{table}[t]
    \centering
    \caption{\textbf{Pretraining corpus composition.}
    Approximate durations sum the retained training segments after
    preprocessing and dataset splitting, before repeated sampling.
    These durations describe corpus composition rather than training
    sampling probabilities.}
    \label{tab:pretraining-corpus}
    \begin{adjustbox}{max width=\linewidth}
\begin{tabular}{@{}p{0.72\linewidth}r@{}}
        \toprule
        Source & Duration (h) \\
        \midrule
        \multicolumn{2}{@{}l}{\textit{Robot demonstrations}} \\
        AgiBot World Beta~\citep{agibot2025world} & 2,345 \\
        Galaxea Open-World~\citep{jiang2025galaxea} & 450 \\
        RoboMIND V2~\citep{hou2026robomind2} & 319 \\
        RoboCOIN~\citep{wu2026robocoin} & 283 \\
        DROID~\citep{khazatsky2025droid} & 268 \\
        RoboMIND V1~\citep{wu2025robomind} & 190 \\
        Baihu: Qingloong subset~\citep{openloong2025baihu} & 148 \\
        Other robot demonstrations & 557 \\
        \textbf{Robot subtotal} & \textbf{4,560} \\
        \midrule
        Human manipulation videos (private corpus) & 15,487 \\
        \midrule
        \textbf{Total} & \textbf{20,047} \\
        \bottomrule
    \end{tabular}
    \end{adjustbox}
\end{table}

\paragraph{Training segments and temporal labels.}
We organize pretraining data into successful task or subtask segments,
each paired with an instruction describing that segment. Segment
boundaries follow available source annotations or delimit a complete
successful execution.
Source-specific preprocessing includes removing trailing inactivity and
combining short adjacent human-video subtasks, together with their
instructions, in temporal order. For DROID, we retain the DreamZero
release's processed timeline, in which idle frames have already been
removed. Temporal labels are normalized over each retained segment,
so subtask segments measure progress toward completing the instructed
subtask. Training queries use the inputs in
Appendix~\ref{app:model-inputs}; the temporal labels and auxiliary
trend supervision are defined in
Appendix~\ref{app:progress-trend}.

\paragraph{Pretraining and initialization.}
We initialize the model from Qwen3.5-4B and pretrain it with the progress
and auxiliary trend objectives in Appendix~\ref{app:progress-trend},
omitting available goal images with probability $0.5$.
Joint training starts from the Stage-1 checkpoint listed in
Table~\ref{tab:pretraining-joint-settings}, before the pretraining schedule ends.

\paragraph{Data sampling across training stages.}
During pretraining, robot demonstrations and human videos are sampled
with probabilities of 70\% and 30\%, respectively.
For joint training, we use the real-robot training split of the
event-annotated dataset described in Appendix~\ref{app:event-dataset},
supplemented with 5,452 simulated rollouts from
RoboFAC~\citep{ye2026robofac}.\footnote{
The RoboFAC simulation subset also includes 292 validation trajectories,
which are excluded from training. This subset is outside the real-robot
collection reported in Appendix~\ref{app:event-dataset}.
}
Across the real and simulated data, the rollout training pools contain
17,572 normal successful trajectories and 49,571 event-bearing trajectories.

Replay uses a fixed subset of 147,456 successful task or subtask segments
from the pretraining training set. These segments retain their original
temporal labels, temporal transformations, and goal-image dropout.
Robot and human replay data follow the same 70\%/30\% sampling ratio
used during pretraining.

Joint training allocates 25\%, 25\%, and 50\% of the model-query budget
to replay, normal successful rollouts, and event-bearing rollouts,
respectively. Both rollout pools include real and simulated executions.
Normal successful rollouts provide temporal labels and valid trend labels,
while event-bearing rollouts provide within-trajectory progress relations
as defined in Appendix~\ref{app:event-losses}.
Available goal images in the rollout pools are retained without
additional dropout.

Mixture proportions are measured in individual model
queries, including event references and comparison
observations. For each sampled event, we select at most
eight supported comparison observations, together with
its required reference observations.
Event groups remain intact when constructing batches.
Across the total query budget, progress stall, regression
and recovery, no further progress, and unrecovered regression
receive approximately 11.26\%, 3.48\%, 23.54\%,
and 11.73\%, respectively.
This sampling allocation determines event exposure;
the relation loss itself uses the within-event and
across-event averaging defined in
Appendix~\ref{app:event-losses}.

\paragraph{Optimization and training budget.}
Both stages update the language backbone, the final three vision-encoder
layers, the visual merger, and both prediction heads using AdamW
(weight decay $0.01$), BF16 mixed precision, and gradient-norm clipping at $1.0$.
Learning rates use linear warmup followed by cosine decay.
Table~\ref{tab:pretraining-joint-settings} gives parameter-group learning
rates, query budgets, and checkpoint steps separately from scheduler horizons.

\begin{table}[t]
    \centering
    \caption{
        Pretraining and rollout joint-training settings.
        Learning rates are base values before scheduling.
        Query counts include reference observations
        used in relation supervision.
    }
    \label{tab:pretraining-joint-settings}
    \begin{adjustbox}{max width=\linewidth}
\begin{tabular}{lcc}
        \toprule
        Setting & Pretraining & Rollout joint training \\
        \midrule
        Checkpoint step & 50,000 & 10,000 \\
        Scheduler horizon & 500,000 & 40,000 \\
        Warmup updates & 25,000 & 2,000 \\
        Global queries per update & 256 & 80 \\
        Queries to checkpoint & 12.8M & 0.8M \\
        Language learning rate
            & $10^{-5}$ & $2\times10^{-6}$ \\
        Vision learning rate
            & $10^{-6}$ & $5\times10^{-7}$ \\
        Prediction-head learning rate
            & $2\times10^{-5}$ & $10^{-5}$ \\
        Progress-loss weight $\lambda_p$ & 1 & 1 \\
        Trend-loss weight $\lambda_t$ & 1 & 1 \\
        Relation-loss weight $\lambda_{\mathrm{rel}}$ & 0 & 10 \\
        Training seed & 42 & 42 \\
        \bottomrule
    \end{tabular}
    \end{adjustbox}
\end{table}

\paragraph{Validation and checkpoint reporting.}
We monitor joint training with a fixed query plan
covering progress prediction on normal successes,
consistency with annotated progress relations, and retention on
Stage-1 data.
Goal-present and no-goal conditions are evaluated
separately.
The retention set consists of Stage-1 training
segments held out from the Stage-2 replay pool.
Test partitions are excluded from this monitoring
plan.

Validation measurements at the reported checkpoint diagnose progress
calibration, relation consistency, and retention; completing the longer
schedule is not treated as evidence of a better model.

\section{Experimental Details and Additional Results}
\label{app:experiments}

\subsection{Ablation Study}
\label{app:ablation}

\paragraph{Supervision conditions.}
We compare five reward-model variants on LIBERO-10
Tasks~1, 3, 7, and 9.
All variants use Qwen3-VL-4B-Instruct and start from
the same checkpoint trained for 2,000 updates using
temporal labels from normal successful rollouts.
Each variant then undergoes 2,000 additional updates
with the supervision specified in
Table~\ref{tab:ablation-supervision-settings}.
We preserve chronological input histories and disable
auxiliary trend supervision throughout these ablations.
Variant names refer only to the supervision used during
continued training.

\begin{table}[t]
    \centering
    \caption{
        Supervision used during continued training.
        All five variants start from the same checkpoint trained
        with temporal labels from normal successful rollouts.
        Relation-loss-off retains the
        mixed-sampling recipe of Progress + Relation but
        disables the relation loss.
    }
    \label{tab:ablation-supervision-settings}
    \begin{adjustbox}{max width=\linewidth}
\begin{tabular}{lccc}
        \toprule
        Variant
        & \shortstack{Temporal\\labels}
        & \shortstack{Trajectory\\preference}
        & \shortstack{Progress\\relations} \\
        \midrule
        Progress-only         & Yes & --  & --  \\
        Progress + Preference & Yes & Yes & --  \\
        Relation-only            & --  & --  & Yes \\
        Progress + Relation      & Yes & --  & Yes \\
        Relation-loss-off        & Yes & --  & --  \\
        \bottomrule
    \end{tabular}
    \end{adjustbox}
\end{table}

\paragraph{Training data.}
For the controlled supervision ablations, we use $\pi_0$ rollouts
covering all ten LIBERO-10 tasks.
We select 2,000 normal successful rollouts.
This pool supplies temporal labels for the shared initialization
and for continued training in variants that retain temporal supervision.
An additional training pool contains 2,559 rollouts, comprising
839 event-containing successes and 1,720 failures.
Relation and trajectory-preference supervision use this same
additional pool.

The two training pools and the held-out validation and test sets,
containing 128 and 139 rollouts respectively, are mutually disjoint
at the trajectory level.
Relation and preference supervision use the same trajectory splits.
The held-out sets cover the same ten tasks as training, while
downstream reinforcement learning evaluates Tasks~1, 3, 7, and 9.

\paragraph{Sampling and supervision.}
Temporal labels are constructed by normalizing time within normal
successful segments.
Progress-only uses clips from the normal-success pool throughout
continued training, while Progress + Relation allocates half of its
clip budget to this pool and half to event data.
Event clips receive relation supervision and no temporal labels.
Relation-loss-off retains the same 50/50 sampling recipe but
sets the relation-loss weight to zero, controlling for the
sampling change relative to Progress-only.
Relation-only uses event data without temporal supervision;
any padding clips carry no supervision.
Under the fixed training budget, this condition therefore increases
event exposure as well as removing temporal supervision.

\paragraph{Trajectory preference supervision.}
Progress + Preference uses the same normal-success pool and
additional rollout pool as Progress + Relation.
Preference supervision is derived from trajectory outcomes,
without using event categories or boundaries.
For preference supervision, we pair a successful and
a failed trajectory from the same task and balance
their presentation order.
Each trajectory is represented by eight uniformly
sampled frames, together with its start and goal images.
The model jointly processes the two trajectories,
and a separate preference head is trained with binary
cross-entropy to favor the successful trajectory.
Each update uses 40 normal-success clips and 14 trajectory pairs,
matching the normal-success clip count and total image and
visual-token exposure of Progress + Relation.
Over 2,000 updates, this gives 80,000 normal-success clip exposures
and 28,000 pair exposures.
Relation supervision samples local event groups, whereas
preference supervision samples whole-trajectory pairs.
During downstream RL, the preference-trained model
provides rewards through its progress head.
Held-out evaluation of the comparison head confirms high
accuracy on same-task success--failure preferences
(Appendix~\ref{app:ablation-preference-diagnostics}).

\paragraph{Optimization.}
We use AdamW with weight decay $0.01$ and 100 warmup
updates.
Learning rates are $2\times10^{-6}$ for the language
backbone, $5\times10^{-7}$ for the trainable vision
parameters, and $10^{-5}$ for the prediction heads.
We update the language backbone, the final three
vision-encoder layers, the visual merger, and the
heads used by each variant.
The progress, relation, and preference losses have weights
1, 10, and 1, respectively, when active.
The progress and relation objectives are defined in
Appendices~\ref{app:progress-trend}
and~\ref{app:event-losses}; auxiliary trend supervision
is disabled here.

\paragraph{Additional baselines.}
We additionally evaluate the publicly released
Robometer checkpoint~\citep{liang2026robometer}
without further fine-tuning, and a Sparse baseline
that uses only the environment reward.
Our choice of Qwen3-VL-4B-Instruct matches the backbone
used by Robometer.
Robometer is an external reward-model baseline,
separate from the Progress + Preference variant
trained under our shared initialization and data protocol.

\paragraph{Residual reinforcement learning.}
\label{app:ablation-rl}
We evaluate the reward models using a residual SAC framework
adapted from PLD~\citep{xiao2025pld}.
The $\pi_{0.5}$ base policy and reward model remain frozen
throughout RL training.
RL updates the residual actor, critics, and SAC temperature
parameter, while the visual backbone remains frozen.
Within each task, all conditions use the same base-policy
checkpoint and residual architecture.
At each decision step, the residual policy observes two camera
views, robot proprioception, and the proposed base-action chunk,
and predicts corrections for the next $H=5$ environment steps.
We first compose the action chunk as
\begin{equation}
    A_t = A_t^{\mathrm{base}} + \alpha \Delta A_t,
    \qquad
    \Delta A_t \in [-1,1]^{H\times 7},
    \qquad
    \alpha=0.1.
    \label{eq:ablation-residual-action}
\end{equation}
Corrections apply to all seven action dimensions.
We then clip the gripper component of $A_t$ to $[-1,1]$
before executing the resulting action chunk.
Network architecture and optimization settings are summarized
in Table~\ref{tab:ablation-sac-settings}.

\paragraph{Progress-based reward computation.}
For learned-reward conditions, we evaluate the progress
potential in Equation~\ref{eq:progress-reward} at action-chunk
boundaries.
For a chunk containing $k\leq H$ executed environment steps,
the training reward is
\begin{equation}
    R_t^{\mathrm{shaped}}
    =
    R_t^{\mathrm{env}}
    +
    \beta\left[
        \gamma^k(1-d_t)p_\theta(x_{t+k})
        -
        p_\theta(x_t)
    \right],
    \label{eq:ablation-chunk-reward}
\end{equation}
where $R_t^{\mathrm{env}}$ is the accumulated environment
reward within the chunk, $\gamma=0.998$, and $\beta=1$.
The indicator $d_t$ is one at an episode termination or
truncation boundary and zero otherwise; the next-state
potential contribution is therefore zero at these boundaries.
SAC bootstrapping uses the same duration-dependent discount
and episode-boundary mask.
Reward inference runs asynchronously with environment
interaction; transitions enter the replay buffer after their
shaped rewards have been computed.
Sparse uses environment rewards alone.

\paragraph{Optimization and interaction budget.}
RL updates use only online experience, with no offline
replay or critic pretraining.
Updates begin after the replay buffer contains 1,000 chunk
transitions.
Each learner iteration performs one critic update on a
128-transition batch, followed by four critic updates on
32-transition minibatches and one actor and temperature
update on the corresponding full batch.
Each run receives a budget of 1.2 million environment steps.
We use five environment seeds, $\{7,8,9,17,27\}$, for every
task and condition.

\begin{table}[t]
    \centering
    \caption{Shared residual SAC settings for the supervision
    ablations. Replay entries correspond to action chunks.}
    \label{tab:ablation-sac-settings}
    \begin{adjustbox}{max width=\linewidth}
\begin{tabular}{lc}
        \toprule
        Setting & Value \\
        \midrule
        Optimizer & AdamW \\
        Learning rate & $3\times10^{-4}$ \\
        Network weight decay & $10^{-4}$ \\
        Gradient clipping norm & 1.0 \\
        Replay capacity & 250,000 chunks \\
        Replay batch size & 128 chunks \\
        Target-network update coefficient & 0.005 \\
        Initial SAC temperature & 0.01 \\
        \midrule
        Actor hidden layers & $3\times256$ \\
        Critic hidden layers & $3\times256$ \\
        Number of critics & 2 \\
        Actor and critic normalization & LayerNorm \\
        Visual encoder & Pretrained ResNet-18 \\
        Visual backbone & Frozen \\
        \bottomrule
    \end{tabular}
    \end{adjustbox}
\end{table}

\paragraph{Policy evaluation.}
We evaluate policy snapshots in a separate environment
every 50 completed training episodes, using 50 evaluation
episodes per snapshot.
Evaluation uses deterministic residual actions while
retaining the base policy's default stochastic sampling.
The initial evaluation sets all residual actions to zero
and measures base-policy performance.
Evaluation success is the fraction of successful episodes
according to the environment's task-success criterion;
evaluation trajectories are not added to the training replay.

\paragraph{Evaluation and curve aggregation.}
Following the evaluation protocol above, we align curves
by linear interpolation between recorded observations,
without extrapolation or additional smoothing.
Curves show equally weighted means across five environment
seeds, with shaded bands indicating the mean
$\pm 1.96\,\mathrm{SEM}$.
Here, $\mathrm{SEM}$ is the sample standard deviation
across seeds divided by $\sqrt{5}$, and bands are clipped
to the valid success-rate range.
Each learned-reward condition uses the same reward-model
checkpoint across environment seeds; the bands therefore
summarize variation across downstream runs.
Result tables report means $\pm$ sample standard deviations
across seeds.

\paragraph{Summary metrics.}
For a given task and condition, let $\widetilde{S}_s(t)$
denote the linearly interpolated evaluation success rate
in $[0,1]$ for seed $s$ at environment step $t$.
The normalized AUC, expressed as a percentage, is
\begin{equation}
    \mathrm{AUC}_s
    = \frac{100}{T}
    \int_0^T \widetilde{S}_s(t)\,\mathrm{d}t.
    \label{eq:ablation-eval-auc}
\end{equation}
Within each task, all seven conditions and five seeds
are compared over the same evaluation interval $[0,T]$,
which is covered by every run.
For Tasks~1, 3, 7, and 9, the endpoints are
1,175,842, 1,176,236, 1,179,116, and 1,178,331
environment steps, respectively.
This metric measures average evaluation success over
a shared interaction interval.
Last-evaluation success uses each run's last periodic
evaluation within the 1.2-million-step training budget.
For the four-task average, we first average the task
scores within each seed, then compute the mean and
sample standard deviation across the five seeds.

\begin{figure}[t]
    \centering
    \includegraphics[width=\linewidth]{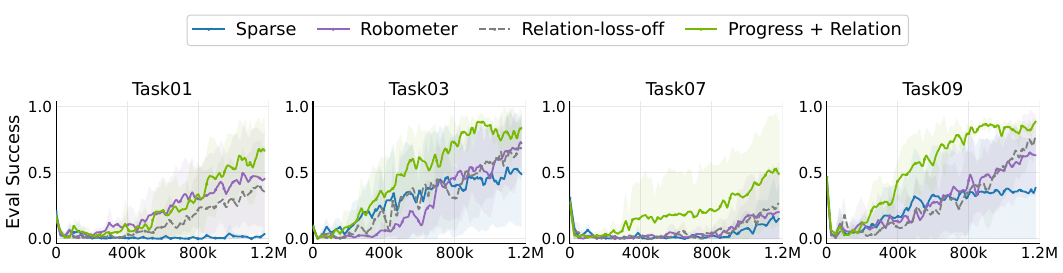}
    \caption{
        \textbf{Additional controls and reward-model baselines.}
        Independent evaluation success of Progress + Relation,
        Relation-loss-off, Sparse, and Robometer on LIBERO-10
        Tasks~1, 3, 7, and 9.
        Lines show means across five environment seeds;
        shaded bands indicate $\pm 1.96\,\mathrm{SEM}$.
        The horizontal axis shows environment interaction steps.
    }
    \label{fig:app-ablation-controls-curves}
\end{figure}

\paragraph{Complete RL results.}
Table~\ref{tab:ablation-full-rl-results} reports all seven
conditions on the four tasks.
The independent-evaluation curves for the four supervision variants are
shown in Figure~\ref{fig:ablation-supervision};
Figure~\ref{fig:app-ablation-controls-curves} shows the additional controls
and reward-model baselines.
Progress + Relation achieves the highest mean evaluation AUC
on all four tasks and the highest mean last-evaluation
success on all four tasks.
It also exceeds Relation-loss-off in mean evaluation AUC
on every task, with a four-task average of 36.7\%
versus 19.1\%.
This comparison supports a contribution from the relation
objective beyond the mixed-sampling recipe.
Progress + Relation also exceeds Progress + Preference in
mean evaluation AUC (36.7\% versus 19.4\%), despite the latter's
high held-out preference accuracy
(Table~\ref{tab:ablation-heldout-preference}).
Relative to Sparse and Robometer, joint supervision
achieves higher four-task averages for both metrics.
On Task~3, Progress + Relation achieves
higher mean evaluation AUC than Robometer
(48.9\% versus 26.5\%) and higher mean last-evaluation
success (88.8\% versus 70.4\%).

\begin{table}[t]
    \centering
    \setlength{\tabcolsep}{3pt}
    \caption{
        \textbf{Complete RL results on four LIBERO-10 tasks.}
        Values are percentages, reported as mean $\pm$
        sample standard deviation across five environment seeds.
        \textit{Average} first averages the four tasks
        within each seed.
        Bold indicates the highest mean in each column
        within each panel.
    }
    \label{tab:ablation-full-rl-results}
    \begin{adjustbox}{max width=\linewidth}
\begin{tabular}{lccccc}
        \toprule
        Method & Task 1 & Task 3 & Task 7 & Task 9 & Average \\
        \midrule
        \multicolumn{6}{l}{\textbf{(a) Normalized evaluation AUC}} \\
        \addlinespace[2pt]
        Sparse & $1.4 \pm 1.0$ & $30.4 \pm 29.6$ & $3.5 \pm 5.3$ & $24.5 \pm 30.9$ & $14.9 \pm 8.1$ \\
        Robometer & $22.1 \pm 19.8$ & $26.5 \pm 15.2$ & $5.6 \pm 9.1$ & $29.1 \pm 22.8$ & $20.8 \pm 11.9$ \\
        Progress-only & $6.6 \pm 7.8$ & $29.4 \pm 18.5$ & $5.5 \pm 8.6$ & $37.0 \pm 25.4$ & $19.6 \pm 10.1$ \\
        Progress + Preference & $16.1 \pm 18.7$ & $31.0 \pm 29.1$ & $6.0 \pm 9.0$ & $24.4 \pm 23.2$ & $19.4 \pm 14.3$ \\
        Relation-only & $6.8 \pm 12.8$ & $13.4 \pm 15.4$ & $1.1 \pm 0.2$ & $11.8 \pm 23.4$ & $8.3 \pm 7.9$ \\
        Relation-loss-off & $13.2 \pm 16.3$ & $30.4 \pm 22.1$ & $5.7 \pm 7.0$ & $27.2 \pm 17.7$ & $19.1 \pm 7.6$ \\
        Progress + Relation & $\mathbf{24.2} \pm 15.7$ & $\mathbf{48.9} \pm 14.3$ & $\mathbf{19.8} \pm 26.3$ & $\mathbf{53.8} \pm 8.8$ & $\mathbf{36.7} \pm 7.8$ \\
        \midrule
        \multicolumn{6}{l}{\textbf{(b) Last periodic evaluation success}} \\
        \addlinespace[2pt]
        Sparse & $4.0 \pm 7.9$ & $49.2 \pm 44.3$ & $15.6 \pm 32.7$ & $39.6 \pm 48.2$ & $27.1 \pm 24.4$ \\
        Robometer & $46.0 \pm 42.2$ & $70.4 \pm 27.2$ & $19.6 \pm 36.2$ & $62.4 \pm 34.7$ & $49.6 \pm 21.7$ \\
        Progress-only & $27.2 \pm 36.9$ & $69.6 \pm 38.6$ & $14.8 \pm 28.7$ & $72.8 \pm 25.3$ & $46.1 \pm 15.2$ \\
        Progress + Preference & $49.6 \pm 39.8$ & $57.2 \pm 35.1$ & $22.8 \pm 39.0$ & $58.0 \pm 32.9$ & $46.9 \pm 30.5$ \\
        Relation-only & $14.8 \pm 33.1$ & $39.2 \pm 36.6$ & $0.0 \pm 0.0$ & $17.6 \pm 39.4$ & $17.9 \pm 14.6$ \\
        Relation-loss-off & $39.6 \pm 38.4$ & $67.2 \pm 29.9$ & $27.2 \pm 34.7$ & $78.0 \pm 8.6$ & $53.0 \pm 14.8$ \\
        Progress + Relation & $\mathbf{66.4} \pm 28.5$ & $\mathbf{88.8} \pm 12.9$ & $\mathbf{52.8} \pm 48.6$ & $\mathbf{90.8} \pm 5.0$ & $\mathbf{74.7} \pm 14.2$ \\
        \bottomrule
    \end{tabular}
    \end{adjustbox}
\end{table}

\subsubsection{Held-Out Trajectory Preference Prediction}
\label{app:ablation-preference-diagnostics}

\paragraph{Evaluation protocol.}
To assess whether Progress + Preference learned the
trajectory comparisons used for supervision, we evaluate
the trained preference head of the same checkpoint used
in the RL ablations, without further training or checkpoint
selection.
The held-out split contains 46 successful and 93 failed
trajectories across all ten LIBERO-10 tasks, with no trajectory
overlap with the training or validation splits.
We evaluate all 433 unique within-task success--failure pairs,
including 73 pairs from the four RL tasks (Tasks~1, 3, 7, and 9).
Labels are determined by recorded terminal outcomes.
The 103 trajectories used in the progress diagnostics
(Appendix~\ref{app:ablation-diagnostics}) form a subset of this
split; the present evaluation uses all 139 trajectories because
it requires terminal outcomes rather than unambiguous local
event annotations.
Inputs follow the training protocol: eight uniformly sampled
frames per trajectory, together with its start image and a
fixed goal image from the training setup.
We randomize the primary A/B order with approximately balanced
labels within each task (seed 20260923), then evaluate every pair
again with the order reversed and its label inverted.

\paragraph{Metrics and uncertainty.}
Following the preference prediction metric in
Robometer~\citep{liang2026robometer}, accuracy is the fraction
of comparisons whose predicted ordering matches the ground truth;
here, a correct prediction selects the successful trajectory.
We additionally report binary cross-entropy (BCE) and order-swap
diagnostics.
Macro averages weight tasks equally, whereas pooled results
weight trajectory pairs equally.
We compute percentile 95\% intervals using 10,000 bootstrap
replicates, independently resampling successful and failed
trajectories within each task and weighting each pair by the
product of its endpoints' resampling multiplicities.
The two orders of a pair remain together.
Thus, shared trajectories and mirrored comparisons are not
treated as independent observations.
The intervals are conditional on the fixed checkpoint and task
set and do not capture reward-model training variability.
This is a same-task success--failure evaluation, rather than
a reproduction of Robometer's out-of-distribution benchmark.

\paragraph{Results and order sensitivity.}
Table~\ref{tab:ablation-heldout-preference} reports per-task and
aggregate results.
Primary accuracy averaged equally across ten tasks is 95.0\%
(95\% interval: 89.9--99.0\%), with BCE 0.1053;
accuracy on the four RL tasks is 98.3\% (93.3--100.0\%).
Pooling all 433 pairs gives 96.3\% primary accuracy (417/433)
and 95.6\% after swapping A/B (414/433).
The predicted preferred trajectory is unchanged by the swap
for 92.8\% of pairs.
Nevertheless, the model is sensitive to presentation order:
across both orders, pooled accuracy is 92.6\% when the successful
trajectory is in position A and 99.3\% when it is in position B.
Task~5 has only two successful and five failed trajectories;
its primary and swapped accuracies are 70.0\% and 90.0\%,
respectively.
We report both orders rather than selecting the more favorable one.

\begin{table}[t]
\centering
\setlength{\tabcolsep}{4pt}
\caption{Held-out preference prediction on LIBERO-10.
S/F denotes successful/failed trajectories. Accuracies are percentages.
Brackets give trajectory-bootstrap 95\% intervals for primary accuracy;
BCE is also measured on the primary order. Aggregate rows weight tasks
equally. A degenerate interval at 100\% reflects no observed errors,
not certainty about population accuracy.}
\label{tab:ablation-heldout-preference}
\begin{adjustbox}{max width=\linewidth}
\begin{tabular}{@{}lrrccc@{}}
\toprule
Task & S/F & Pairs & Primary accuracy & Swapped accuracy & BCE \\
\midrule
0 & 6/8 & 48 & 100.0 [100.0, 100.0] & 93.8 & 0.0210 \\
1 & 4/2 & 8 & 100.0 [100.0, 100.0] & 100.0 & 0.0187 \\
2 & 3/8 & 24 & 100.0 [100.0, 100.0] & 100.0 & 0.0167 \\
3 & 5/3 & 15 & 93.3 [73.3, 100.0] & 100.0 & 0.0765 \\
4 & 7/12 & 84 & 97.6 [89.3, 100.0] & 96.4 & 0.0663 \\
5 & 2/5 & 10 & 70.0 [20.0, 100.0] & 90.0 & 0.4478 \\
6 & 7/10 & 70 & 92.9 [78.6, 100.0] & 92.9 & 0.2140 \\
7 & 5/4 & 20 & 100.0 [100.0, 100.0] & 100.0 & 0.0138 \\
8 & 4/31 & 124 & 96.0 [87.9, 100.0] & 94.4 & 0.1631 \\
9 & 3/10 & 30 & 100.0 [100.0, 100.0] & 100.0 & 0.0155 \\
\midrule
All ten (macro) & 46/93 & 433 & 95.0 [89.9, 99.0] & 96.7 & 0.1053 \\
RL four (macro) & 17/19 & 73 & 98.3 [93.3, 100.0] & 100.0 & 0.0311 \\
\bottomrule
\end{tabular}
    \end{adjustbox}
\end{table}

\paragraph{Relation to downstream reward learning.}
The preference head achieves high held-out accuracy in comparing
successful and failed trajectories from the same task.
Our RL rewards are shaped using progress predictions at individual
time steps, which the preference head's pairwise outputs do not
directly provide.
Although preference training updates the shared backbone,
accurately ranking whole trajectories does not by itself ensure
that the progress head provides an effective shaping potential.
On the four RL tasks, Progress + Preference achieves a mean
evaluation AUC of 19.4\%, compared with 36.7\% for
Progress + Relation (Table~\ref{tab:ablation-full-rl-results}).

\subsubsection{Progress Prediction Diagnostics}
\label{app:ablation-diagnostics}
To diagnose Relation-only's poor RL performance, we evaluate progress
decreases and output compression on 103 held-out LIBERO-10 trajectories with
unambiguous reviewed annotations, covering all ten tasks and both outcomes; the RL
ablations cover four tasks. We use the fixed reward-model checkpoints from
those ablations, identical query frames and goals across conditions, and
original $[0,1]$ predictions without per-model or per-trajectory rescaling.
\emph{Relation-loss-off} retains the joint-training sampling recipe with the
relation loss disabled.

\paragraph{Relation satisfaction rate.}
We compute this metric on the \emph{unrecovered regression}
subset of the 103 held-out trajectories: 46 trajectories
spanning nine tasks.
These annotations identify achieved progress that is lost
and remains unrecovered at the end of the execution.
For each trajectory, we compare the prediction at the
annotated reference frame with predictions at 16 uniformly
spaced frames from 1.5 seconds after the reference to the
end of the trajectory.
A comparison satisfies the relation when the event-frame prediction
is strictly lower than the reference prediction; ties
count as unsatisfied comparisons.
We average the relation satisfaction rate within each trajectory,
then within each task, and finally equally across the
nine tasks.

\paragraph{Within-trajectory prediction range.}
To examine whether predictions distinguish different
trajectory stages, we sample nine evenly spaced frames
from the beginning to the end of each of the 103 trajectories.
For trajectory $i$, we compute
\[
    \Delta_i
    =
    \max_{t \in S_i} p_{i,t}
    -
    \min_{t \in S_i} p_{i,t},
    \qquad |S_i|=9,
\]
where $p_{i,t}$ is predicted progress and $S_i$ is the
sampled frame set.
We report the unweighted mean of $\Delta_i$ across
trajectories.
This measure diagnoses output compression; a larger range
alone does not establish more accurate progress predictions.

\begin{table}[t]
    \centering
    \caption{\textbf{Progress prediction diagnostics.}
    The relation satisfaction rate is averaged equally across
    nine tasks containing unrecovered regression annotations.
    Prediction range is averaged across 103 held-out
    trajectories.}
    \label{tab:ablation-progress-diagnostics}
    \begin{adjustbox}{max width=\linewidth}
\begin{tabular}{@{}lcc@{}}
        \toprule
        Supervision
        & \shortstack{Relation satisfaction\\rate (\%)}
        & \shortstack{Mean prediction\\range} \\
        \midrule
        Progress-only         & 41.62 & 0.5928 \\
        Progress + Preference & 48.57 & 0.5676 \\
        Relation-only            & 58.86 & 0.0021 \\
        Progress + Relation      & 54.47 & 0.5097 \\
        Relation-loss-off        & 44.89 & 0.5816 \\
        \bottomrule
    \end{tabular}
    \end{adjustbox}
\end{table}

\paragraph{Results and interpretation.}
Relation-only has the highest relation satisfaction rate but a mean prediction
range of 0.0021, versus 0.5097 for joint supervision
(Table~\ref{tab:ablation-progress-diagnostics}). Across all held-out queries,
its predictions are approximately 0.0010--0.0051. Thus, some correct relative
orderings coexist with nearly identical, near-zero scores. With our fixed
shaping coefficient, these values would yield small progress-dependent
chunk-reward terms (Equation~\ref{eq:ablation-chunk-reward}), consistent with
a weakened shaping signal. Joint supervision improves relation satisfaction
over Progress-only while maintaining variation. Together with the RL results
in Section~\ref{sec:ablation}, this supports temporal labels for scale
anchoring and relations for local progress constraints.

\begin{figure}[t]
    \centering
    \includegraphics[width=\linewidth]{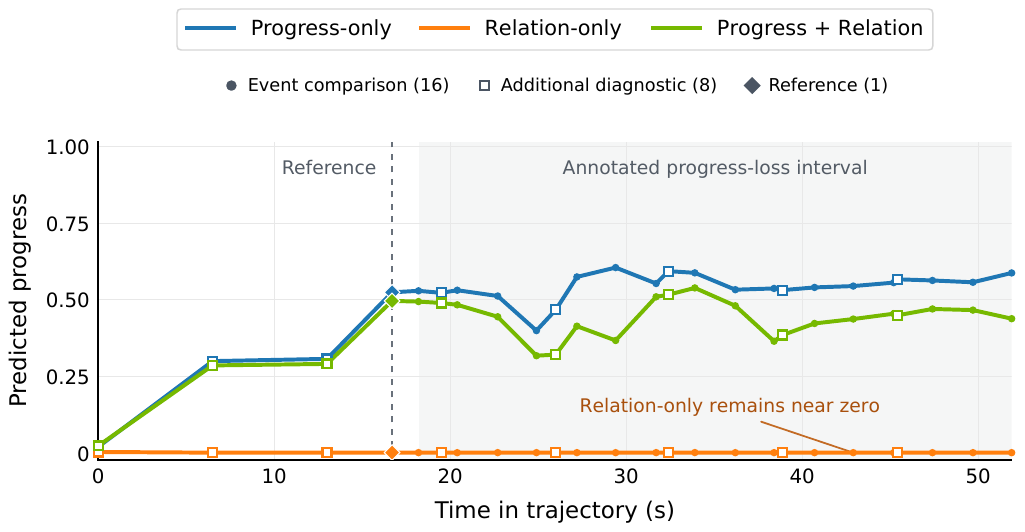}
    \caption{\textbf{A held-out example of progress-scale collapse.}
    Predicted progress on a LIBERO-10 Task~7 trajectory,
    shown on a common, unnormalized scale.
    Each curve connects 25 sampled predictions with straight lines:
    16 event-comparison
    points (filled circles), eight additional trajectory-diagnostic
    points (open squares), and one reference point (diamond).
    The dashed line marks the reference time, and shading indicates
    the progress-loss comparison interval.
    Only filled circles enter the reported decrease counts;
    the five open squares within the shaded interval are excluded.}
    \label{fig:ablation-scale-collapse}
\end{figure}

\paragraph{An example of progress-scale collapse.}
On the held-out LIBERO-10 Task~7 failure in
Figure~\ref{fig:ablation-scale-collapse}, Relation-only predicts
0.0013--0.0036 yet satisfies 15 of 16 decrease comparisons. All three models
use the same 16 uniformly sampled frames (filled circles), compared with
their own reference predictions. Progress-only satisfies 2 of 16, while
joint supervision satisfies 14 of 16; both retain substantial score
variation. Correct decrease judgments can therefore coexist with a severely
compressed progress scale.

\subsection{Policy Learning in Simulation}
\label{app:policy-learning-details}

\subsubsection{Reward-Model Adaptation}
\label{app:reward-adaptation}

We adapt reward models for four LIBERO suites: Spatial,
Object, Goal, and LIBERO-10, each containing ten tasks.
Event2Reward starts from the model obtained after rollout joint training
described in Section~\ref{sec:joint-learning}.
Robometer~\citep{liang2026robometer} and
Robo-Dopamine~\citep{tan2025robodopamine} start from their
publicly released pretrained checkpoints.

Within each suite, Event2Reward and Robometer use the same
200 adaptation trajectories.
The Spatial set contains 114 expert demonstrations,
58 suboptimal successful rollouts, and 28 failed rollouts;
the Object, Goal, and LIBERO-10 sets each contain 75, 64,
and 61 trajectories of these categories, respectively.
Suboptimal and failed rollouts are collected using
\texttt{pi0\_libero} and manually annotated and checked.
Each method follows its own adaptation recipe consistently
across suites for 500 updates, and the final checkpoint is
used for policy learning.
Event2Reward uses temporal labels and auxiliary trend supervision
on expert demonstrations, together with progress relations derived
from annotated events in suboptimal and failed executions.
Robometer retains its progress, preference, and success objectives.
Table~\ref{tab:reward-adaptation-settings} summarizes the optimization
settings used consistently across the four suites.

\begin{table}[t]
    \centering
    \caption{Reward-model adaptation settings for LIBERO.
    Learning rates are base values before scheduling.
    Event2Reward batches count flattened input clips (model queries),
    whereas Robometer batches count trajectory pairs.
    Global batches include all devices and gradient accumulation.}
    \label{tab:reward-adaptation-settings}
    \begin{adjustbox}{max width=\linewidth}
\begin{tabular}{lcc}
        \toprule
        Setting & Event2Reward & Robometer \\
        \midrule
        Optimizer updates & 500 & 500 \\
        Optimizer & AdamW & AdamW \\
        Language learning rate & $2\times10^{-6}$ & $2\times10^{-5}$ \\
        Vision learning rate & $5\times10^{-7}$ & $2\times10^{-5}$ \\
        Prediction-head learning rate & $10^{-5}$ & $2\times10^{-5}$ \\
        Trainable language parameters & All & All \\
        Trainable vision modules & Last 3 blocks and merger & Entire encoder \\
        Trainable prediction heads & Progress, trend
            & \shortstack{Progress, preference,\\success} \\
        Per-device microbatch & 10 clips & 8 pairs \\
        Devices & 2 & 2 \\
        Gradient accumulation & 4 & 1 \\
        Global batch per update & 80 clips & 16 pairs \\
        Warmup updates & 25 & 50 \\
        Schedule after warmup & Cosine & Cosine \\
        Weight decay & 0.01 & 0.01 \\
        \bottomrule
    \end{tabular}
    \end{adjustbox}
\end{table}

Robo-Dopamine uses task-specific one-shot adaptation.
For each task, we select one expert demonstration from
the official LIBERO data and adapt the pretrained model
for two epochs using its released training objective.
The final frame of the demonstration serves as the goal image.
This produces one reward model per task, whereas Event2Reward and
Robometer each use one adapted model per suite.
Sparse uses only the environment reward.
All reward models remain fixed during policy learning,
and each adapted model is reused across environment seeds.

\subsubsection{Policy Learning and Evaluation}
\label{app:simulation-rl}
\label{app:policy-learning-metrics}
For each suite, we jointly fine-tune
$\pi_{0.5}$~\citep{physicalintelligence2025pi05} on one expert demonstration
from each of its ten tasks. Within each task, all reward conditions share
the frozen base policy, while task-specific residual policies are trained
by RL.

We use the residual SAC framework and online-update protocol in
Appendix~\ref{app:ablation}, with action-chunk length $H=5$.
LIBERO-Goal and LIBERO-Object use residual scale $\alpha=0.2$
and discount $\gamma=0.998$; LIBERO-Spatial uses $\alpha=0.5$
and $\gamma=0.99$; LIBERO-10 uses $\alpha=0.1$
and $\gamma=0.998$.
Reward construction and all other SAC settings are unchanged.
Reward-model adaptation trajectories are excluded from policy replay.

For online reward inference, Event2Reward uses the task instruction,
the episode's initial main-camera image, a fixed goal image from
the same view, and the latest five consecutive main-camera
observations.
Robometer uses the instruction and up to eight main-camera frames
sampled from the observed trajectory prefix, taking the
current-frame progress clipped to $[0,1]$.
Robo-Dopamine uses forward inference with fixed demonstration
reference and goal images, together with main- and wrist-camera
observations at the episode start and current step.
Its percentage score is divided by 100 and clipped to $[0,1]$.

The environment provides a sparse reward of 1 upon success and
0 otherwise, with no additional failure penalty.
Episodes terminate upon success or are truncated after 220, 280,
300, and 520 environment steps for Spatial, Object, Goal, and
LIBERO-10, respectively, in both training and independent evaluation.

Each method is evaluated with five environment seeds
per task, following the seeds and independent-evaluation protocol in
Appendix~\ref{app:ablation}. Quantitative budgets are 200k training
environment steps per run for Spatial, Object, and Goal, and 1.2M for
LIBERO-10. The first three suites' curves extend to 240k steps, but
quantitative statistics exclude evaluations beyond 200k.
Evaluation interactions are excluded from both the training budget and replay.

\paragraph{Metrics.}
Policy success is taken from the last periodic independent
evaluation within the corresponding suite's training budget.
Since evaluations are triggered by completed episodes,
this evaluation need not occur exactly at the budget endpoint.
We first average success rates over the ten tasks within
each seed, then report the mean and sample standard
deviation across seeds. Policy success, AUC, and steps to target success
use independent policy evaluations.

\paragraph{Normalized evaluation AUC.}
For each run, we compute normalized evaluation AUC as in
Equation~\ref{eq:ablation-eval-auc} using the corresponding
interaction budget: $T=200{,}000$ for Spatial, Object, and Goal,
and $T=1{,}200{,}000$ for LIBERO-10.
We use the recorded step-zero base-policy evaluation
and linearly interpolate between subsequent evaluations
within the comparison budget.
The last observed success rate is held constant from its
evaluation step to $T$, so no evaluation beyond the budget
contributes to the integral.
We first average the ten task AUCs within each seed, then
report the mean and sample standard deviation across the
five seeds in Table~\ref{tab:simulation-auc}.

\begin{table}[t]
    \centering
    \caption{\textbf{Normalized evaluation AUC on LIBERO.}
        AUC is normalized by the training budget: 200k environment steps
        for Spatial, Object, and Goal, and 1.2M for LIBERO-10.
        Values (\%) are reported as mean $\pm$ sample standard deviation
        across five environment seeds, after averaging the ten task AUCs
        within each seed. Robometer denotes the adapted model.
        Bold indicates the highest mean in each suite.}
    \label{tab:simulation-auc}
    \begin{adjustbox}{max width=\linewidth}
\begin{tabular}{@{}lcccc@{}}
        \toprule
        Method & LIBERO-Spatial & LIBERO-Object & LIBERO-Goal & LIBERO-10 \\
        \midrule
        Sparse & $22.44 \pm 3.67$ & $20.90 \pm 1.21$ & $40.10 \pm 1.65$ & $15.61 \pm 5.18$ \\
        Robo-Dopamine & $22.44 \pm 1.67$ & $24.42 \pm 1.94$ & $37.20 \pm 1.78$ & $16.21 \pm 3.34$ \\
        Robometer & $20.05 \pm 2.77$ & $24.83 \pm 3.63$ & $33.72 \pm 2.49$ & $17.32 \pm 6.04$ \\
        Event2Reward & $\mathbf{25.55} \pm 1.66$ & $\mathbf{33.06} \pm 2.28$ & $\mathbf{44.91} \pm 1.51$ & $\mathbf{24.91} \pm 4.31$ \\
        \bottomrule
    \end{tabular}
    \end{adjustbox}
\end{table}

\paragraph{Sample efficiency.}
We measure the training interactions needed to first reach
a target suite-average success rate.
At each recorded evaluation step, we use the most recent
available evaluation for every run, including each run's
own step-zero evaluation before its first subsequent evaluation.
We average the ten task scores within each seed and then
average the five seed means.
The first step at which this suite-average curve reaches
or exceeds the target is the reported crossing time.
This is a crossing of the aggregate curve, not an average
of per-task or per-run crossing times.
Table~\ref{tab:simulation-target-steps} reports targets of
50\%, 60\%, and 70\%.

For each target and suite, we compare Event2Reward with the
baseline that reaches the target in the fewest interactions.
If both have a recorded crossing, the percentage reduction is
$100(1-\tau_{\mathrm{Event2Reward}}/\tau_{\mathrm{baseline}})$,
using unrounded crossing steps.
At 60\% success, these steps are 132,806 versus 156,630
for Event2Reward and Robo-Dopamine on LIBERO-Goal, and
160,052 versus 199,163 for Event2Reward and Robometer on
LIBERO-Object, giving reductions of 15.2\% and 19.6\%.
If a method has no recorded crossing within the comparison
budget, we mark the entry as unreached and do not substitute
the budget endpoint as a crossing time.

\begin{table}[t]
    \centering
    \caption{\textbf{Environment interactions to first reach a target
        suite-average success rate.}
        The comparison budget is 200k environment steps for Spatial,
        Object, and Goal, and 1.2M for LIBERO-10.
        Each suite-average curve averages the ten tasks within each
        environment seed and then averages the five seed means.
        A dash denotes no recorded crossing within the budget.
        Bold indicates the earliest crossing in each suite at each target.}
    \label{tab:simulation-target-steps}
    \begin{adjustbox}{max width=\linewidth}
\begin{tabular}{@{}lcccc@{}}
        \toprule
        Method & LIBERO-Spatial & LIBERO-Object & LIBERO-Goal & LIBERO-10 \\
        \midrule
        \multicolumn{5}{l}{\textbf{(a) Target: 50\% mean success}} \\
        \addlinespace[2pt]
        Sparse & -- & -- & $107{,}329$ & -- \\
        Robo-Dopamine & -- & $173{,}777$ & $126{,}216$ & -- \\
        Robometer & -- & $177{,}230$ & $162{,}312$ & -- \\
        Event2Reward & -- & $\mathbf{141{,}119}$ & $\mathbf{102{,}873}$ & $\mathbf{1{,}149{,}903}$ \\
        \midrule
        \multicolumn{5}{l}{\textbf{(b) Target: 60\% mean success}} \\
        \addlinespace[2pt]
        Sparse & -- & -- & $162{,}154$ & -- \\
        Robo-Dopamine & -- & -- & $156{,}630$ & -- \\
        Robometer & -- & $199{,}163$ & -- & -- \\
        Event2Reward & -- & $\mathbf{160{,}052}$ & $\mathbf{132{,}806}$ & -- \\
        \midrule
        \multicolumn{5}{l}{\textbf{(c) Target: 70\% mean success}} \\
        \addlinespace[2pt]
        Sparse & -- & -- & -- & -- \\
        Robo-Dopamine & -- & -- & -- & -- \\
        Robometer & -- & -- & -- & -- \\
        Event2Reward & -- & $\mathbf{178{,}441}$ & $\mathbf{164{,}080}$ & -- \\
        \bottomrule
    \end{tabular}
    \end{adjustbox}
\end{table}

\subsubsection{Complete Task Results}
\label{app:simulation-task-results}

\begin{figure}[p]
    \centering
    \includegraphics[width=\linewidth]{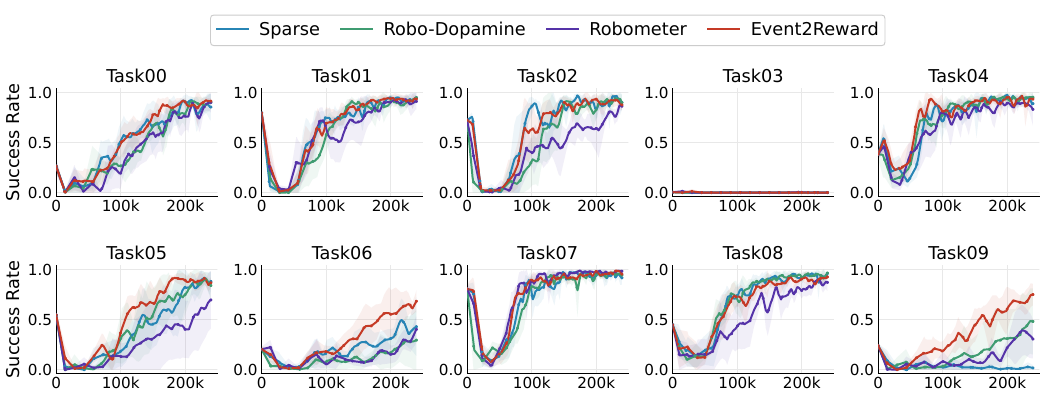}
    \caption{\textbf{Independent evaluation on all ten LIBERO-Goal tasks.}
        Task IDs follow the original suite ordering.
        The horizontal axis shows training environment steps.
        Curves extend to 240k steps; quantitative comparisons use 200k.
        Lines and shaded bands show the mean and $\pm1.96\,\mathrm{SEM}$
        across five environment seeds.}
    \label{fig:full-goal-eval}
\end{figure}

\begin{figure}[p]
    \centering
    \includegraphics[width=\linewidth]{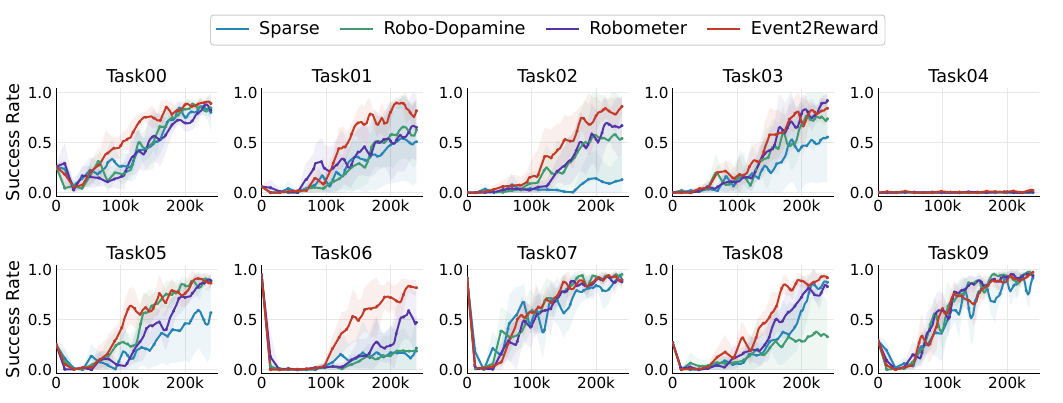}
    \caption{\textbf{Independent evaluation on all ten LIBERO-Object tasks.}
        Task IDs follow the original suite ordering.
        The horizontal axis shows training environment steps.
        Curves extend to 240k steps; quantitative comparisons use 200k.
        Lines and shaded bands show the mean and $\pm1.96\,\mathrm{SEM}$
        across five environment seeds.}
    \label{fig:full-object-eval}
\end{figure}

\begin{figure}[p]
    \centering
    \includegraphics[width=\linewidth]{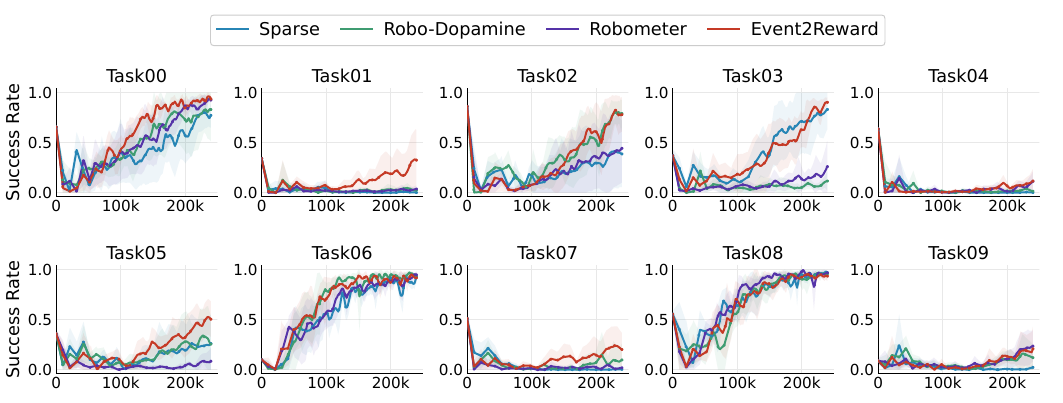}
    \caption{\textbf{Independent evaluation on all ten LIBERO-Spatial tasks.}
        Task IDs follow the original suite ordering.
        The horizontal axis shows training environment steps.
        Curves extend to 240k steps; quantitative comparisons use 200k.
        Lines and shaded bands show the mean and $\pm1.96\,\mathrm{SEM}$
        across five environment seeds.}
    \label{fig:full-spatial-eval}
\end{figure}

\begin{figure}[p]
    \centering
    \includegraphics[width=\linewidth]{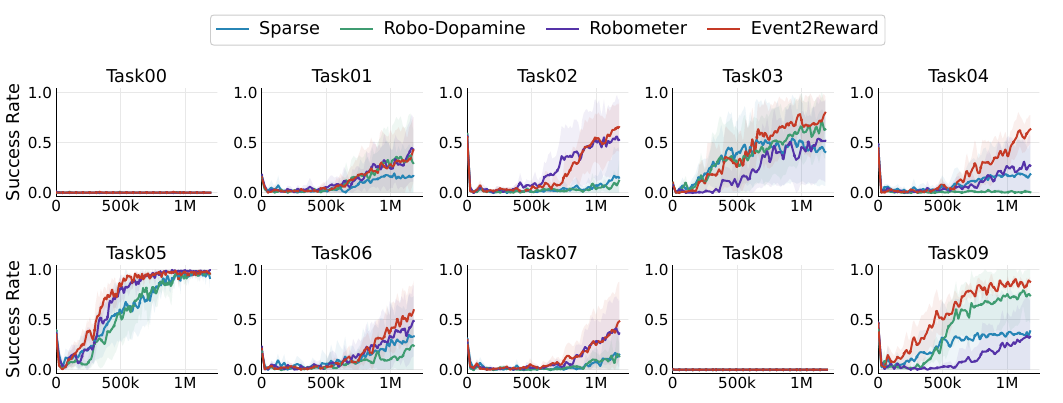}
    \caption{\textbf{Independent evaluation on all ten LIBERO-10 tasks.}
        Task IDs follow the original suite ordering.
        The horizontal axis shows training environment steps within
        the 1.2M-step budget; each task is plotted over its fixed
        common observation interval without extrapolation.
        Lines and shaded bands show the mean and $\pm1.96\,\mathrm{SEM}$
        across five environment seeds.}
    \label{fig:full-libero10-eval}
\end{figure}

Figures~\ref{fig:full-goal-eval}, \ref{fig:full-object-eval},
\ref{fig:full-spatial-eval}, and~\ref{fig:full-libero10-eval}
show independent-evaluation curves for all ten tasks in each suite,
ordered Task00--Task04 (top) and Task05--Task09 (bottom).
Curves show means across five environment seeds with
$\pm1.96\,\mathrm{SEM}$ bands clipped to $[0,1]$. We linearly interpolate
between observations on a shared grid containing recorded evaluation
steps, without additional smoothing or extrapolation. Step zero evaluates
the shared frozen base policy with zero residual actions.

\begin{table}[ht]
    \centering
    \footnotesize
    \renewcommand{\arraystretch}{0.85}
    \captionsetup{skip=4pt}
    \caption{\textbf{Tasks shown in Figure~\ref{fig:simulation-selected-tasks}.}
    Task IDs follow Figures~\ref{fig:full-goal-eval}--\ref{fig:full-libero10-eval};
    on each task, Event2Reward improves final success over the strongest baseline.}
    \label{tab:figure4-tasks}
    \begin{tabular}{llll}
    \toprule
    Figure~\ref{fig:simulation-selected-tasks} & Suite & Task ID & Instruction \\
    \midrule
    Task 1 & LIBERO-Object & Task02 & pick up the salad dressing and place it in the basket \\
    Task 2 & LIBERO-Goal & Task06 & put the cream cheese in the bowl \\
    Task 3 & LIBERO-Goal & Task09 & put the wine bottle on the rack \\
    Task 4 & LIBERO-Object & Task08 & pick up the chocolate pudding and place it in the basket \\
    Task 5 & LIBERO-Object & Task00 & pick up the alphabet soup and place it in the basket \\
    Task 6 & LIBERO-Goal & Task05 & push the plate to the front of the stove \\
    Task 7 & LIBERO-Object & Task01 & pick up the cream cheese and place it in the basket \\
    Task 8 & LIBERO-Object & Task03 & pick up the bbq sauce and place it in the basket \\
    \bottomrule
    \end{tabular}
\end{table}

Goal, Object, and Spatial curves extend through 240k environment steps,
while quantitative comparisons use 200k. LIBERO-10 curves stay within
the 1.2M-step budget and each task's fixed common observation interval,
ending at its last supported step rather than extending to the nominal
budget endpoint. This plotting endpoint differs from each run's last
periodic evaluation used in Table~\ref{tab:libero-policy-performance}.
AUC and target-crossing statistics follow
Appendix~\ref{app:policy-learning-metrics}.

On some tasks, success initially drops when residual learning begins.
This observation is consistent with the early exploration behavior
discussed in PLD~\citep{xiao2025pld}, where the residual policy starts
to deviate from the base policy and visits potentially suboptimal states.

\clearpage

\clearpage

\clearpage

\subsubsection{Rollout Joint-Training Ablation}
\label{app:rollout-joint-ablation}

\paragraph{Training controls.}
Event2Reward w/o rollout joint training uses the 50,000-update pretraining
checkpoint; Event2Reward uses the 10,000-update joint-training checkpoint
initialized from it. Both share Qwen3.5-4B and the target-adaptation
objective. All conditions use 500 adaptation updates, 80 queries per
update, and training seed 42. With 200 trajectories, the ablated model
uses eight devices and one accumulation step; all other conditions use
two devices and four accumulation steps, preserving the global batch.
Joint training combines real rollouts, 5,452 simulated RoboFAC rollouts,
and pretraining-data replay, adding 10,000 updates (800,000 queries;
Appendix~\ref{app:training-configuration}). The comparison tests this
stage as a whole, without separately controlling data source, scale,
supervision, or additional compute.

\paragraph{Fixed adaptation subset.}
The 40-trajectory subset contains 15 normal successful trajectories
and 25 event-bearing trajectories, sampled from the same 200-trajectory
Object adaptation set used in Section~\ref{sec:simulation}.
Normal successes cover all ten tasks, with at least one per task.
The four event categories retain 7, 5, 11, and 2 trajectories for
no further progress, unrecovered regression, progress stall, and
recoverable regression, respectively.
Event counts follow the original category proportions, with sampling
stratified by task. All variants use the same subset, fixed before
training without consulting model predictions or RL outcomes,
and share ten fixed goal images;
nine are sourced outside the 40-trajectory subset. The reduced-data
condition therefore uses 40 adaptation trajectories plus these shared
goal images.

\paragraph{RL and evaluation.}
Each condition uses one reward-model checkpoint and 50 RL runs on ten Object
tasks with environment seeds $\{7,8,9,17,27\}$. All share the frozen Object
one-shot $\pi_{0.5}$ checkpoint at step 49,999 and Object residual SAC
settings from Appendix~\ref{app:policy-learning-details}: residual scale
$0.2$, discount $0.998$, and chunk horizon five. Runs train for 300k
environment steps, while Table~\ref{tab:rollout-joint-ablation} retains the
200k comparison budget. Independent evaluation and last-periodic success
follow Appendix~\ref{app:policy-learning-metrics}. AUC follows that section's
fixed-budget rule: it includes each run's initial evaluation, interpolates
between in-budget observations, and holds the last value to the budget
endpoint; no out-of-budget observation contributes.

\paragraph{Uncertainty and scope.}
Pairing the task-averaged results by environment seed, the 200k AUC gain
from rollout joint training is $+3.39$ percentage points with 200
adaptation trajectories, with a nominal 95\% paired $t$ interval of
$[-2.46,9.24]$; with 40 trajectories, it is $+4.13$ points with interval
$[-0.63,8.89]$ (four degrees of freedom). These intervals reflect
environment-seed variation conditional on one reward-model training seed
and one fixed 40-trajectory subset. The positive mean differences do not
establish statistical significance or equivalent performance after
reducing adaptation data by 80\%. Both methods already include the same
pretraining and relation-supervised target adaptation; this comparison
addresses the incremental rollout joint-training stage.

\paragraph{Full training budget.}
Table~\ref{tab:rollout-joint-ablation-300k} reports the same metrics at
the actual 300k-step training budget. Rollout joint training increases
mean AUC by 3.23 points with 200 trajectories and 4.77 points with 40
trajectories; last periodic evaluation success increases by 4.44 and 6.20
points, respectively. The final periodic evaluations occur between
290,617 and 300,000 environment steps. Benefits vary across tasks:
Task02 and Task06 account for 5.68 points of the 6.20-point mean success
gain in the 40-trajectory comparison. These results complement the fixed
200k comparison rather than changing its reporting budget.

\begin{table}[!htbp]
    \centering
    \setlength{\tabcolsep}{4pt}
    \caption{\textbf{Rollout joint-training ablation at the full 300k budget.}
        Evaluation AUC and last periodic evaluation success (\%) use
        the same task and seed aggregation as
        Table~\ref{tab:rollout-joint-ablation}. Values are mean $\pm$
        sample standard deviation across five environment seeds.}
    \label{tab:rollout-joint-ablation-300k}
    \begin{adjustbox}{max width=\linewidth}
\begin{tabular}{@{}lccc@{}}
        \toprule
        Method & \shortstack{Adaptation\\trajectories} & Eval AUC & Success \\
        \midrule
        Event2Reward w/o rollout joint training & 200 & $45.49 \pm 2.25$ & $80.20 \pm 0.62$ \\
        Event2Reward & 200 & $48.72 \pm 1.60$ & $84.64 \pm 1.28$ \\
        \midrule
        Event2Reward w/o rollout joint training & 40 & $42.28 \pm 4.00$ & $77.72 \pm 9.06$ \\
        Event2Reward & 40 & $47.05 \pm 2.54$ & $83.92 \pm 2.06$ \\
        \bottomrule
    \end{tabular}
    \end{adjustbox}
\end{table}

\begin{figure}[!htbp]
    \centering
    \includegraphics[width=\linewidth]{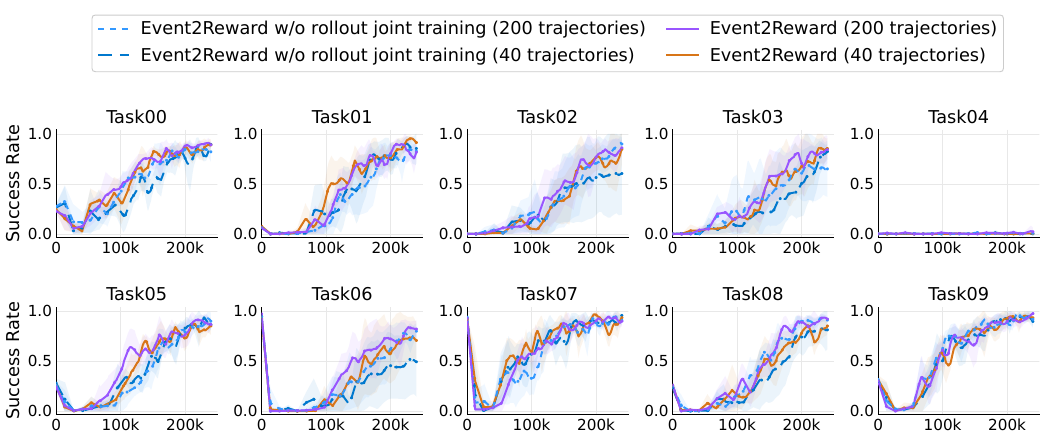}
    \caption{\textbf{Rollout joint-training ablation: independent evaluation.}
        Each task compares Event2Reward with Event2Reward w/o rollout joint
        training using 200 or 40 adaptation trajectories.
        Lines show means across five environment seeds and bands show
        $\pm1.96\,\mathrm{SEM}$, clipped to $[0,1]$.
        Curves extend to 240k training environment steps; the main
        quantitative comparison uses 200k.
        Curves are linearly interpolated only within the current runs'
        common observed interval for each task, with no additional
        smoothing or extrapolation.}
    \label{fig:object-stage2-eval}
\end{figure}

\FloatBarrier
\section{Real-World Tasks and Evaluation Protocols}
\label{app:real-world-details}

\subsection{Insertion Tasks}
\label{app:real-world-insertion-tasks}

The three SERL tasks use the insertion setups shown in
Figure~\ref{fig:real-world-tasks} (top). Insertion success is judged
by a human operator according to the task-specific criteria below.

\paragraph{Plug Insert.}
The robot starts with the plug already held in its gripper, aligns it
with the target socket, and inserts it fully. An episode succeeds when
the plug is fully seated in the socket. The maximum episode length is
190 steps.

\paragraph{RAM Insert.}
The robot starts with the RAM module already held in its gripper, aligns
it with the target memory slot, and presses it into place. An episode
succeeds when the module is fully inserted and seated; engagement of the
retaining clips is not required. The task has an insertion tolerance of
approximately 2\,mm. The maximum episode length is 400 steps.

\paragraph{Ethernet Insert.}
The connector initially rests on the tabletop. The robot must grasp it,
lift the connector and the attached cable, and insert it into the target
Ethernet port. An episode succeeds when the connector is fully seated
in the port. The maximum episode length is 300 steps.

\subsection{Object-Manipulation Tasks}
\label{app:real-world-object-tasks}

The three DSRL tasks use a $\pi_0$ base policy on Franka robots.
Success is assessed at the episode level according to the task-specific
criteria below.

\paragraph{Stack Bowls.}
Three bowls are placed on the tabletop, with one bowl already at the
central target position. The robot must place the other two bowls into
the central bowl to form a stack of three. An episode succeeds only when
all three bowls form a stable stack after the gripper releases them.
The maximum episode length is 800 environment steps.

\paragraph{Press Buttons.}
Blue, pink, and yellow buttons are placed on the tabletop. The robot
must press them downward in the order blue, pink, then yellow. An episode
is successful only when all three buttons are pressed in this order.
The maximum episode length is 500 environment steps.

\paragraph{Plate Vegetables.}
The robot must pick up a cabbage, an ear of corn, and a pepper from the
tabletop and place all three on the target plate. An episode succeeds
only when all three objects remain stably on the plate after the gripper
releases them. The maximum episode length is 800 environment steps.

\subsection{Training and Evaluation Protocols}
\label{app:real-world-protocols}

\paragraph{SERL offline demonstrations.}
Each SERL insertion task uses an offline buffer of 20 demonstrations,
shared by all four reward conditions for policy learning.
We do not perform behavior-cloning pretraining.

\paragraph{SERL initialization and autonomous rollouts.}
Each SERL run initializes its online replay buffer with 10 newly collected
human-assisted trajectories, using the same protocol across reward
conditions. Subsequent training and evaluation are autonomous.
Initialization time counts toward the training budget, but initialization
trajectories are excluded from rollout-success statistics.

\paragraph{DSRL initialization and online learning.}
For each task, we use the 20 policy demonstrations to fine-tune $\pi_0$
with LoRA. All reward conditions for that task start from the same
fine-tuned checkpoint. During online reinforcement learning, $\pi_0$
remains frozen and only the latent-noise control policy is trained.
DSRL uses action chunks of 25 environment steps.
Online exploration and evaluation are autonomous.

\paragraph{Reward-model adaptation.}
Robo-Dopamine follows its official one-shot adaptation protocol.
Robometer and Event2Reward share five successful and five suboptimal
adaptation trajectories per task. For Event2Reward, we follow
Appendix~\ref{app:annotation-procedure} to annotate suboptimal
executions with event types, reference states, event intervals, and
recovery states where applicable, without framewise absolute progress
scores. With our annotation tool, annotating all five suboptimal
trajectories of a task, from opening the tool to completing the last
annotation, took about one minute of human effort. Both methods use 500 updates with their respective
simulation-adaptation learning rates, batch sizes, and trainable parameter
groups (Appendix~\ref{app:reward-adaptation}). Final adapted checkpoints
remain frozen throughout online policy learning.

\paragraph{Reward annotation and computation.}
The sparse task reward is 1 upon human-judged task success and 0
otherwise, with no additional failure penalty.
Learned-reward conditions use potential-based reward shaping (PBRS), with
predicted progress as the potential, following
Equation~\ref{eq:progress-reward}. Shaping rewards are computed at
action-chunk boundaries with $\beta=1$ and $\gamma=0.999$.
The potential at absorbing terminal states is set to zero.
For SERL with learned rewards, rewards for offline demonstration
trajectories are precomputed with the corresponding reward model before
the transitions enter the offline buffer. During online collection, the
actor completes an episode without waiting for reward inference. Rewards
for the collected trajectory are computed during the subsequent reset,
and the transitions enter the replay buffer once their rewards are available.

\paragraph{Reset randomization and episode termination.}
Each insertion task randomizes the starting position at every reset with
independent perturbations of approximately 1\,cm along each of the
$X$, $Y$, and $Z$ axes. Episodes end upon success, declared failure, or
the applicable step limit. Declared failure triggers a reset, with no
further attempt in that episode. During SERL training, a human operator
may terminate substantially deviating, potentially unsafe motion;
this ends the episode without providing corrective actions.
Our experiments evaluate task performance in the reported settings and do not
establish safety for deployment beyond these settings.

\paragraph{Training budget and independent evaluation.}
Each reward condition uses three independent runs per task, each with
30 minutes of online wall-clock training, including initial buffer
warm-up where applicable, resets, and reward computation. Each 30-minute
checkpoint is evaluated over 20 autonomous episodes (60 per task and
condition across the three policies). We report the mean and sample
standard deviation of per-policy success rates across runs; setting-level
means weight the three tasks equally before rounding. Human-aborted
episodes count as failures and remain in the denominator.

\paragraph{Plug-insertion training-time comparison.}
One additional run per method uses the same SERL framework and measures
wall-clock time to the first ten consecutive successful autonomous
training episodes: 15 minutes for Event2Reward and 45 minutes for Sparse.
Only post-warm-up autonomous episodes enter the window; both times
include warm-up collection and resets. This training-rollout metric
differs from the three-run independent evaluation after 30 minutes
in Figure~\ref{fig:real-world-policy-performance}.

\end{document}